\documentclass[sigconf]{acmart}

\usepackage{tikz}
\usepackage{amsmath}

\usepackage{stfloats}
\usepackage{filecontents}
\usepackage{verbatim}
\usepackage{enumerate}

\usepackage{epsfig} 
\usepackage{setspace}
\usepackage{balance}
\usepackage{amsfonts}
\usepackage{epsfig}  
\usepackage{amssymb}
\usepackage[normalem]{ulem}
\usepackage{amscd}
\usepackage{amsmath} 
\usepackage{microtype}
\usepackage{array}
\usepackage{chngpage}
\usepackage{graphicx}
\usepackage{epsfig}
\usepackage{multirow}
\usepackage{caption}
\usepackage{listings}
\usepackage{subfigure}
\usepackage{adjustbox}
\usepackage{ulem}
\usepackage{lipsum}
\usepackage{float}
\usepackage{pifont}
\usepackage{bm}
\usepackage{color,xcolor}
\usepackage{booktabs} 
\usepackage{verbatim}
\usepackage{amssymb}
\usepackage{pifont}
\usepackage{array,multirow}
\usepackage{makecell}
\usepackage{tikz}
\usepackage{CJK}
\usepackage{appendix}  
\usepackage{float}
\usepackage{threeparttable}
\usepackage{colortbl}
\usepackage{bm} 
\usepackage{titlesec}
\usepackage{appendix}
\usepackage{cleveref}
\usepackage[utf8]{inputenc}
\crefname{section}{§}{§§}
\Crefname{section}{§}{§§}
\usepackage{fancyhdr}

\newcommand{\reftable}[1]{Table \ref{#1}}
\newcommand{\tabincell}[2]{\begin{tabular}{@{}#1@{}}#2\end{tabular}}
\newcommand{\ignore}[1]{}

\def\BibTeX{{\rm B\kern-.05em{\sc i\kern-.025em b}\kern-.08emT\kern-.1667em\lower.7ex\hbox{E}\kern-.125emX}}
    
\copyrightyear{2019} 
\acmYear{2019} 
\acmConference[CCS '19]{2019 ACM SIGSAC Conference on Computer and Communications Security}{November 11--15, 2019}{London, United Kingdom}
\acmBooktitle{2019 ACM SIGSAC Conference on Computer and Communications Security (CCS '19), November 11--15, 2019, London, United Kingdom}
\acmPrice{15.00}
\acmDOI{10.1145/3319535.3354259}
\acmISBN{978-1-4503-6747-9/19/11}

\begin{document}
\cfoot{\thepage}

\title{Seeing isn't Believing: Towards More Robust Adversarial Attack Against Real World Object Detectors}

\author[els]{Yue Zhao$^{1,2}$, Hong Zhu$^{1,2}$, Ruigang Liang$^{1,2}$, Qintao Shen$^{1,2}$, Shengzhi Zhang$^{3}$, Kai Chen$^{1,2}$}

\authornote{Corresponding Author}

\affiliation{\normalsize{$^{1}$SKLOIS, Institute of Information Engineering, Chinese Academy of Sciences, China}}
\affiliation{$^{2}$School of Cyber Security, University of Chinese Academy of Sciences, China}
\affiliation{$^{3}$Department of Computer Science, Metropolitan College, Boston University, USA}
\email{{zhaoyue, zhuhong, liangruigang, shenqintao}@iie.ac.cn,shengzhi@bu.edu,chenkai@iie.ac.cn}

\begin{abstract}
Recently Adversarial Examples (AEs) that deceive deep learning models have been a topic of intense research interest. Compared with the AEs in the digital space, the physical adversarial attack is considered as a more severe threat to the applications like face recognition in authentication, objection detection in autonomous driving cars, etc. In particular, deceiving the object detectors practically, is more challenging since the relative position between the object and the detector may keep changing. Existing works attacking object detectors are still very limited in various scenarios, e.g., varying distance and angles, etc.

In this paper, we presented systematic solutions to build robust and practical AEs against real world object detectors. Particularly, for Hiding Attack (HA), we proposed the \textit{feature-interference reinforcement (FIR)} method and the \textit{enhanced realistic constraints generation (ERG)} to enhance robustness, and for Appearing Attack (AA), we proposed the \textit{nested-AE}, which combines two AEs together to attack object detectors in both long and short distance. 
We also designed diverse styles of AEs to make AA more surreptitious. Evaluation results show that our AEs can attack the state-of-the-art real-time object detectors (i.e., YOLO V3 and faster-RCNN) at the success rate up to 92.4$\%$ with varying distance from 1$m$ to 25$m$ and angles from $-60^\circ$ to $60^\circ$\footnote{Demos of attacks are uploaded on the website: https://sites.google.com/view/ai-tricker}. Our AEs are also demonstrated to be highly transferable, capable of attacking another three state-of-the-art black-box models with high success rate.

\end{abstract}

%
%



%

%

%
\maketitle


\small{\textbf{ACM Reference Format:}}

\noindent Yue Zhao, Hong Zhu, Ruigang Liang, Qintao Shen, Shengzhi Zhang, Kai Chen. 2019. Seeing isn’t Believing: Towards More RobustAdversarial Attack Against Real World Object Detectors. In \textit{2019 ACMSIGSAC Conference on Computer and Communications Security (CCS ’19), November 11–15, 2019, London, United Kingdom.} ACM, New York, NY, USA, 14 pages. https://doi.org /10.1145/3319535.3354259
\normalsize{}
\vspace{-4pt}

\section{Introduction}
\label{sec:intro}

Object detection deals with recognizing instances of semantic objects from images or video clips, which has been  widely applied in many areas including face detection, object tracking, and safety critical tasks such as autonomous driving and intelligent video surveillance. Especially in autonomous driving systems, object detectors are widely adopted to undertake the perception tasks such as recognizing traffic signs, pedestrians, cars, traffic lights, traffic lanes, etc. However, the last few years have seen the security concerns over object detectors, because DNNs are known to be vulnerable to adversarial examples (AEs). AEs are well-crafted malicious inputs that can deceive DNNs into making wrong predictions.  Early researches mainly focus on studying adversarial examples against the image classifiers in the digital space only, i.e., computing the perturbations, adding them back to the original image, and feeding them directly into classification systems. 
Recently, Tom et al. has shown that the AEs against the image classifiers are also possible in the physical world by taking pictures of them and feeding the pictures into the classifier\cite{brown2017adversarial}.

Compared to the image classifiers, the object detectors are more challenging to attack as the AEs need to mislead not only the label predictions but also the object existence prediction (whether there is an object). More importantly, unlike classifiers always working on stationary images, object detectors are commonly applied in an environment where the relative position between the objects and the object detectors may keep changing due to the relative motion of both, e.g., the object detectors on the fast-moving autonomous driving vehicles or walking pedestrians under the intelligence surveillance systems. On one hand, such relative motion between the objects and the object detectors makes the distance and viewing angle between them change dynamically. On the other hand, the moving objects may cause the surrounding illuminations and/or environmental background to change almost all the time. Both of them will significantly impact the effectiveness of practical AEs, thus demanding more robust AEs against the object detectors.

Until very recently, there are a few studies attacking object detectors in the physical space, e.g., ~\cite{song2018physical}~\cite{DBLP:journals/corr/abs-1804-05810}.
Their main approach to improving the robustness of the generated perturbations is to extend the image transformations (e.g., change the size of AEs to simulate different distances~\cite{brown2017adversarial}). However, due to the capability of the approach, the distances and angles are very limited, e.g., at most 12 meters and $15^\circ$ in ~\cite{DBLP:journals/corr/abs-1804-05810} (More detailed discussion and comparison are presented later in Section~\cref{success rate of HA and AA}). Actually, the object detectors on an autonomous driving vehicle may be able to recognize the traffic lights at the distance about 20$m$\footnote{According to the stopping distance table provided on Queensland Government website~\cite{Queensland}, the braking distance of a car at the speed of 60km/h on a dry road is 20$m$.}, and traffic signs at the roadside over the angles of $30^\circ$. So practically attacking object detectors requires the AEs be effective at the longer distance and wider angle. Moreover, these studies are also limited in exploring the impact of illuminations and background on the AEs against object detectors.
In real world situations, to deceive an intelligent surveillance camera, the moving AEs should continue being effective in various scenarios, e.g., from sunshine to shadow, or from driveway to grass, etc. Unfortunately, to the best of our knowledge, existing adversarial attacks are still far away from robustly deceiving the real world object detectors.

In this paper, we aim to generate robust AEs to attack the state-of-the-art object detectors used in the real world, especially with the long distances, wide angles and various real scenarios. To better demonstrate the improvement over existing studies (e.g.,~\cite{song2018physical}), we consider two existing types of AEs: Hiding Attack (HA), which makes the object detector fail to recognize the object, and Appearing Attack (AA), which makes the object detector mis-recognize the AE as the desire object specified by the attacker. We propose several novel techniques to enhance the robustness of the attack.

Particularly, for HA, we propose two novel techniques to improve robustness: \textit{Feature-interference reinforcement (FIR)} and \textit{Enhanced realistic constraints generation (ERG)}. Rather than optimizing the final prediction layer of DNN, FIR enforces the generated AEs to impact both hidden layers and the final layer. In this way, the features of the target object that attackers want to hide are revised by our AEs at the ``early'' stage in the process of classification, which is shown to be more robust against the changes of physical scenarios. Based on the observation that the object detectors ``remember'' the background of an object (they are trained using the images containing both the object and the background where the object usually appears.), ERG generates AEs using a series of ``reasonable'' backgrounds in an automatic way. We leverage the semantics of the object to search for the reasonable backgrounds on the Internet, and synthesize the object and its transformations (e.g., different sizes and/or angles) together with the reasonable backgrounds. In this way, our AEs are more robust against various real world backgrounds.  

For AA, we propose \textit{nested-AE}, which decouples the task of the varying-distance attack into two pieces: the long distance attack and the short distance attack, and produces two separate AEs\ignore{(a big one and a small one)} accordingly. The two AEs are then assembled in a nested fashion to build a single AE, with one AE targeting the long distance attack and the other one targeting the short distance attack. Finally, we also implement diverse styles of AEs to make them more surreptitious, and the batch-variation to accelerate the convergence during the generation of AEs.

We evaluated the AEs generated by our solutions against multiple state-of-the-art object detectors in different physical environments systematically\footnote{We have contacted the developers of all the object detectors that we successfully attacked in this paper, and are waiting for their responses.}. They can attack YOLO V3~\cite{yolov3} and Faster RCNN~\cite{ren2015faster}, with the success rate over 60$\%$ and 78$\%$ respectively in different outdoor environments. Furthermore, they are robust enough to different distances (from $1m$ to $25m$), shooting angles (from $-60^\circ$ to $60^\circ$), backgrounds (various scenarios both indoor and outdoor) and illuminations (cloudy day and sunny day), simultaneously. Compared to previous state-of-the-art studies, the attack distance increases 52\% and the angle increase 75\%. We also measured the transferability of the AEs on other black-box models including SSD (Single Shot Detector)~\cite{liu2016ssd}, RFCN (Region based Fully Convolutional Network)~\cite{dai2016r} and Mask RCNN~\cite{he2017mask}. The success rate is up to 90$\%$ and 72$\%$ for indoors and outdoors, respectively. These results indicate that it is feasible to design robust AEs against real world object detectors, which can be a real threat to autonomous driving cars, intelligent surveillance cameras, etc.

\vspace{5pt}\noindent\textbf{Contributions}. Our contributions are outlined as follows:

\vspace {2pt}\noindent$\bullet$\space\textit{New techniques for generating robust AEs against object detectors.} We proposed \textit{feature-interference reinforcement} and \textit{Enhanced realistic constraints generation}. Such techniques leverage manipulation on the hidden layers in DNN and the semantics of the target object, enabling practical adversarial attacks against object detectors with the varying shooting distances and angles, different backgrounds and illumination in the real world.

\vspace {2pt}\noindent$\bullet$\space\textit{Nested AE}.  We design a new kind of AE, which contains two AEs inside, with each targeting a sub-task of the attack (i.e., the long distance attack and the short distance attack). The two AEs are produced accordingly and then assembled in a nested fashion to build a single AE. Such nested AEs significantly improve the robustness of adversarial attack at the various distances.

\vspace {2pt}\noindent$\bullet$\space We evaluated our AEs generated against multiple state-of-the-art object detectors in different physical environments systematically. Results show that they are robust enough to different distances (from $1m$ to $25m$), shooting angles (from $-60^\circ$ to $60^\circ$), backgrounds and illuminations, simultaneously. Furthermore, our AEs are shown to be highly transferable to other four black-box models.

\section{Background}
\label{sec:background}
In this section, we first overview the existing object detectors, especially the breakthrough in this field due to deep learning. Then we summarize the physical adversarial attacks against image classifiers that are closely related to our attack and the limitations of existing adversarial attacks against the object detectors.

\subsection{Object Detection}
Great progress has been made in recent years on object detection due to convolutional neural networks (CNNs)~\cite{Sermanet2013OverFeat, Girshick2015Fast, redmon2016you}. Modern object detectors based on deep learning methods can be classified into two categories: two-stage strategy detectors such as Faster RCNN~\cite{ren2015faster}, RCNN~\cite{girshick2014rich}, SPPNet~\cite{Kaiming2015Spatial}, Fast RCNN~\cite{Girshick2015Fast}, RFCN~\cite{dai2016r}, Mask RCNN~\cite{he2017mask}, Light Head RCNN~\cite{li2017light}, etc., and one-stage detectors including DetectorNet~\cite{zeng2016deep}, OverFeat~\cite{Sermanet2013OverFeat}, YOLO~\cite{redmon2016you}, YOLO V2 and YOLO 9000~\cite{redmon2017yolo9000}, SSD~\cite{liu2016ssd}, YOLO V3~\cite{yolov3}, etc. Below, we detail YOLO V3 and Faster RCNN from the above two categories respectively.

For YOLO, a one-stage region-based framework, class probabilities and bounding box offsets are predicted directly with a single feed forward CNN network. This architecture leads to a faster processing speed. 
Due to such excellent efficiency and high-level accuracy, YOLO is a good choice in real-time processing systems, such as the traffic light detection module in Apollo~\cite{ApolloPlatform} (an open platform for autonomous driving), and the object detection module in satellite imagery~\cite{DBLP:journals/corr/abs-1805-09512}. Compared with V1 and V2, YOLO V3 improves a lot in the detection of tiny and overlaid objects, which is important for autonomous driving that always needs to detect traffic signs far away at its braking distance. 

Faster RCNN, a two-stage detection framework, includes a pre-processing step for region proposals and a category-specific classification step to determine the category labels of the proposals~\cite{ren2015faster}. Faster RCNN is proposed to improve the RCNN, which is quite computationally expensive in spite of  high object detection accuracy~\cite{Liu2018Deep}. Instead of using the time-consuming selective search algorithm on the feature map to identify the region proposals, Faster RCNN uses a separate network to make the region proposals. Hence Faster RCNN is much faster than its predecessors and can even be used for real-time object detection.

\begin{figure}

\centering
\epsfig{figure=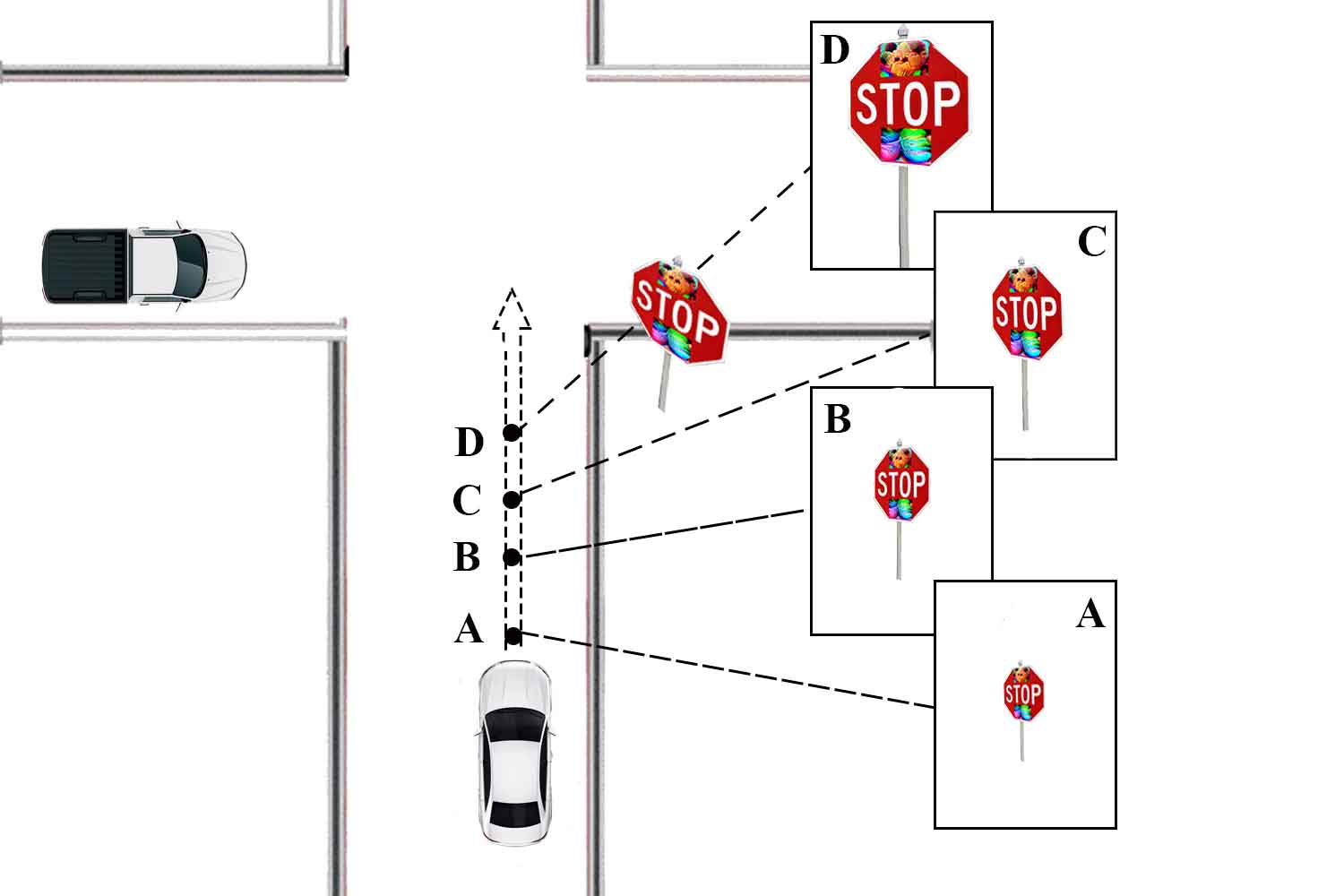, width=0.4\textwidth} 
\caption{\textbf{A Real World Example of Hiding Attack (Stop Sign) Against the Object Detector on an Autonomous Driving Car.}}
\label{Mimic diagram of appearance attack.}
\vspace{-3mm}
\end{figure}

\subsection{Physical Adversarial Examples}
Many researches have explored the adversarial attacks against the image classifiers. In the early works, AEs are studied only in the digital space, but now the physical adversarial attack against deep learning models attracts more attention. For example, in the work of the physical attack against face detection presented in~\cite{sharif2016accessorize}, the authors printed sun glasses that are capable of deceiving the state-of-the-art face recognition system. Tom et al. presented a method to create adversarial image patches that could be printed and attached to the target to fool image classifiers~\cite{brown2017adversarial}. Ivan et al. proposed similar attacks on road signs to deceive image classifiers~\cite{DBLP:journals/corr/EvtimovEFKLPRS17}. These works raised serious safety and security concerns especially for the safety-critical systems.

Successfully attacking one or a few frames in a video stream (using the techniques to fool image classifiers as above) is not enough for object detectors. In contrast, practical adversarial attacks against object detectors should keep adversarial samples work on most, if not all, of the frames. Lu et al.~\cite{lu2017no} demonstrated that physical adversarial samples against image classifiers cannot transfer to object detectors (i.e., YOLO and Faster RCNN) in standard configuration. The recent studies~\cite{song2018physical, DBLP:journals/corr/abs-1804-05810} generated AEs against the object detector under some physical conditions. However, these works are limited in the aspects of the longer distance, multiple angles, various illuminations, etc.

\section{Attack Approach}
\label{Attack approach}

\begin{figure}
\centering
\epsfig{figure=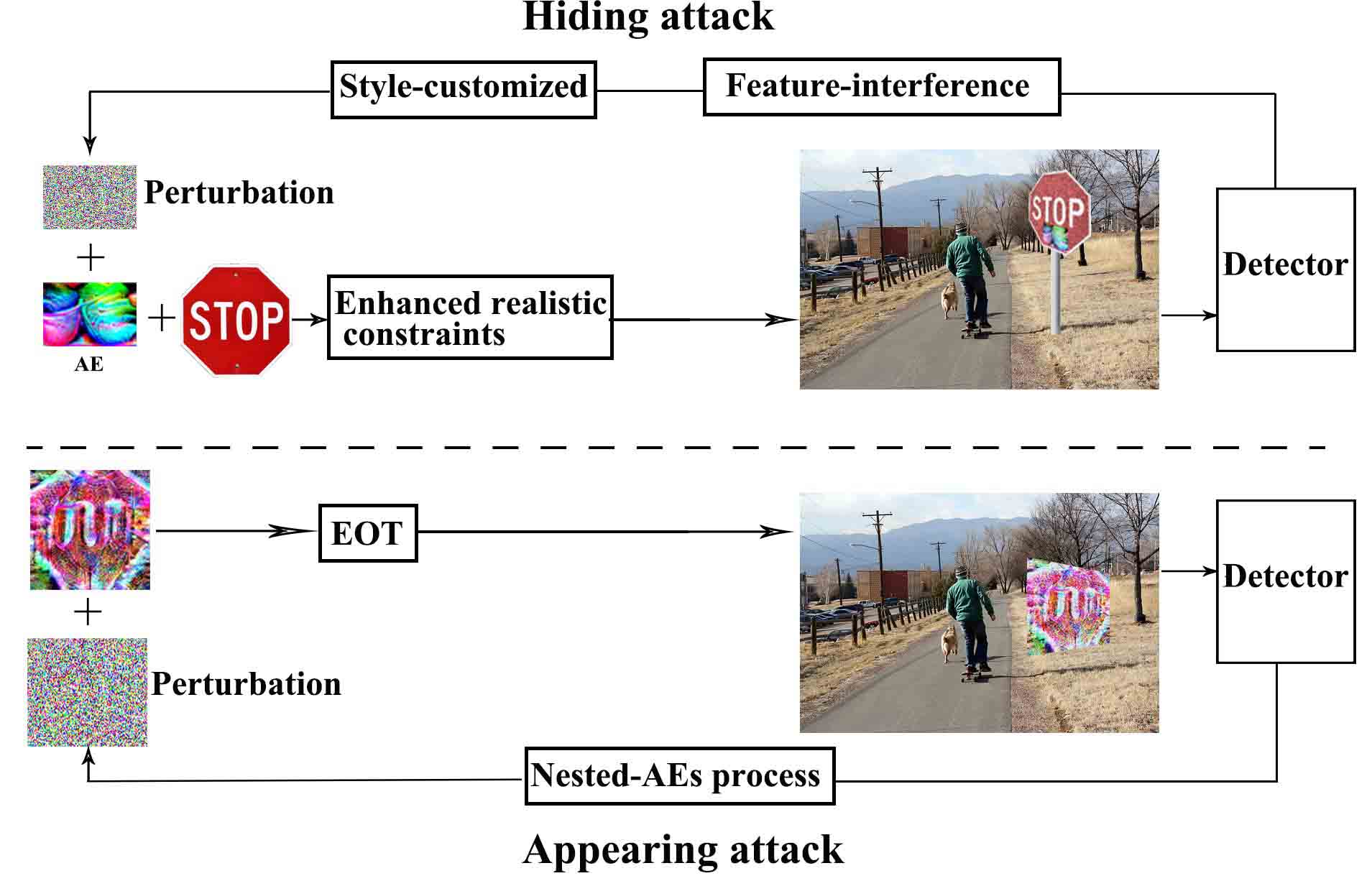, width=0.47\textwidth} 
\caption{\textbf{The Proposed Solutions to Generate Robust AEs (Top: Hiding Attack; bottom: Appearing Attack)}}
\label{Implemantation.}
\vspace{-2mm}
\end{figure}

Building robust AEs against the object detectors in the real world is non-trivial considering the relative motion between objects and detectors, varying environments, etc. Figure~\ref{Mimic diagram of appearance attack.} illustrates a real world example of an adversarial attack (HA of a stop sign) against the object detector on a running autonomous driving car. The distances and angles between the car and the AE (the stop sign) keep changing as the car moves towards the AE. Hence, at different positions, e.g., A, B, C, and D as shown in the figure, the perturbations in the AE captured by the object detector demonstrate different sizes, shapes, reflection of light, etc. Such relative motion between the object and the detector imposes the requirement of highly robust AEs, which although being static themselves, should tolerate reasonable changes in terms of size, shape, illumination, etc.

To generate robust and practical AEs, we proposed a suite of solutions for Hiding Attack (HA) and Appearing Attack (AA) respectively. As shown in Figure~\ref{Implemantation.}, for HA, we proposed feature-interference reinforcement ( FIR, Section~\cref{feature-erosion}) and Enhanced realistic constraints generation (ERG, Section ~\cref{enhanced image transformation}); for AA, we propose nested-AE (Section~\cref{nested AEs}). Finally, we presented the style-customized AEs (Section~\cref{style-customized AEs}) to make them more surreptitious and the batch-variation (Section~\cref{sec:batch-variation momentum}) to accelerate the convergence during the generation of AEs.

\vspace{2mm}
\noindent\textbf{Threat Model.} In this paper, we focus on the white-box adversarial attack, which means we need to access the target model (including its structure and parameters). Meanwhile, we also did some preliminary experiments on the black-box adversarial attack by measuring the transferability of our AEs, where we assume we do not know any details of the target black-box models.

\subsection{Feature-interference Reinforcement}
\label{feature-erosion}
To generate AEs, most of the existing studies design an objective function or a loss function to minimize the difference between the deep learning model's prediction value and the expected value. 
Since object detectors extract the high-dimensional features of the object and give predictions based on those extracted features, AEs can be enhanced by perturbing the original object's features 
``earlier'' in the hidden layers (i.e., before the output layer). Thus, in this paper, besides misleading the prediction results, our loss function is also designed to make the AEs be able to perturb the object’s features at the hidden layers of the model. Such perturbation prevents the original features of the object from transferring to the later layers especially the final layer, thus further misleading the prediction results.

\begin{figure}
\centering
\epsfig{figure=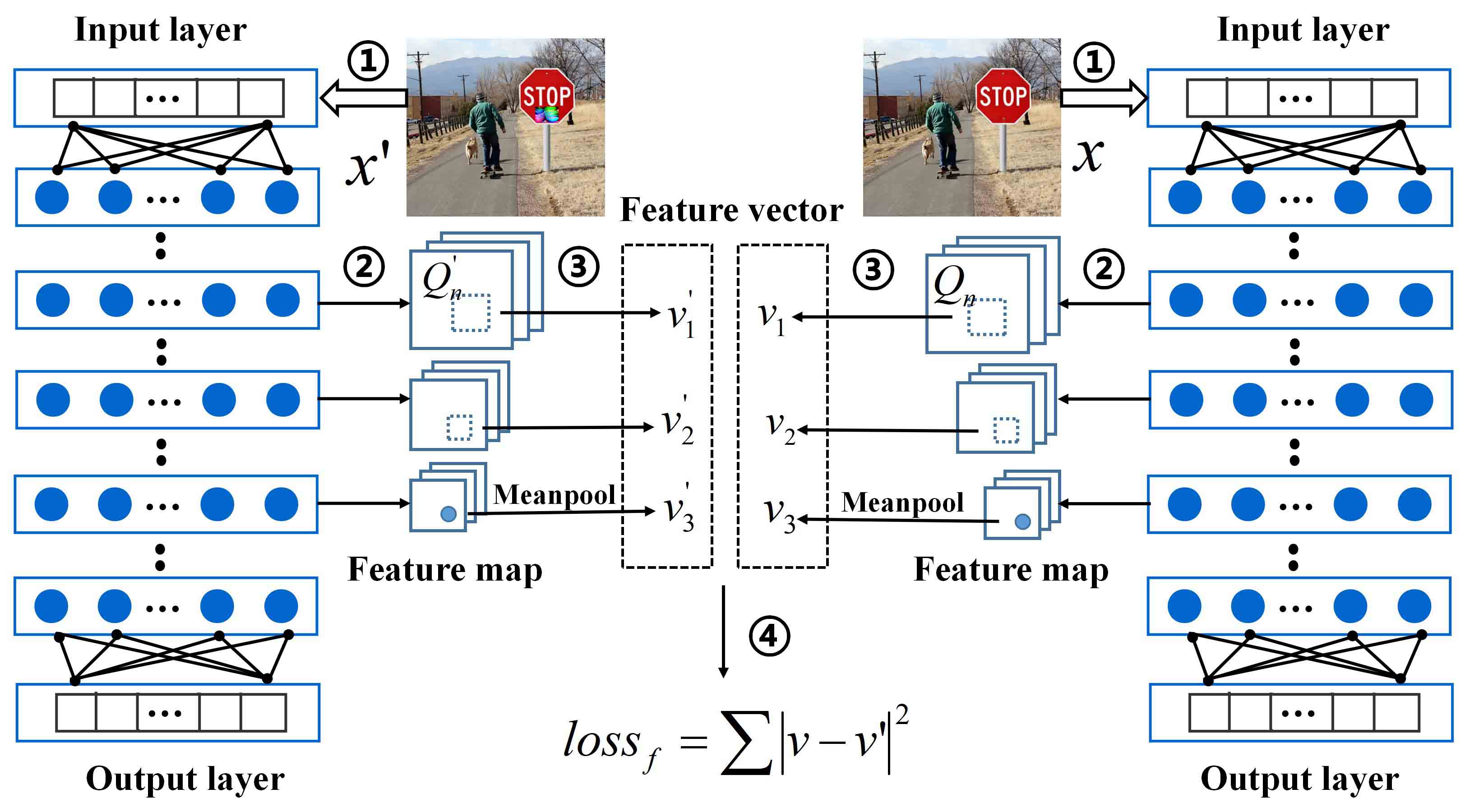, width=0.47\textwidth} 
\caption{\textbf{Feature Interference Reinforcement (Using Stop Sign as an Example).}}
\label{feature-erosion.}
\end{figure}

Figure~\ref{feature-erosion.} shows an example using a stop sign as the attack target to illustrate the high-level idea of feature interference attack (). The two DNN models are identical from the same object detector. Let $Q_n$ denote the DNN hidden layer's feature map from the target object in the input image $x$ (on the right) and $Q^{'}_{n}$ the feature map from the object in the input image $x'$ (on the left). Both of the two input images have the same background, the same stop sign of the same size, the perspective angle and the light condition. The only difference is that the adversarial perturbation is added on the stop sign in $x'$. 

In Step {\textcircled{\small{1}}}, both images $x$ and $x'$ are provided as inputs into DNN model and the corresponding neurons in the hidden layers are activated. Then in Step {\textcircled{\small{2}}}, we extract the feature map $Q_n$ and $Q^{'}_{n}$ from the same hidden layer in an automatic way (We will elaborate  the selection of the hidden layers in the last paragraph of this Section.). The full feature map of the hidden layer is the feature map of the whole input image. Since we only care about the features of the target object, which is the stop sign in this example, we extract a small feature map that is most related to the stop sign based on its relative coordinates in the input image. Such extraction is possible because the hidden layers of the object detectors are composed of convolution layers, which preserve the geometric features of the input image. We set the coordinates of two points of the square region of the stop sign in the input image: the top left point $(x,y)$ and the bottom right point$(x',y')$. Then the coordinates of the small feature map (related to the stop sign) in the full feature map can be calculated as $\lambda \cdot (x,y)$ and $\lambda \cdot(x',y')$ respectively, where $\lambda$ is the ratio of the width of the full feature map and that of the input image. 

In Step {\textcircled{\small{3}}}, we pool $Q_n$ and $Q^{'}_{n}$ to form the feature vectors $v$ and $v'$. We extract one feature value from each feature map, using the mean pooling. For example, we get a group of feature maps of the stop sign from one hidden layer with the size of 3*3*256, which means there are 256 convolution kernels in this layer. Usually, each kernel filters one kind of typical feature. Therefore, we get a 3*3 feature map from each filter in Step {\textcircled{\small{2}}} and then pool it to one single value. With the mean pooling, we get the vector with the size of 256. Then we normalize it and get the feature vector $v$ for the input image $x$ ( $v'$ for the input image $x'$ can be obtained in a similar way.).

In Step {\textcircled{\small{4}}}, we use the function $loss_{f}=\sum |v-v'|^{2}$ to measure the difference of features from the hidden layer.
Finally, our loss function is describe as follows:
\begin{equation}
\begin{split}
&  \alpha\cdot C_{N}^{box}+\beta\cdot p_{N}(y^{N}|S)+c\cdot (loss_f)^{-1}
\end{split}
\label{Equation 1}
\end{equation}
Given an input, an object detector makes several most possible predictions. Each prediction $y^N$ is represented in the form of $<C_{N}^{box}, p_N>$. $N$ is the index of the prediction, which outputs the result of the target object (e.g., the stop sign). $S$ denotes the class probabilities space. $C_{N}^{box}$ is the box confidence and $p_N$ is a vector of probability distribution over all classes. In order to hide the prediction of a particular object $y^N$, either $C_{N}^{box}$ or the particular object's probability in $p_N$ should be less than the thresholds that control whether the object can be detected. Furthermore, the difference of hidden layer features $loss_{f}$ between the target object (to be hidden) and the perturbed object should be maximized. Hence, our loss function of HA is defined as in Equation~\ref{Equation 1}, where $\alpha$, $\beta$ and $c$ are the parameters to adjust the weights\footnote{In the evaluation of this paper, $\alpha$, $\beta$ and $c$ are 1, 1 and 0.1 respectively.}. Based on our evaluation, FIR is demonstrated to be effective to enhance the robustness of AEs: it contributes 7\% in the increase of the distance and the angle.

\begin{figure*}

\centering
\epsfig{figure=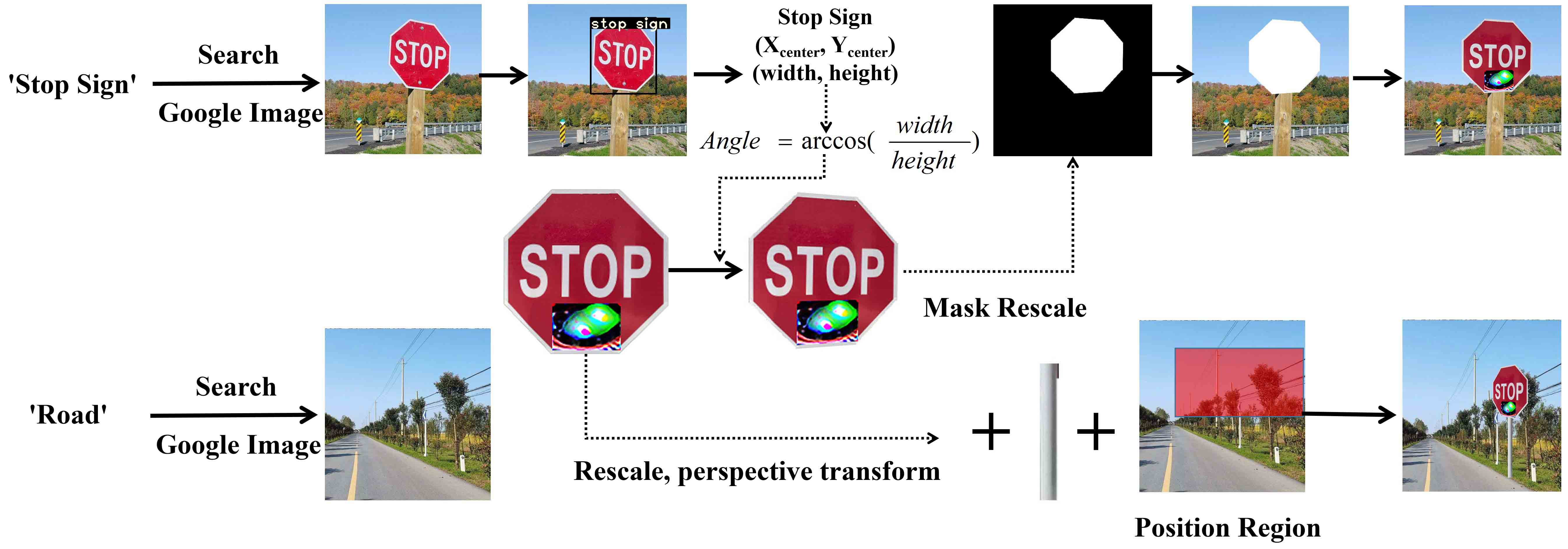, width=0.8\textwidth} 
\caption{\textbf{The Workflow of Enhanced Realistic Constraints Generation.}}
\label{Enhanced-RS method.}
\end{figure*}

To reduce the complexity of the object function, instead of all hidden layers, we select a few hidden layers to optimize the third item of Equation~\ref{Equation 1}. Take YOLO V3 as an example. YOLO V3 network architecture consists of the feature extraction part (the darknet backbone) and the detection part. In order to extract features of diverse scales, the hidden layers can be viewed as being divided into groups, with each group extracting feature maps of different sizes. Within each group however, the consecutive hidden layers extract the feature maps of the same size. Therefore, focusing on the feature extraction part of YOLO V3, we just choose the last hidden layer from each group, i.e., one hidden layer for each different size. For the optimization of the selected hidden layers, the former layers usually impact the result more than the latter layers, no matter the effect is positive or negative, since the gradients of the former layers are typically larger. However, giving a significant weight to the former layers may disturb the optimization of the other two items in Equation~\ref{Equation 1}. Therefore, we adjust the parameter $c$ accordingly to find an optimal solution.

\subsection{Enhanced Realistic Constraints Generation}
\label{enhanced image transformation}


In the prior studies, Expectation over Transformations (EOTs) are applied to build an adversarial attack in the physical world. EOT is to add random distortions in the optimization to make the perturbation more robust, but simulating realistic situations with random image transformations as in the existing works is not enough. The reason is that we observed that object detectors have a certain ``knowledge'' to the background of the object and the object semantic. This ``knowledge'' make detectors be sensitive to the relationship of the object and different background as well as the object semantic integrity. More precisely, the former means whether the object is in the reasonable environment and whether this object is in the reasonable position, while the latter means whether the object appears to be in reasonable integrity. Experimental results about such observation  are shown in Section~\cref{The analysis of factors}.

We can leverage these sensitivities to expose the perturbations that are not so robust in the optimization process and then optimize them to make them more robust. 
For instance, for a stop sign with adversarial perturbations, even if unrobust, the object detectors are still likely to be able to recognize it correctly if the background environment is outdoor and there is a pole equipped below the stop sign. 
Then we can optimize it based on the gradients in the backward propagation until it is robust enough.
Hence, ERG is proposed to generate more realistic constraints (reasonable background and reasonable object semantic integrity) in a systematic way. Figure ~\ref{Enhanced-RS method.} illustrates ERG method using the stop sign as an example. We find the necessary background using a search engine (e.g., Google) using two approaches. The first one is to search the target directly using the name (e.g., using the word ``stop sign''). Then the search engine will return the image of the target with a real background. The second approach is to use the words related to the semantic of the target (e.g., using the word ``road''). Then the search engine will return suitable images, but may be without the target. We detail how to leverage the background search using two different approaches.

For the background images containing the target (e.g., stop sign), we leverage the original background to generate AEs, which makes the generated image more realistic. Such AEs are shown to be more robust. The basic idea is to extract the target object, perform various transformations on the object and put it back to the original image to replace the original target. Detailed steps are as follows. We first utilize an object detector to get the coordinates of the target object in the image, the size of its bounding box and the $(width, height)$ of this bounding box. Then the approximate angle of the original target in the image can be estimated using the formula: $arccos(\frac{width}{height})$. Thirdly, we re-scale the perturbed target with the size of the bounding box and apply the perspective transformation method on it with the estimated angle. We also apply the random gray-scale transformations to simulate the illumination changes. Finally, we replace the original target with the transformed one in the image by cutting out the area of the original target in the image and adding the transformed one. The original target can be easily identified, since we have got the coordinates and (width, height) of it in the image with the object detector as mentioned above.

For the background images without the target, but with related semantics, we first choose a position region where the target is most likely to appear if it were in the image. 
Position region is set as the rectangle area mostly near or in the region of the semantic-related object in each image. Hence, we can get the region of the semantic-related object of each image with the image semantic segmentation tools and then set the position region of the target based on the results and semantics~\cite{ImageSegmentation}. 
As for the “stop sign”, the semantic-related object is the road and then the position region of the stop sign should be the rectangle area beside the road. After locating the position region, we apply different transformations to the target object, e.g., random size for re-scaling, random angle for perspective transformation and random gray-scale. Finally, we add the transformed target to the position region in the image for AE generation. This approach may find out some images unrelated to the target. Our idea is to double check the semantics of the searched image by Google, and to verify whether the image is related to the keyword in the searching.

\begin{figure}

\centering
\subfigure[Mesh grids in the short distance.]{
\begin{minipage}[t]{0.25\textwidth}
\epsfig{figure=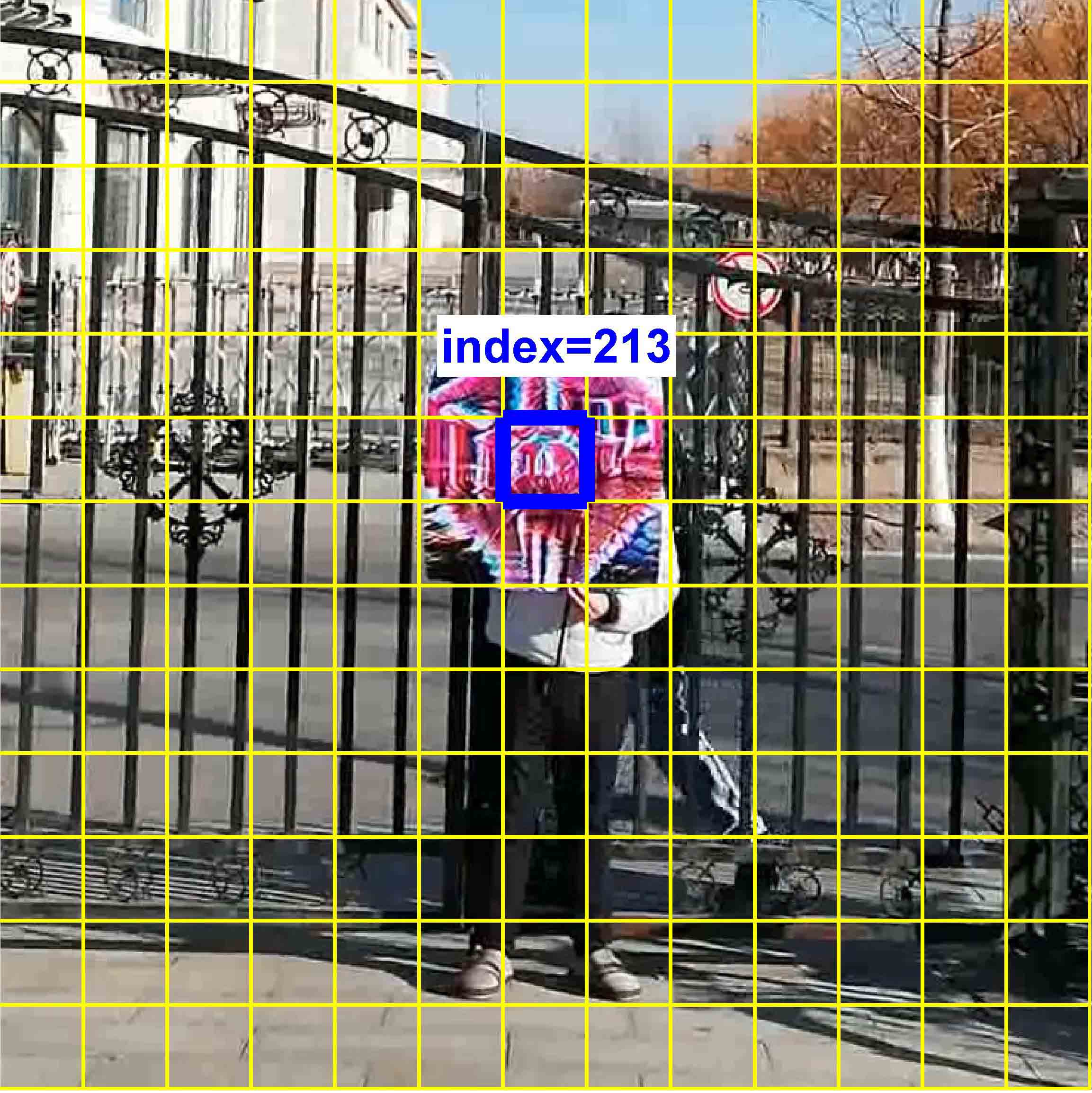, width=0.85\textwidth} 
\end{minipage}%
}%
\subfigure[Mesh grids in the long distance.]{
\begin{minipage}[t]{0.25\textwidth}
\epsfig{figure=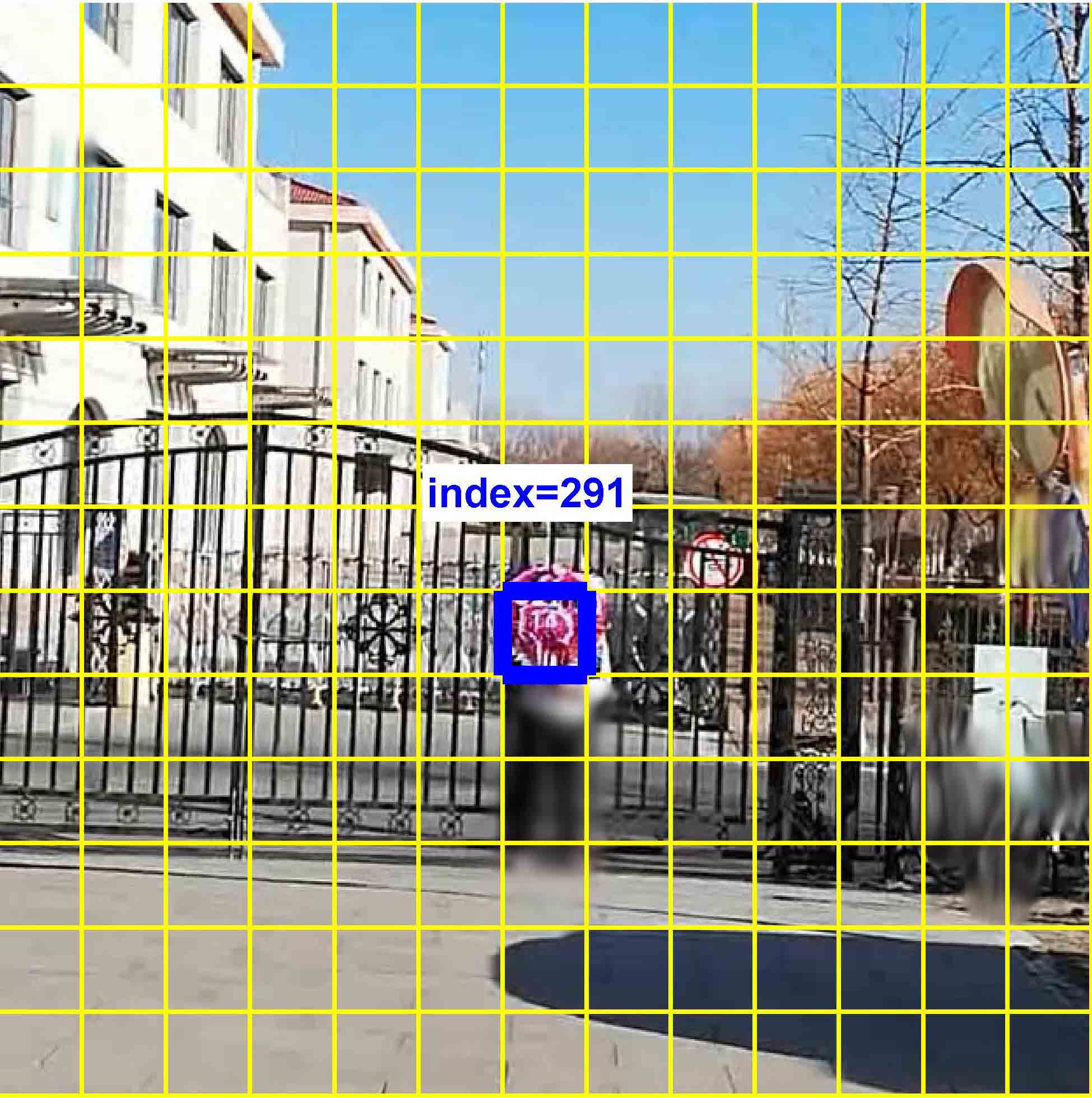, width=0.85\textwidth} 
\end{minipage}%
}%

\caption{\textbf{Mesh Grids in the Short Distance View and the Long Distance View.}}
\label{mesh fig.}
\end{figure}

Besides the realistic constraints generation, we also consider the color saturation constraint to overcome the printer chromatic aberration. As we know, printers are unable to reproduce the color of the original digital image accurately. So in the real world, the generated digital perturbations cannot be exactly the same way as they were supposed to be. Neither can camera lens be able to capture the color perfectly. Thus, the perturbations on the AEs captured by them also lose fidelity before feeding them to the object detectors. Such chromatic aberration also introduces difficulties when attacking the real world object detectors. Interestingly, we find that for the images with low saturation, color printers can usually reproduce them with less chromatic aberration. Therefore, we utilize color saturation function to impose restrictions on the perturbation, that is, for each pixel of the perturbation, we limit its color saturation to be lower than a threshold. In this way, the generated AEs can always be with low saturation, thus more suitable for printing.

\subsection{Nested AEs}
\label{nested AEs}

Inspired by the observation that an object ``looks'' smaller when captured in long distance and bigger in short distance, recent object detectors, such as YOLO v3, are designed to use more than one scales (e.g., three scales for YOLO v3: big, medium, and small scale) to measure the scope of the objects, thus improving the accuracy of object detection, especially the small objects or the objects in the long distance\footnote{This is also the reason why YOLO v3 performs much better than the previous versions (e.g., YOLO v1 and v2) on detecting small objects or objects in the long distance.}. An object is considered to be detected as long as the object detector model detects an object from any one of the three scales. 
However, compared to the partial model that identifies big or medium objects, the other part of the model detecting small objects (referred to as $Model_s$) is easier to be deceived since it relies on fewer pixels in the video frame (also with few features) to detect objects.

Based on the observation above, we always target $Model_s$ in different distances since it is easier to be deceived. In particular, considering the AE in long distance as shown in the right of Figure~\ref{mesh fig.}, the entire AE together only appears as a few number pixels in the video frame, to attack $Model_s$. In contrast, to attack $Model_s$ in short distance as shown in the left of Figure~\ref{mesh fig.}, only the central area of the AE takes effect. Such AE with the central area (for short distance attack) integrated into the whole area (for long distance attack) is named as \textit{nested AE} in this paper. Note that, the central area and the whole area of the nested AEs should not interfere with each other. The formal design of nested AEs is as below:

\begin{equation}
  X^{adv}_{i+1}=Clip\begin{Bmatrix}
X_i+\varepsilon sign(J(X_i)),\quad \quad \quad \quad S_p \leqslant S_{thres}\\  X_i+\varepsilon M_{center}sign(J(X_i)),\quad S_p>S_{thres}
\end{Bmatrix}
\end{equation}
where $X_i$ is the origin AE generated with random noise,  $X^{adv}_{i+1}$ denotes the modified AE, $J(\cdot)$ is the gradients of the input $X_i$, and $Clip(\cdot)$ normalizes all elements in inputs into the range of [0, 255]. If the size of the AE (referred to as $S_p$) is less than or equals to the threshold $S_{thres}$, we regard it as a long distance attack and modify the full AE. Otherwise, we view it as a short distance attack and only modify the center of the AE. Overall, decoupling the task of varying distance attack into two sub-tasks, long distance attack and short distance attack, enables robust AE generation in the scenarios of varying distance. With the help of nested AE, we achieved high attack success rate at the distance from 6$m$ to 25$m$ in our evaluation.

\vspace{5pt}\noindent \textbf{Loss Function based on Nested AEs.}
In order to implement the proposed nested AEs for Appearance Attack (AA), a loss function should be designed to increase the probability of the target and suppress the probabilities of other objects during the prediction process. Different  from image classifiers, object detectors need to identify all the recognizable objects in every single video frame. Thus, we should first locate the position where the AE appears, and then design the loss function based on this position. 

The object detectors divide each video frame into several different $m \times n$ grids based on the scales. Note that based on the design principle discussed above, we only target $Model_s$. So $m$ and $n$ are fixed in the model (e.g., $m=52$ and $n=52$ in YOLO v3). In Figure~\ref{mesh fig.}, we use a mesh grid of $m=13$ and $n=13$ as an example, to illustrate the position of AE. From the figure, we find that the AE is in the box with blue border. Then we could map the position to the prediction results (usually expressed by tensors). The index of the tensor is referred to as $N_p$, which can be calculated based on the size of AE $P_{size}$ and the center position of AE $P_{position}$. For instance, in the left of Figure~\ref{mesh fig.}, $N_p$ of the grid where the center position of the AE locates is $213$. Note that in different frames of the video, the position of the AE can change, e.g., in the right of Figure~\ref{mesh fig.}, so $N_p$ changes to $291$ accordingly. Hence, $N_p$ should be re-calculated for each video frame. 

Once $N_p$ is calculated, we define the loss function as follows:

\vspace{-8pt}
\begin{equation}
\begin{split}
& N_p=f(P_{size},P_{position}), \\
&     1-C_{N_p}^{box}+\beta \sum\left |p_{N_p,j}-y_j  \right |^2
\end{split}
\end{equation}

\noindent where $f(\cdot)$ is the function to calculate $N_p$. The loss function is composed of two parts. The first part is $1-C_{N_p}^{box}$, where $C_{N_p}^{box}$ is the box confidence of the prediction at the index $N_p$. 
The second part $\beta \sum\left |p_{N_p,j}-y_j \right |^2$ calculates the sum of the square of the differences between the probabilities of other predictions (denoted by $p_{N_p,j}$) and the target $y_j$ we set at the index $N_p$. Hence, minimizing the loss function will maximize the confidence of the target object at $N_p$, and meanwhile minimize the possibilities for other objects to be detected.

\subsection{Style-customized AEs}
\label{style-customized AEs}
The prior work ~\cite{song2018physical} customized the shape of the perturbations to mimic vandalism (e.g., graffiti on stop sign), to make them surreptitious. However, such unitary style may only work for limited scenarios. It is obvious that diverse styles provide more choices to make perturbations adapt to different attack environments, thus making them more surreptitious. In this paper, we introduce diverse styles to mimic the graffiti or advertisements, namely Pattern-controlled AE, shape-controlled AE, color-controlled AE, and text-based AE (via the combination of shape-controlled AE and color-controlled AE). Note that the style-customized AEs are designed specifically for HA in this paper, rather than AA. The AE of AA is a single poster to be individually placed somewhere reasonable, unlike the AE of HA, which is typically attached on the stop sign. The style-customized AE can be added in the AE generation as a nice-to-have option to make AE more surreptitious. In particular, shape-control can be added with a $Mask$ matrix when modifying the AE, while the pattern-control and color-control can be implemented through their loss functions.

\begin{figure}
\centering
\includegraphics[width=0.45\textwidth]{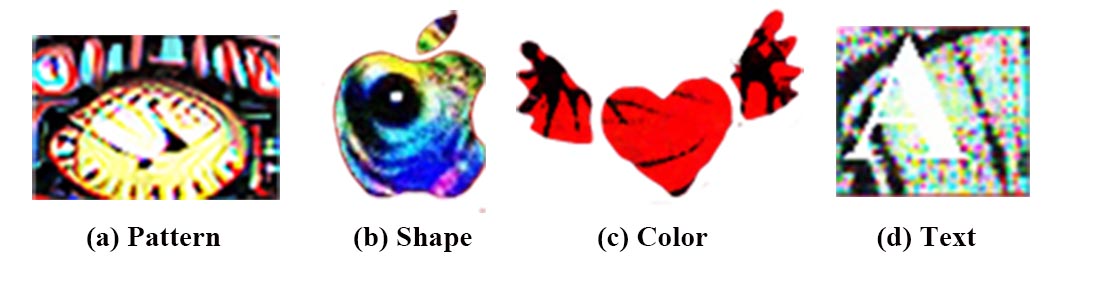} 
\caption{\textbf{The Style-customized AEs} 
\label{style-customized AEs figure}
} 
\end{figure}

\vspace{5pt}\noindent \textbf{Pattern-controlled AE.} Pattern control aims to generate AEs with specified patterns such as a clock, a person or even a car. Figure~\ref{style-customized AEs figure} (a) gives an example of AE with a clock inside. We can leverage the target object detector itself to generate the pattern in AE, because the object detectors are trained using images of objects, so they should contain the information of the objects that it has learned in the training process. To implement the idea of recovering the object from the model, we design a pattern control loss function, written as: 
$\sum\left | p_{j}-1 \right |^{2}$
, where $j$ is the target class that we desire the pattern to be, and $p_j$ is the probability of class $j$ in each prediction we aim to modify. Minimizing the loss function will improve the probability of the target class, so the adversarial sample will be crafted to be similar with the target class from the model’s perspective.

\vspace{5pt}\noindent \textbf{Shape-controlled AE.} Shape control makes AE in a specific shape, like a butterfly or the shape of love. Figure~\ref{style-customized AEs figure} (b) shows an example of AE in the shape of Apple logo. In order to develop the irregular shape, we create a mask to control the shape of AE. Assume $X^{adv}+P$ is the function to modify the adversarial samples where $P$ is the perturbation. We slightly change it to be $X^{adv}+P \cdot Mask$ where $Mask$ is a matrix with the same dimension as $X^{adv}$, with all values either 0 or 1. Hence all the elements with the value of 1 form the shape as desired, and are allowed to be modified in the training process. In this way, the generated AE will be in the specific shape.

\vspace{5pt}\noindent \textbf{Color-controlled AE.} We can control the color used to build AEs. Figure~\ref{style-customized AEs figure} (c) gives an example of AE with red hue. To generate such style, the original AE needs to be colorful. Then we can adjust the color of AE as needed through a loss function, defined as below:
\begin{equation}
loss_{color}=\sum_{pixel\epsilon X_i}\frac{pixel_R+pixel_G+pixel_B }{pixel_T}
\end{equation}
where $Pixel_R$, $Pixel_G$, $Pixel_B$ are R (red), G (green) and B (blue) values of each pixel and $Pixel_T$ is the value of the target color. Since each pixel of the image contains RGB with different weights, we can tune the weights of each pixel to generate an AE with a primary color, e.g., assigning more weights on R than G and B in the AE generation. In this way, the color hue can be controlled. 

\vspace{5pt}\noindent \textbf{Text-based AE.} We could further generate texts in AEs to mimic small advertisements. Figure~\ref{style-customized AEs figure} (d) gives an example of AE with letter ``A'' inside. We combine shape-controlled AE and color-controlled AE to implement the Text-based AE. Specifically, we create a mask with letter-like shape (e.g., ``A'', ``B'', ``C'') and control the color used in the shape (i.e., tuning RGB in each pixel of the shape). For example, we can use gray and white color in the shape, which are more appropriate to display texts.

\subsection{AE Generation with Various Constraints}
\label{sec:batch-variation momentum}
Based on the loss functions, AEs can be generated by iteratively modifying the perturbations with a small step $\varepsilon$ on the direction of gradient calculated. However, the number of constraints we introduced so far may let the generation process converge very slowly or even not converge. To solve the convergence problem, we adopt the \textit{batch-variation} method, which computes the average gradient of all $N$ gradients. Each gradient is computed based on the AE transformed through a specific variation, including the enhanced realistic constraints, different re-scale parameters, perspective transformation angles, etc., to guide the modification of AE. Using the average gradient instead of $N$ variations can stabilize the update directions and reduce the overfitting to limited realistic constraints.  Reducing overfitting further decreases AEs' dependence on the model which is helpful to increase the transferability.
Hence, the batch-variation method increases the convergence speed or improves the transferability of physical AEs to some extent. Based on evaluation results, such an approach to generate AEs is 5$\times$ faster than previous approaches to converge.

\begin{table}
\centering
\footnotesize
\caption{The List of Objects attacked by HA and AA}
\label{Attack Objects}
\begin{tabular}{m{2cm}
<{\centering}|m{4cm}
<{\centering}}
\hline
\textbf{Attack}& \textbf{Objects}\\
\hline
HA & stop sign, car, monitor\\
\hline
AA & stop sign, person, traffic light\\
\hline
\end{tabular}
\label{multiple-objects}
\end{table}

\section{Evaluation} 
\label{sec:evaluation}
We implemented HA and AA for multiple objects, including stop sign, car and monitor in HA, and stop sign, person and traffic light in AA, as shown in ~\reftable{multiple-objects}. 
Due to the space limit, we cannot present evaluation results for all the objects in different physical conditions. To ease the comparison with existing works, we choose the stop sign as an example in this section to elaborate the evaluation results of both HA and AA in various physical conditions, since the other two state-of-the-art physical attacks against detectors also evaluated theirs approaches using the stop sign\footnote{Stop sign is frequently used to evaluate the physical adversarial attack, because it is considered to be highly related to the traffic safety (Practical adversarial attack against stop sign can cause autonomous driving cars to malfunction.).}. We recorded the attacks against other objects and uploaded them on the demo website. 

\subsection{Experimental Setup}

We evaluated AEs in three different kinds of environment settings, indoor (lab) environment, outdoor environment and the real road. 
We purchased a real stop sign, as shown in Figure~\ref{effectiveness against attack.}, for all the related experiments. In HA, the generated AEs are printed using a regular desktop printer, HP Color LaserJet Pro MFP M277dw. Then we cut the stickers out of the printout and attach them to the surface of the stop sign. For AA, we print the generated AE as a 60$cm\times$60$cm$ poster to represent as our AE. We evaluate the effectiveness of AEs in the physical space by shooting videos of the AEs, and running object detectors on the video recordings. The cameras used to shoot video are the built-in cameras of iPhone 6s and HUAWEI nova 3e. The computer used to generate AEs is equipped with an Intel Xeon E5-2620 CPU, a GTX Titan-X GPU and 32GB physical memory.

We evaluated the AEs using YOLO V3 and Faster RCNN, which are the representative models of one-stage detectors and two-stage detectors accordingly. The backbones of the pre-trained models YOLO V3 and Faster RCNN are Darknet-53 and ResNet-101 respectively. Both the two detectors are trained on Common Objects in Context (COCO) dataset~\cite{homeCOCO}~\cite{MSCOCO}. We define the success rate of the physical attack as $f_{succ}=N_{succ}/N_{all}\times 100$, where $N_{all}$ denotes the number of all the frames in a video and $N_{succ}$ denotes the number of the frames in which our attack successfully fools the object detector.


\begin{figure*}

\centering
\subfigure{
\begin{minipage}[t]{0.25\linewidth}
\centering
\epsfig{figure=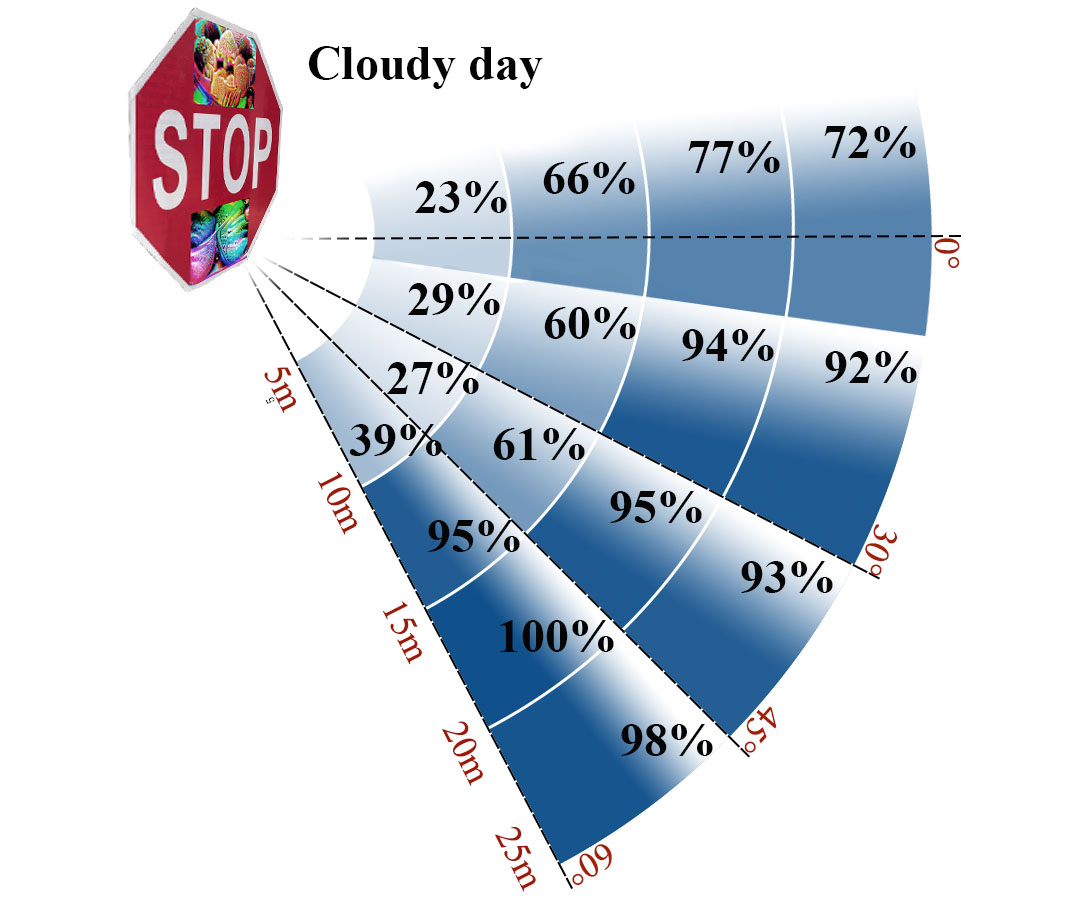, width=1.1\textwidth} 
\end{minipage}%

\begin{minipage}[t]{0.25\linewidth}
\centering
\epsfig{figure=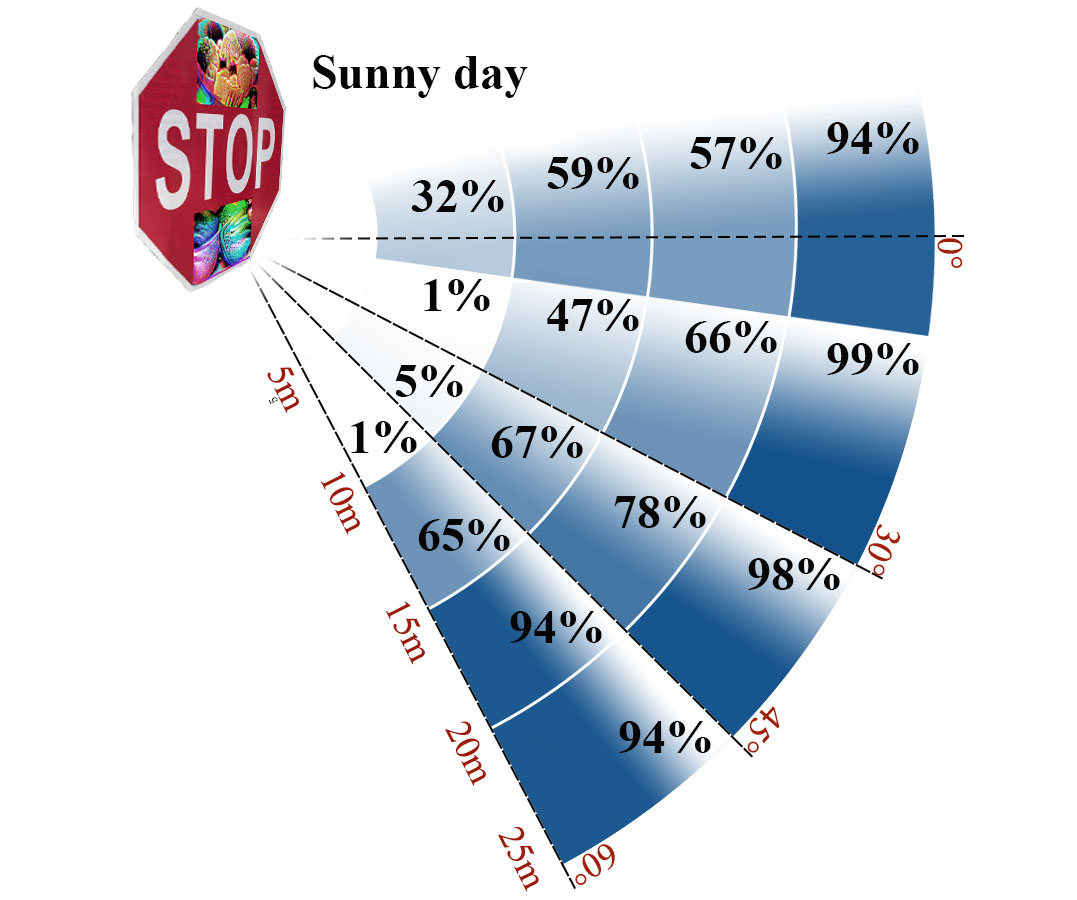, width=1.1\textwidth} 
\end{minipage}%

\begin{minipage}[t]{0.25\linewidth}
\centering
\epsfig{figure=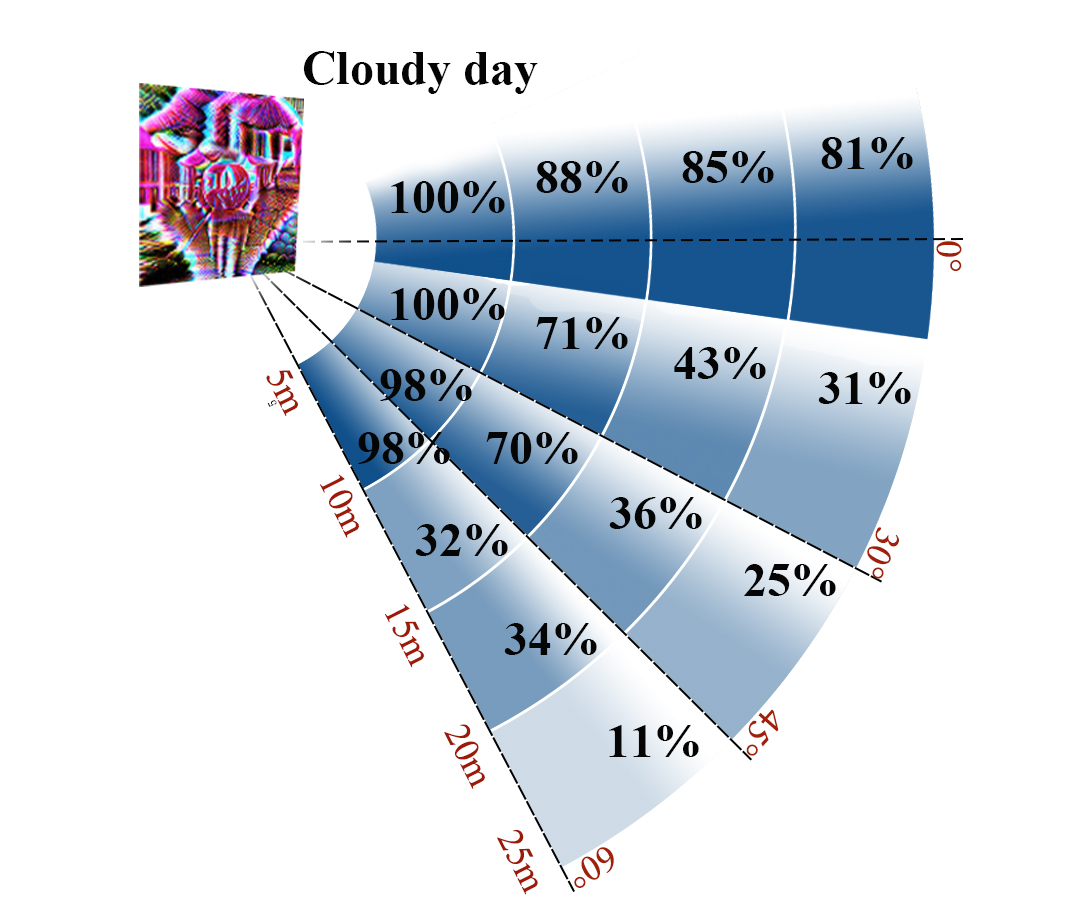, width=1.1\textwidth} 
\end{minipage}%

\begin{minipage}[t]{0.25\linewidth}
\centering
\epsfig{figure=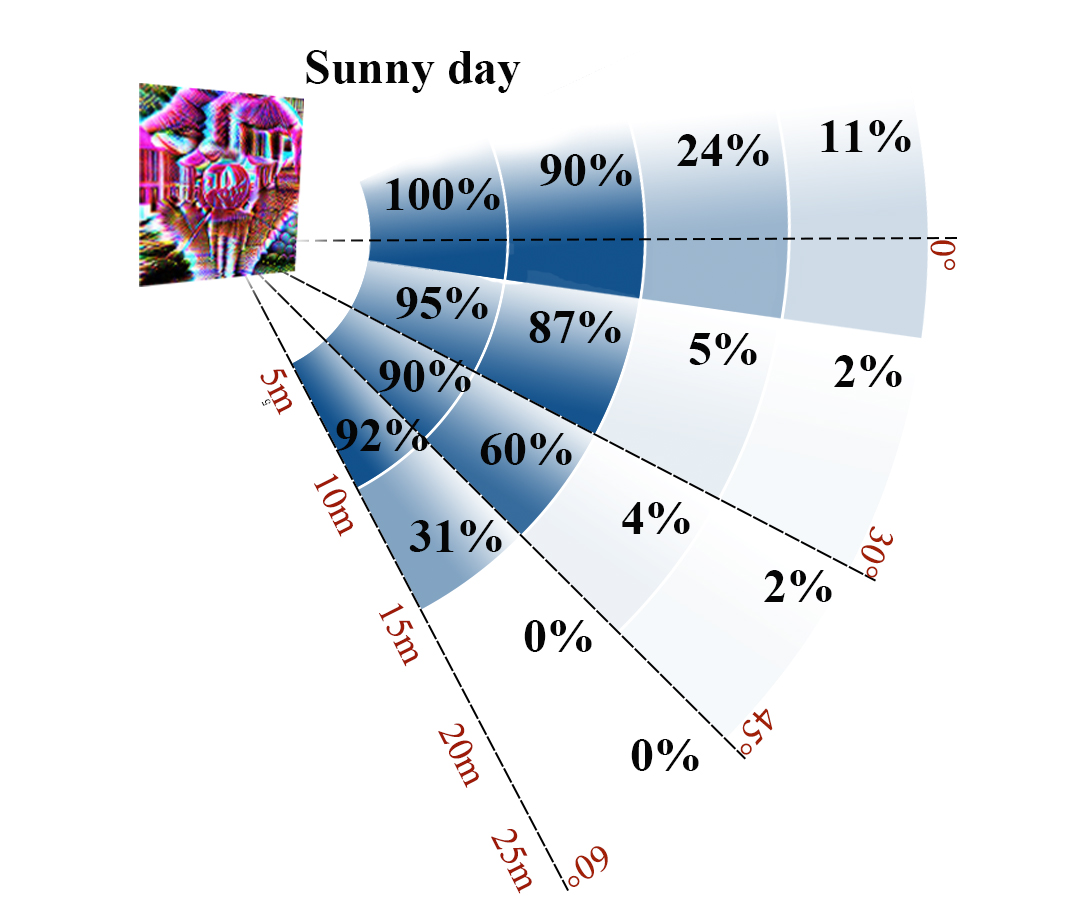, width=1.1\textwidth} 
\end{minipage}%

}%

\caption{\textbf{Success Rate at Different Angles, Distances, and Illuminations. (Left two figures: Hiding Attack; Right two figures: Appearing Attack)  }}
\label{varyingconditions}
\end{figure*}

\subsection{Effectiveness}
\label{success rate of HA and AA}
We evaluated the AEs of HA and AA (generated based on YOLO V3) against various factors including distance, angle, and illumination. We recorded several pieces of video towards the AEs (the stop sign) using an iPhone 6s and a HUAWEI nova 3e. To examine the impact of varying distances and angles, we divided the distances 5$m\sim$25$m$ into five regions (each region is 5$m$), and recorded video in each region (keep moving from $Nm$ to $N+5m$) at the angles $0^\circ$, $30^\circ$, $45^\circ$, and $60^\circ$ respectively. To evaluate the impact of illumination, we repeated the above experiments under different illumination conditions, e.g., at the same time from 1:00 pm to 3:00 pm on a sunny day and a cloudy day respectively. Figure~\ref{varyingconditions} shows the success rate (marked in each corresponding region) of HA and AA on a cloudy day and a sunny day at varying distances and angles. The depth of the background color is used to represent different success rates, i.e., the darker the color, the higher the success rate.

\vspace{2mm}
\noindent\textbf{AEs of HA.} The left two of Figure \ref{varyingconditions} demonstrate the success rate of HA on a cloudy day and a sunny day, respectively. Generally, a higher success rate is achieved at wide angles than narrow angles, and at long distances than short distances. For example, the average success rate for all the four angles is 89$\%$ in the range of 20$m\sim$25$m$, which is larger than the average success rate of 70$\%$ in the range of 10$m\sim$15$m$. The average success rate over the entire distance range (5$m\sim$25$m$) at angle $60^\circ$ is 83$\%$, also larger than 60$\%$ at angle $0^\circ$. HA at the wide angle performs as good as or even better than at the narrow angle. We repeated the same experiments on Yolo v3 with the original stop sign, and found that the capability of detecting the stop sign of YOLO v3 does not decline significantly as the angle increases, and even remain the same when close enough (The success rates are shown in Table~\ref{Success rate of the real stop sign on YOLO V3} in Appendix.). Hence, the better performance of HA at wide angle largely results from our approach, rather than the weaker capability of YOLO v3 at wide angle. Furthermore, we observed that the success rate of HA drops gradually as the distance gets shorter. The reason is the detection capability of the detector increases as the distance becomes closer, thus more difficult to be deceived.

\vspace{2mm}
\noindent\textbf{AEs of AA.} 
The right two of Figure \ref{varyingconditions} demonstrate the success rate of AA on a cloudy day and a sunny day respectively. On the cloudy day, AA achieves over 98$\%$ success rate at all angles within 5$m\sim$10$m$ and over 70$\%$ success rate at $0^\circ\sim45^\circ$ within 10$m\sim$15$m$. Moreover, it keeps a high success rate over 80$\%$ at $0^\circ$ from 0$m$ to 25$m$. In Figure~\ref{varyingconditions}, we can see that apparently the success rate of AA is opposite to that of HA, indicated by the changes of the color. In particular, AA performs better at the short distance and narrow angle, while HA better at the long distance and wide angle. This is because wide angle and long distance will affect the object's features (crafted by AA) captured by the detector. On the sunny day, AA performs great at the distance 5$m\sim$15$m$ and the angle $0^\circ\sim45^\circ$, because the printed AEs can be recorded much more clear on the sunny day, which helps highlight the features of the AEs. However, the success rate degrades rapidly as the distance over 15$m$, since the reflection will impact the success rate of AEs with long distance.  

Overall, AEs of HA demonstrate great robustness against different angles and illumination conditions. Meanwhile, with the distance longer than 10$m$ (For HA, it can be too late to stop when the object detector on autonomous driving cars recognizes the stop sign as close as 10$m$.), AEs of HA can always achieve good success rate. AEs of AA is shown to be robust towards angles up to $60^\circ$ within the distance of 10$m$ (  For AA, the control system of the autonomous driving car may give a decision to immediately slow down or brake when the perception system (e.g., the object detector) detects a stop sign as close as 10$m$.).

\vspace{2mm}
\noindent\textbf{Comparison with the State-of-the-art Attacks.}
We compared our work with two state-of-the-art attacks against the object detectors:  ShapeShifter \cite{DBLP:journals/corr/abs-1804-05810} and Eykholt's method \cite{song2018physical}, and evaluated the improvements introduced by our proposed approaches.

\begin{table*}
\centering
\footnotesize
\caption{Comparison with the State-of-the-Art Attacks
}
\label{Comparison with state-of-the-art attacks}
\begin{tabular}{m{2cm}
<{\centering}|m{2cm}
<{\centering}|m{2cm}
<{\centering}|m{3cm}
<{\centering}|m{4cm}
<{\centering}}
\hline
\textbf{HA}& \textbf{Distance}& \textbf{Angle}& \textbf{Perturbation Area}& \textbf{Transferability}\\
\hline \hline
\textbf{Our method} & \text{$\leqslant 25m$}& \text{$\leqslant 60^\circ$} & \text{$20\% \sim 25\%$}& \text{Faster, YOLO, SSD, RFCN, Mask}  \\ \hline
\textbf{ShapeShifter}& \text{$\leqslant 40' \  (12m)$} &\text{$\leqslant 15^\circ $} & \text{Full stop image except "STOP"}&\text{Unable}  \\ \hline
\textbf{Eykholt's method}& \text{$\leqslant 30' \  (9m)$} &\text{$---$} & \text{$20\% \sim 25\%^*$}&\text{Faster RCNN (18$\%$)}  \\ \hline
\end{tabular}
\vspace{6pt}

\begin{tabular}{m{2cm}
<{\centering}|m{2cm}
<{\centering}|m{2cm}
<{\centering}|m{3cm}
<{\centering}|m{4cm}
<{\centering}}
\hline
\textbf{AA}& \textbf{Distance}& \textbf{Angle}& \textbf{Perturbation Area}& \textbf{Transferability}\\
\hline \hline
\textbf{Our method} & \text{$\leqslant 25m$}& \text{$\leqslant 60^\circ$} & \text{$---$}&\text{Faster, YOLO, SSD, RFCN, Mask}\\ \hline
\textbf{ShapeShifter}& \text{$---$} &\text{$---$} & \text{$---$}&\text{$---$}\\ \hline
\textbf{Eykholt's method}& \text{$\leqslant 10' \  (3m)$} &\text{$---$} &\text{$---$}& \text{$---$} \\ \hline
\end{tabular}

\begin{tablenotes}
\footnotesize
\item[1] \scriptsize { \textbf{\quad \quad \quad \quad \quad *$20\% \sim 25\%$}: We measured the perturbation area ratio  based on the image of AE in paper \cite{song2018physical} . \textbf{*$---$}: We did not get the data from their papers. } 
\end{tablenotes}

\end{table*}

Table \ref{Comparison with state-of-the-art attacks} shows the improvements of our work over the other two state-of-the-art attacks. Overall, our AEs of both AA and HA get the longest attacking distance up to 25$m$ and the widest attacking angle up to $60^\circ$. ShapeShifter evaluated their AEs at a variety of distances (5’ to 40’) by taking photos indoors. 
Their AEs' largest effective angle is $15^\circ$ at the distance within 20'($6m$) and $30^\circ$  at the distance within 10' ($3m$), lack of transferability based on their evaluation. Eykholt et. al. evaluated the disappearance attack within 30'($9m$) and creation attack within 10' ($3m$) both at the angle $0^\circ$, without measuring their attacks at other angles. Moreover, their disappearance attack with the form of sticker could transfer to Faster RCNN with the success rate of 18.9$\%$. Our AEs of HA and AA generated based on YOLO V3 or Faster RCNN can transfer to other black-box models including SSD, RFCN and Mask RCNN with the success rates up to 90$\%$ and 72$\%$ in the indoor environment and the outdoor environment respectively. Details of evaluation results on transferability of our AEs are in Section \cref{Transferbility}.

\begin{figure}
\centering
\includegraphics[width=0.35\textwidth]{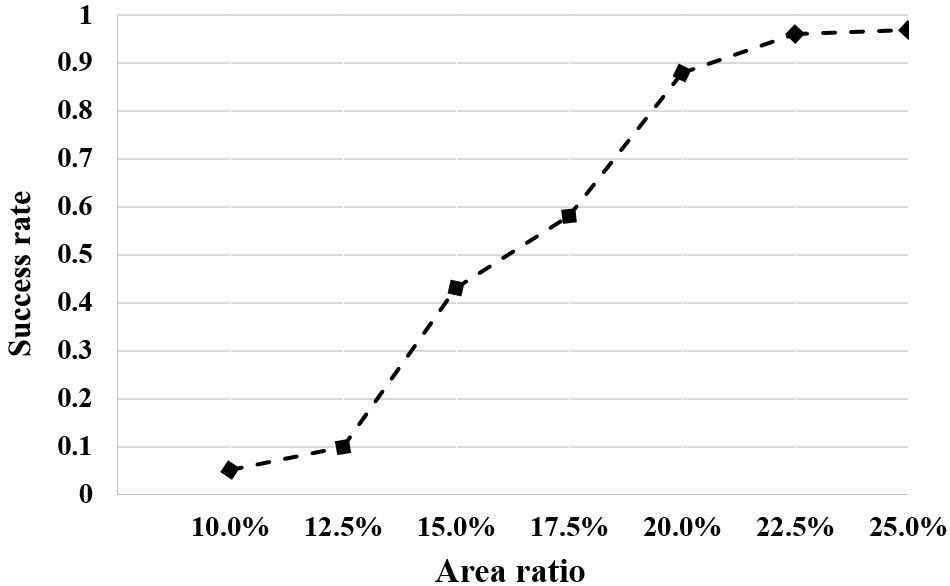} 
\caption{\textbf{The Success Ratio with different Area Ratios (i.e., the Area of the AE to That of the Target Object) 
}
}
\label{effect of area}
\end{figure}

\vspace{2mm}
\noindent\textbf{The Perturbation Area.}
Intuitively, the ratio of the area of the AE to that of the target object also affects the success rate of HA. The success rate of AE increases as its area gets larger. ShapeShifter modified the full stop image except the ``stop'' characters. Our method and Eykholt's method only modified a relatively small region of the stop sign as shown in Table \ref{Comparison with state-of-the-art attacks}.

To get an ``ideal'' perturbation area (by ``ideal'', we mean just large enough to provide good success rate, while remaining less noticeable.), we generated AEs with different areas and tested them in the digital space. The image of the stop sign partially covered by AEs with different areas was applied with random image transformations and added with random backgrounds. Then the success rate of each AE is computed based on the average of 10,000 tests. Figure~\ref{effect of area} shows the success rate varying the ratio. We can observe that the success rate reaches up to 90$\%$ when the ratio is around 20$\%$, and almost keeps stable (around 97$\%$) as the ratio is around 25$\%$. Hence, we conclude that the ``ideal'' ratio of the AEs in this paper should be in the range from 20$\%$ to 25$\%$, which is used in all the experiment in this paper.

\subsection{Performance Improvement of Individual Techniques}

\vspace{2mm}
\noindent\textbf{FIR.}
We demonstrate the performance improvement introduced by FIR by comparing the success rate of AEs generated with and without the reinforcement. For both with and without the reinforcement, AEs are always generated with enhanced realistic constraints generation from YOLO V3. We did not repeat all the experiments in Section \cref{success rate of HA and AA}. Instead, we set the angle at $0^\circ$ and distance ranging from 5$m$ to 25$m$, to evaluate the success rate of AEs. The results showed that FIR improves the average success rate of AEs from 53$\%$ to 60$\%$ for YOLO V3, which indicates a great improvement of robustness against varying distances. Experimental results also demonstrate steady improvement over varying angles.

\vspace{2mm}
\noindent\textbf{ERG.}
We demonstrate the performance improvement introduced by ERG by comparing the success rate of AEs generated with and without the enhancement. For the latter (without enhancement), we generated AEs by applying the transformations and background randomly. We obtained AEs from both YOLO V3 and Faster RCNN, with and without the enhancement. We did not repeat all the experiments in Section \cref{success rate of HA and AA}. Instead, we set the angel at $0^\circ$ and distance ranging from 5$m$ to 25$m$, to evaluate the success rate of AEs. 
The results showed that enhanced realistic constraints generation improves the average success rate of AEs from 31 $\%$ to 53 $\%$ for YOLO V3, and 43 $\%$ to 67 $\%$ for Faster RCNN, which indicates significant improvement of robustness against varying distances. Experimental results also demonstrate steady improvement over varying angles.

\begin{figure}

\centering
\subfigure[]{
\begin{minipage}[t]{0.25\textwidth}
\epsfig{figure=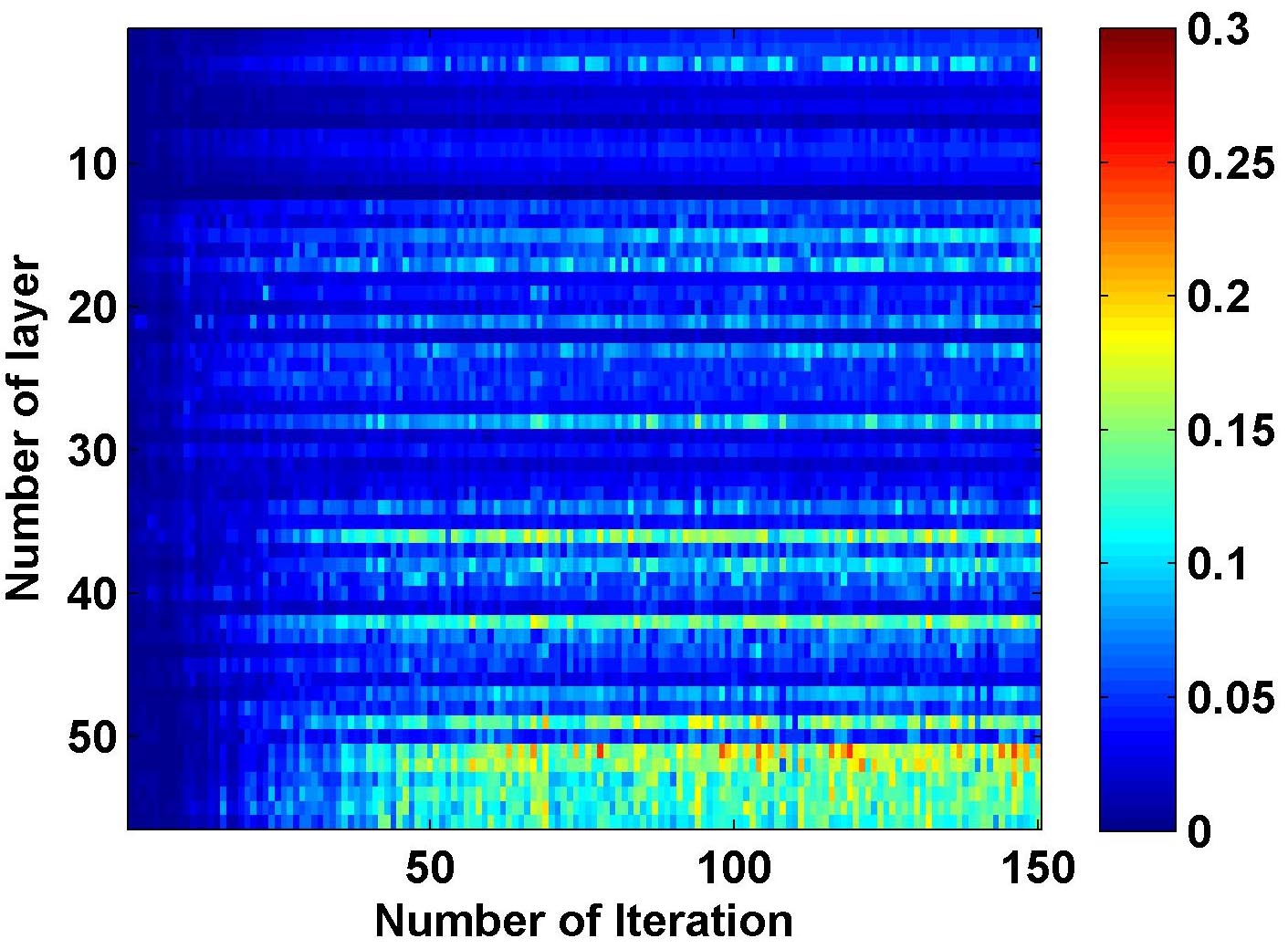, width=0.85\textwidth} 
\end{minipage}%
}%
\subfigure[]{
\begin{minipage}[t]{0.25\textwidth}
\epsfig{figure=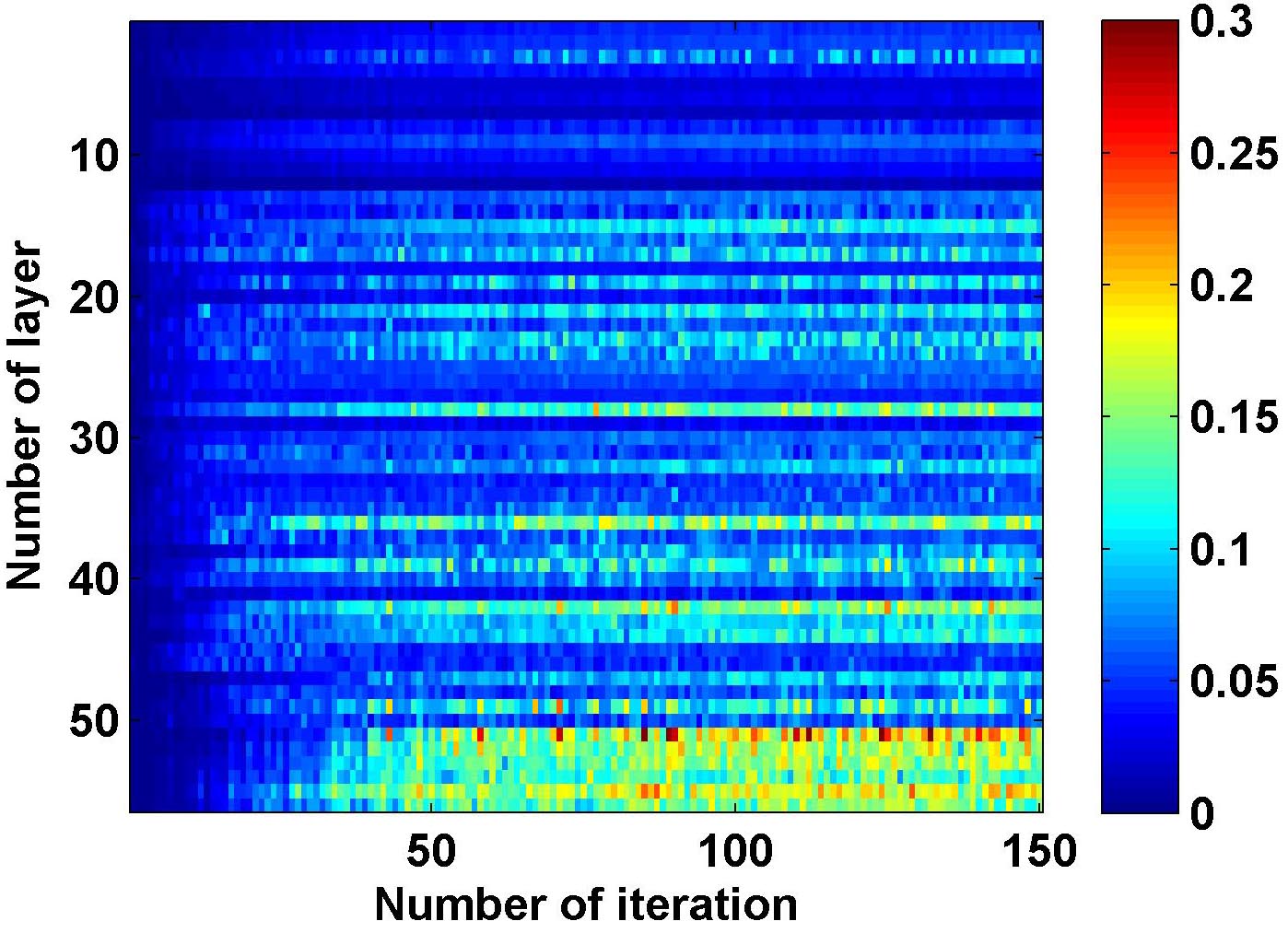, width=0.85\textwidth} 
\end{minipage}%

}%

\centering
\subfigure[]{
\begin{minipage}[t]{0.25\textwidth}
\epsfig{figure=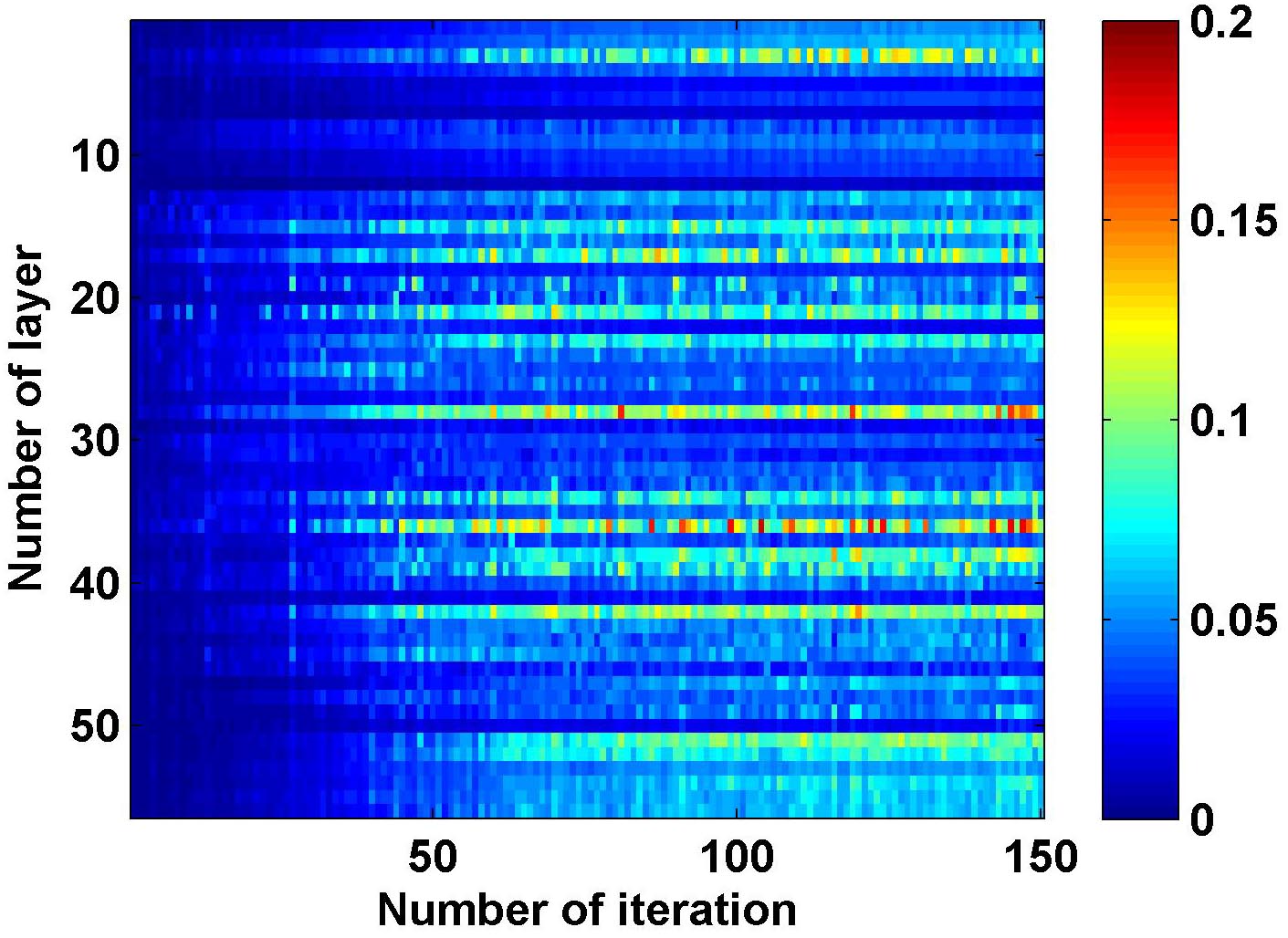, width=0.85\textwidth} 
\end{minipage}%
}%
\subfigure[]{
\begin{minipage}[t]{0.25\textwidth}
\epsfig{figure=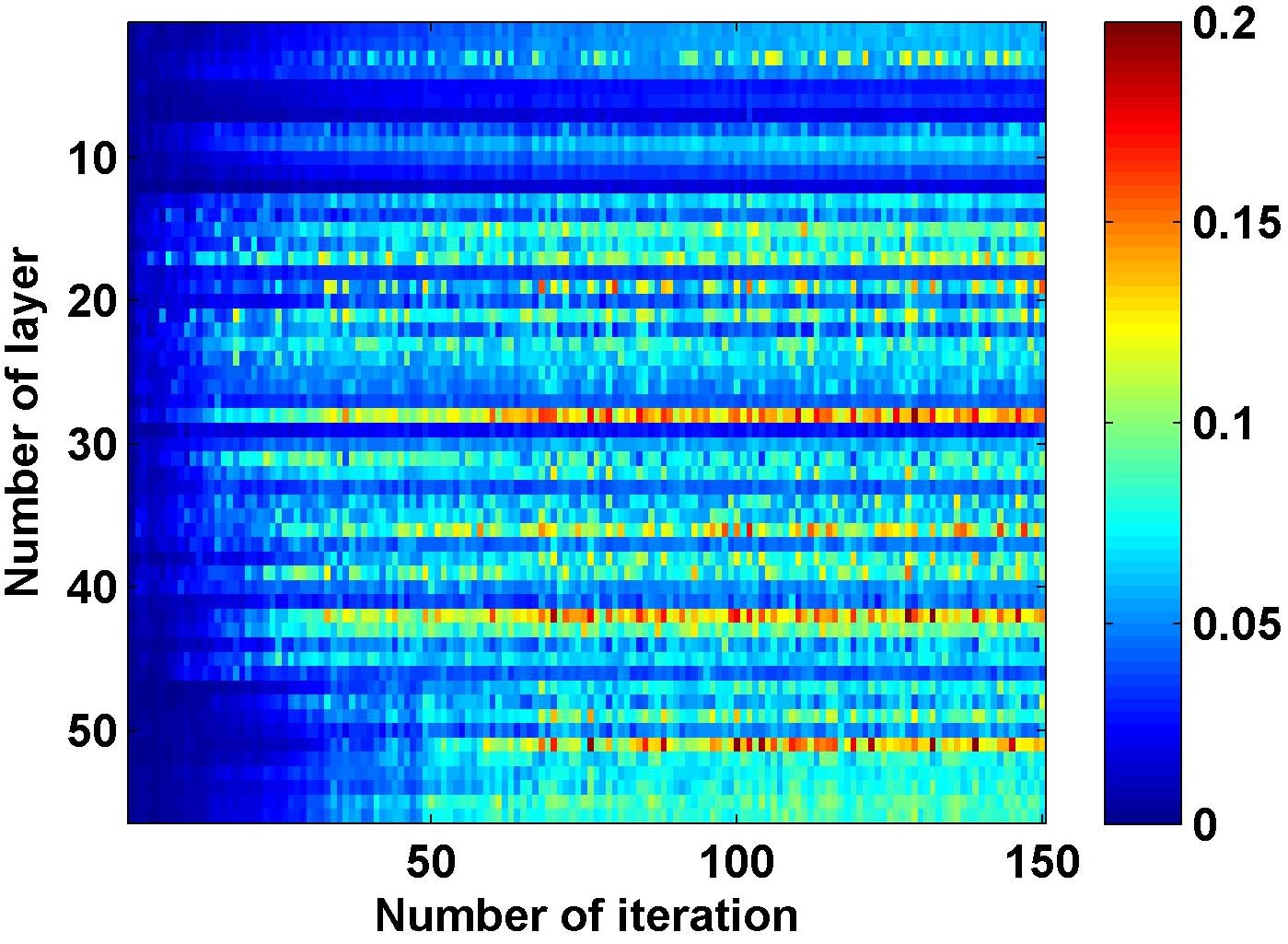, width=0.85\textwidth}
\end{minipage}%

}%
\caption{\textbf{Interference values for different hidden layers of DNN when attacked by AEs. (a) without FIR/ERG; (b) with FIR; (c) with ERG; (d) with FIR/ERG.}}
\vspace{-2mm}
\label{interference fig.}
\vspace{-2mm}
\end{figure}


\begin{table*}
\centering
\footnotesize
\caption{Transferability of AEs }
\label{Transferability}
\begin{tabular}{m{2.5cm}
<{\centering}|m{2.5cm}
<{\centering}|m{2.5cm}
<{\centering}|m{2.5cm}
<{\centering}|m{2.5cm}
<{\centering}|m{2.5cm}
<{\centering}}
\hline
\multicolumn{2}{c|}{\multirow{2}{*}{\textbf{White-box Model}}} & \multicolumn{4}{c}{\textbf{Black-box Model}} \\ \cline{3-6}
\multicolumn{2}{c|}{\text{}}&{\tabincell{c}{\textbf{Faster RCNN} \\ \textbf{/YOLO V3}}}&\textbf{SSD}&\textbf{RFCN}&\textbf{Mask RCNN}\\
\hline \hline
\multirow{2}{*}{\tabincell{c}{\textbf{YOLO V3} \\ \textbf{(Hidding Attack)}}} & \textbf{Indoors (785)*}& \text{21 $\%$}&
\text{71.6 $\%$}&
\text{52.6 $\%$}&
\text{49.7 $\%$}\\ \cline{2-6}
\textbf{}& \textbf{Outdoors (658)} &\cellcolor{gray!30}\text{10 $\%$}
&\cellcolor{gray!30}\text{46 $\%$}
&\cellcolor{gray!30}\text{19.2 $\%$}
&\cellcolor{gray!30}\text{9 $\%$}\\ 
\hline
\multirow{2}{*}{\tabincell{c}{\textbf{YOLO V3} \\ \textbf{(Appearing Attack)}}} & \textbf{Indoors(919)}& \text{ 51.8$\%$}&
\text{ 0$\%$}&
\text{ 20$\%$}&
\text{ 2$\%$}\\ \cline{2-6}
\textbf{}& \textbf{Outdoors (889)} &\cellcolor{gray!30}\text{48.2 $\%$}
&\cellcolor{gray!30}\text{8.4 $\%$}
&\cellcolor{gray!30}\text{47.4 $\%$}
&\cellcolor{gray!30}\text{56.2 $\%$}\\ 
\hline
\multirow{2}{*}{\tabincell{c}{\textbf{Faster RCNN} \\ \textbf{(Hidding Attack)}}} & \textbf{Indoors (701)}& \text{98.7 $\%$}&
\text{90.7 $\%$}&
\text{91 $\%$}&
\text{85.7 $\%$}\\ \cline{2-6}
\textbf{}& \textbf{Outdoors (839)} &\cellcolor{gray!30}\text{76.8 $\%$}
&\cellcolor{gray!30}\text{78 $\%$}
&\cellcolor{gray!30}\text{72 $\%$}
&\cellcolor{gray!30}\text{58 $\%$}\\ 
 \hline

\end{tabular}
\begin{tablenotes}
\item[1] \scriptsize {\textbf{*$(Num)$}: Num is the number of all frames of the test video. 
}
\end{tablenotes}
\end{table*}


\vspace{2mm}
\noindent\textbf{Further understanding by analyzing DNN.}
Besides understanding the performance improvement of FIR and ERG from the viewpoint of success rate, we further try to understand their impact on DNN itself. Specifically, we analyze the impact of perturbation on hidden convolutional layers to make sure that the AEs interfere with the features of the original target in the hidden layer. The more interference, the more robustness of AEs could be obtained.
To perform such analysis, we analyzed all 56 convolutional layers of YOLO V3 before the first residual layer, and measured the inference values variation to the hidden layer features caused by perturbations in 150 iterations of adversarial modification. The interference value for a hidden layer is defined as $loss_{f} / N$. $loss_{f}$ measures the difference of features in the layer (see ~\cref{feature-erosion}), and $N$ is the number of convolution kernels in the layer.

Figure~\ref{interference fig.} (a) shows the interference values for the DNN when attacked by the AEs generated without FIR and EDG. The x-axis shows the number of iterations and y-axis shows different hidden layers. Each value in the figure shows the interference value at a specific layer when attacked by an AE generated in a specific number of iterations. The red value means a large interference value which shows the impact of AE is high; while a blue value shows a small interference value. From this figure, we can see that the latter layers (i.e., the layer close to the output layer) are impacted more than the previous layers (i.e., the layer close to the input layer). With the number of iterations increases, more hidden layers are impacted.  Figure~\ref{interference fig.} (b) and (c) shows the interference values for the DNN when attacked by the AEs generated with FIR and EDG, respectively. Figure~\ref{interference fig.} (d) shows the interference values when both FIR and EDG are involved in the generation of AEs. By comparing the three figures with (a), we can see that for the same number of iterations, more hidden layers are impacted. This might explain the high robustness of the AEs generated through FIR and ERG.

\begin{table*}[h]
\centering
\footnotesize
\caption{Success Rate of the Style-customized AEs }
\label{Success rate of practical attack on YOLO V3}
\begin{tabular}{m{1.5cm}
<{\centering}|m{1.5cm}
<{\centering}|m{1.5cm}
<{\centering}|m{1.5cm}
<{\centering}|m{1.5cm}
<{\centering}|m{1.5cm}
<{\centering}|m{1.5cm}
<{\centering}|m{2.3cm}
<{\centering}}
\hline
\multicolumn{3}{c|}{\multirow{2}{*}{\textbf{Style-customized AEs }}} & \multicolumn{4}{c|}{\textbf{Hiding attack}} & \multirow{2}{*}{\textbf{Appearing attack}}\\ \cline{4-7}
\multicolumn{3}{c|}{\text{}}&\textbf{Pattern}&\textbf{Shape}&\textbf{Text}&\textbf{Color}&\text{}\\
\hline \hline

\multirow{4}{*}{\textbf{YOLO V3}} & \multirow{2}{*}{\textbf{Indoors}} & \text{$f_{succ}$}& \text{92.4 $\%$ (784)}& \text{99.5 $\%$ (803)} & \text{99.4 $\%$ (716)}& \text{97 $\%$ (986)}& \text{88 $\%$ (1894)}
\\ \cline{3-8}
\text{}& \text{}& \text{$max_{t}(f^{100}_{succ})$} &\text{99 $\%$}&\text{100 $\%$}&\text{100 $\%$}&\text{99 $\%$} &\text{99 $\%$ } \\

\cline{2-8}
\text{}&\multirow{2}{*}{\textbf{Outdoors}}& \cellcolor{gray!30}\text{$f_{succ}$}  &\cellcolor{gray!30}\text{53 $\%$ (963)} &\cellcolor{gray!30}\text{76.5 $\%$ (613)}  &\cellcolor{gray!30}\text{65 $\%$ (604)} &\cellcolor{gray!30}\text{30 $\%$ (738)} &\cellcolor{gray!30}\text{91.7 $\%$ (1788)}
 \\ \cline{3-8}
\text{}&\textbf{}& \text{$max_{t}(f^{100}_{succ})$} &\text{93 $\%$}&\text{99 $\%$} &\text{94 $\%$} &\text{81 $\%$} &\text{99 $\%$}\\
\hline

\multirow{4}{*}{\textbf{Faster RCNN}} &\multirow{2}{*}{\textbf{Indoors}} & \text{$f_{succ}$} & \text{83.6 $\%$ (885)}& \text{89 $\%$ (784)} & \text{86.2 $\%$ (802)}& \text{28 $\%$ (696)}& \text{$---$}\\ \cline{3-8}
\text{}&\text{}& \text{$max_{t}(f^{100}_{succ})$} &\text{ 100$\%$}&\text{ 100$\%$}&\text{ 100$\%$}&\text{ 68$\%$} &\text{$---$} \\

\cline{2-8}
\text{}&\multirow{2}{*}{\textbf{Outdoors}}& \cellcolor{gray!30}\text{$f_{succ}$} & \cellcolor{gray!30}\text{78.1 $\%$ (717)}& \cellcolor{gray!30}\text{87.8 $\%$ (614)} & \cellcolor{gray!30}\text{79.6 $\%$ (766)}& \cellcolor{gray!30}\text{31.2 $\%$ (715)}& \cellcolor{gray!30}\text{$---$}
 \\ \cline{3-8}
 \text{}&\textbf{}&
 \text{$max_{t}(f^{100}_{succ})$} &\text{ 99$\%$}&\text{99 $\%$} &\text{ 99$\%$} &\text{ 70$\%$}& \text{$---$}\\
 \hline

\end{tabular}

\begin{tablenotes}
\footnotesize
\item[1] \scriptsize {\textbf{\quad \quad \quad *$max_{t}(f^{100}_{succ})$}: The best success rate$/$100 frames at the distance over 10$m$ .                 \textbf{*$f_{succ}$}: Success rate of total frames. \text{*($Num$): The number of total frames of this video.}
}
\end{tablenotes}
\end{table*}

\subsection{Transferability}
\label{Transferbility}
 
To evaluate the transferability of AEs, we fed our video clips recorded in the above experiments to some black-box models provided in Tensorflow detection model zoo. The black-box models include one-stage detectors such as SSD and two-stage detectors such as RFCN and Mask RCNN\footnote{Pre-trained black-box models are downloaded on the website:\\
https://github.com/tensorflow/models/blob/master/research/object$\_$detection /g3doc/detection$\_$model$\_$zoo.md}. We measured all video clips with the threshold\footnote{Typically the detector will give $N$ predictions. If the probability of the target in one prediction is lower than the threshold of 0.5, then this prediction will be filtered out.} of 0.5, which is the default value in the Tensorflow Object Detection API. Because SSD is known to have a poor performance in detecting small objects (Based on our testing, the longest distance over which SSD can detect the original stop sign is 15$m$), we fed the truncated video clips to SSD, in which the farthest shooting distance is about 12.5$m$.

\vspace{2mm}
\noindent\textbf{Transferability of AEs based on Faster RCNN.} We used the two video clips containing HA, recorded indoors and outdoors respectively, and fed them to the other four black-box models. As shown in Table \ref{Transferability}, AEs of Faster RCNN show quite good transferability performance on both the one-stage and two-stage black-box models. The video recorded in the indoor environment obtains high success rates over 90$\%$ in almost all black-box models. While in the outdoor environment, the highest success rates range from 58$\%$ to 78$\%$. Given the results above, AEs based on Faster RCNN demonstrate high transferability to black-box models even against varying angles, long distances and different experimental environments.

\vspace{2mm}
\noindent\textbf{Transferability of AEs based on YOLO V3.} The experiments about AEs of YOLO V3 include HA and AA. For HA attack, the transferability of AEs based on YOLO V3 is lower than that of Faster RCNN. The performances between indoors and outdoors for YOLO V3 also diverge significantly. For example, in the indoor environment, the success rates of AEs are mostly over 50$\%$, but for the outdoor environment, the success rates are mostly below 20$\%$. For AA, AEs perform better than HA. The success rates in outdoor environments achieve 48.2$\%$, 47.4$\%$ and 56.2$\%$ on Faster RCNN, RFCN and Mask RCNN respectively. It is interesting that the transferability of AA and HA are opposite on the same black-box model, i.e., the transferability to Mask RCNN in the outdoor environment with the success rate 9$\%$ for HA and the success rate 56.2$\%$ for AA. Such results could be explained by the detectors’ sensitivity to the stop sign. For instance, Mask RCNN was trained to be very sensitive to the stop sign, so hiding will be more difficult, but ``making up'' is relatively easier.
\vspace{-2pt}

\subsection{Effects of Style-customized AEs} 
We evaluated four different styles AEs against YOLO V3 and Faster RCNN respectively. We record the object with AEs attached, beginning from 25$m$ away and ending at about 1$m$ away, while keeping the camera facing to the stop sign during the whole recording (The frame rate is 30 frames per second). As shown in Table~\ref{Success rate of practical attack on YOLO V3},  four different style-customized AEs include 
: \textit{pattern-controlled AE} with the rectangle shapes and the specified patterns such as the clock for YOLO V3 and the person for Faster RCNN, \textit{shape-controlled AE} with various shapes such as the butterfly and Apple logo, \textit{text-based AE} with English letters and \textit{color-controlled AE} with the specified color hue based on the semantics of the target to be hidden. For example, to hide a stop sign, we need to choose the red color hue, since it is similar to the background color of a stop sign, which makes the adversarial patches more surreptitious.
Since a captured video usually lasts very long (e.g., several minutes), to perform a fine-grained measurement, we also define the success rate for every 100 consecutive frames $f^{100}_{succ}$(the frame rate is 30 frames per second). 
 Hence, $f^{100}_{succ}$ can be a good indicator to see within every 3.3 seconds, whether we have accumulated enough success (the number of successfully-attacked frames) to fool the object detector making wrong decisions.

As shown in \reftable{Success rate of practical attack on YOLO V3}, all AEs against two target models have good performances indoors, with $max_{t}(f^{100}_{succ})$ ($max(f^{100}_{succ})$ of the whole video duration time $t$) always over 99$\%$ and $f_{succ}$ mostly over 90$\%$. For outdoor environments, except the AEs of color-controlled AE, YOLO V3 gets $f_{succ}$ over 50$\%$ and $max_{t}(f^{100}_{succ})$ over 93$\%$ respectively, while Faster RCNN gets $f_{succ}$ over 80$\%$ and $max_{t}(f^{100}_{succ})$ over 99$\%$. 
However, AEs of color-controlled AE against both YOLO V3 and Faster RCNN achieve only about 30$\%$ success rate in the outdoor environment. The possible reason is the style of red color hue is originally too similar to the background color of the stop sign.

Overall, style-customized AEs can achieve a good performance in the physical attack against both YOLO V3 and Faster RCNN. Based on our experience, the area of AE has a greater impact than the shape on its success rate. However, making a larger AE is definitely easier to be noticed.

\begin{table}
\centering
\footnotesize
\caption{Success Rate in the Real-road Driving Tests}
\label{Success rate of the real-road driving test}
\begin{tabular}{m{3cm}
<{\centering}|m{2cm}
<{\centering}|m{2cm}
<{\centering}}
\hline
\textbf{Success rate}& \textbf{Straight road}& \textbf{Crossroad}\\
\hline \hline
\textbf{HA($6km/h$)} & \text{75$\%$}& \text{64$\%$}\\ \hline
\textbf{AA($6km/h$)}& \text{63$\%$} &\text{81$\%$} \\ \hline
\textbf{HA ($30km/h$)}& \text{72$\%$} &\text{60$\%$} \\ \hline
\textbf{AA ($30km/h$)}& \text{76$\%$} &\text{78$\%$} \\ \hline
\end{tabular}

\end{table}

\subsection{Real-road Driving Test}
To simulate the scenario of an object detector working on an autonomous driving car, we mounted the smartphone HUAWEI nova 3e on top of the glove compartment inside the car, recording the AEs while the car is running at different speeds. Such simulation can evaluate our AEs against varying angles, distances, and speeds. We did the experiment on a sunny day in two scenarios. In the first scenario, we placed the stop sign with HA patches or the AA poster (on a stick) on the right side of a straight road, as shown in Figure~\ref{real road driving tests} (a) and (c). Then the car started moving from 25 $m$ away, and passed by the stop sign or poster, at the speed of about 6$km/h$ and 30$km/h$ respectively. In the second scenario, the stop sign with HA patches or the AA poster were placed on one corner of the crossroad, as shown in Figure~\ref{real road driving tests} (b) and (d). We started the car from 25 meters away and made a left turn when passing the crossroad. For both of the two scenarios, a passenger sitting in the front of the car recorded the video towards the stop sign (i.e., AE). 

Table \ref{Success rate of the real-road driving test} shows the success rates of both HA and AA in the scenarios of straight road and corner road are always above 60$\%$ and even up to 81$\%$. The success rates at the speed of 30$km/h$ (similar to the speed of the real local driving) indicate that our AEs could potentially cause serious problems for autonomous driving cars. During the real road driving test, we find that HA always performs better at the long distance than at the short distance, while AA is the opposite, which also align with the experimental results in ~\cref{success rate of HA and AA}. In terms of illumination effects, we find that the ideal situation is with high illumination, but not direct shooting on the AEs, which may cause heavy reflection and downgrade the performance.

\subsection{Efficiency}
We evaluated the time required to generate AEs. For each attack, we performed the AE generation process ten times, and calculated the average time. For both AA and HA, the number of iterations in the generation process is 50,000. Without batch-variation, it took two hours and thirty minutes to finish all the iterations for AA, and two hour and fifty-five minutes for HA. After adopting batch-variation, we only need to modify the AEs 500 times. Therefore, the generation speed of AEs has been greatly improved. Specially, it took thirty minutes for AA, and thirty-five minutes for HA. The reason why HA takes a little longer than AA for both with and without batch-variation is that the image transformation used in HA introduces extra complexity. Overall, the generation efficiency of the AEs is significantly improved by the batch-variation.

\section{Discussion}
\label{Discussion}

\subsection{Impact of Backgrounds and Object integrity}
\label{The analysis of factors}

\begin{table*}
\centering
\footnotesize
\caption{\textbf{Success Rate of AEs }(Indoors vs Outdoors, with pole vs without pole, Physical vs Digital)}
\label{successrate}
\begin{tabular}{m{2cm}
<{\centering}|m{2cm}
<{\centering}|m{2.5cm}
<{\centering}|m{2.5cm}
<{\centering}|m{2.5cm}
<{\centering}|m{2.5cm}
<{\centering}}
\hline
\multicolumn{2}{c|}{\multirow{2}{*}{\textbf{Success Rate ($f_{succ}$)}}}&\multicolumn{2}{c|}{\textbf{Physical$^{\#}$}}&\multicolumn{2}{c}{\textbf{Digital$^*$}}\\
\cline{3-6}
\multicolumn{2}{c|}{\text{}}&\textbf{Indoors}&\textbf{Outdoors}&
\textbf{Indoors}&\textbf{Outdoors}\\
\hline \hline
\multirow{2}{*}{\textbf{YOLO V3}} &\textbf{With pole}& \text{92.4 $\%$}&
\textbf{53$\%$ $\ / \ $31$\%$}&
\text{89 $\%$}&
\text{73 $\%$ $\ / \ $53$\%$}\\ 
 \cline{2-6}
 \textbf{}&\textbf{Without pole}& \text{100 $\%$}&
\text{76.6 $\%$}&
\text{97 $\%$}&
\text{87 $\%$}\\ 
 \hline
\multirow{2}{*}{\textbf{Faster RCNN}} &  \textbf{With pole}& \text{75.1 $\%$}&
\textbf{67 $\%$ $\ / \ 43\%$}&
\text{93 $\%$}&
\text{74 $\%$ $\ / \ 63\%$}\\ 
 \cline{2-6}
 \textbf{}& \textbf{Without pole}& \text{91.1 $\%$}&
\text{81 $\%$}&
\text{99 $\%$}&
\text{94 $\%$}\\ 
 \hline

\end{tabular}

\begin{tablenotes}
\footnotesize
\item[1] \scriptsize {\textbf{$\#$}The average total frames of all videos in physical experiments is about 900 frames.                \textbf{$*$}The number of test images in digital space is 10000.
\textbf{$*$}Success rate / success rate*: Success rate* is the success rate of AE generated with non-enhanced realistic simulation method.
}

\end{tablenotes}
\end{table*}

As mentioned previously, we guess the effectiveness of object detectors is prone to be impacted by backgrounds and the integrity of the target object (e.g., a stop sign with the pole). Here we want to evaluate on this point. Particularly, we evaluated and analyzed the sensitivities of object detectors to different backgrounds and the object integrity in the digital space and physical world, respectively.
For evaluating the impact of background, we evaluated the AEs generated using the indoor background (unreasonable) and the outdoor background (reasonable), respectively. Also, for object integrity, we evaluated an AE targeting stop sign with a pole (reasonable) and without a pole (unreasonable), respectively. The distance ranges from 1$m$ to 25$m$, while keeping the camera facing the stop sign during the whole recording (The frame rate is 30 frames per second). Results in both physical space and digital space are shown in Table~\ref{successrate}. Generally, AEs perform better in the digital space than in the physical space as expected.

\vspace{2mm}
\noindent\textbf{Background.} As shown in Table~\ref{successrate}, no matter the stop sign with the pole installed or not, the success rate of AEs of Faster RCNN is about 10$\%$ higher when indoors than that in outdoor background. In the digital space, the success rate of AEs against stop sign is 19$\%$ higher with the indoor background than that with the outdoor background. For YOLO V3, the differences are even more clear. The success rate of AEs is 40$\%$ higher when indoors than that in outdoor background. These results demonstrate that in an outdoor background environment, the stop sign is even harder to attack. Fortunately, the stop sign is usually put outdoors. So the real situation is outdoor, which we leverage to generate AEs.

\vspace{2mm}
\noindent\textbf{Object integrity.} For AEs of Faster RCNN, the success rate of the stop sign without the pole is about 14$\%$ higher than that with the pole in the physical space; and this number is 20$\%$ outdoors in the digital space. For AEs of YOLO V3, the success rates of the stop sign without pole are 7$\%$ and 23$\%$ higher than that with pole indoors and outdoors, respectively, in the physical world. In the digital space, the success rate of stop sign without a pole are 14$\%$ and 8$\%$ higher than that with a pole outdoors and indoors, respectively.  Apparently, both YOLO V3 (one-stage detector) and Faster RCNN (two-stage detector) are sensitive to object integrity. 
To make evaluations more reliable, the stop sign is always installed with the pole in all the experiments.

The experimental results show that the background and object integrity are important for adversarial attacks. So to be more realistic, we apply different backgrounds and keep the object integrity when generating AEs, which also increases the robustness of the generated AEs.

\subsection{Impacts of Attacks against Real World Object Detectors} 
Object detectors are becoming widely used in the areas of autonomous driving, intelligent video surveillance and etc. Compromising object detectors in a surreptitious way could incur a significant loss to people's property and even life. First, deep-learning based perception module is fundamental to enable autonomous driving vehicles, providing crucial information about the driving environment, e.g., traffic lights, traffic lanes, road signs, etc. The object detector is one of the core units in the perception module, relying on the input from the camera sensors~\cite{stavens2011learning}. If adversarial example against the traffic sign can successfully deceive the object detector, the perception module may present false information to the control system of the car. Then the system would probably make wrong decisions, which could result in traffic accidents\footnote{Although perception module also relies on information from other sensors like LIDAR, RADAR, etc., to make decisions, they cannot present the semantics of the objects, e.g., stop sign, traffic lights, etc.}. Second, Intelligent Surveillance System (ISS) is able to automatically analyze the image, video, audio or other types of surveillance data without or with limited human intervention. The adversary attack against the object detector in ISS may cause it unable to identify dangerous persons/objects or anything that needs to be monitored. Such kind of detection failure due to attacks also poses great threats to the safety of the people's property and life. 

\subsection{Potential Defense}
To the best of our knowledge, there does not exist any general defense mechanism for adversarial attacks against object detectors, since researchers are still investigating the feasibility of such attacks in the real world. Therefore, we study the defense mechanisms of adversarial attacks against image classifiers, and discuss the possibility of applying such defense solutions for object detectors. 

The defense mechanisms we consider can be grouped into three categories: (1) Modifying the inputs to disturb or even remove the adversarial perturbations, e.g., JPEG compression \cite{dziugaite2016study}, randomization \cite{xie2017mitigating}, median-filter \cite{xie2018feature}, image re-scaling \cite{guo2017countering}, etc. Furthermore, Fangzhou et al. \cite{liao2018defense} proposed to train a guided denoiser to remove the perturbations of AEs. However none of these pixel-based image processing, transformation and denoising methods are very likely to defeat our AEs. The reason is that our AEs are generated with various transformations and random noise, thus those approaches might not be able to disturb the perturbations in our AEs effectively. The guided denoiser trained based on a large amount of AEs (including ours) against object detectors is potentially an effective defense solution. However, building the corpus of AEs is not an easy task. (2) Improving the models such as the adversarial training \cite{tramer2017ensemble}, defense distillation \cite{papernot2016distillation} and gradients obfuscation \cite{athalye2018obfuscated}. However, such defense is limited to re-attack and transferable attack. We can bypass it through transferability or generating new AEs against the improved models. (3) Defeating AEs with GAN. A classifier can be trained to distinguish whether the input is adversarial or not using GAN~\cite{samangouei2018defense}. However, such GAN needs to be trained based on our AEs to defeat our attack.

\section{Related work}
\label{sec:related work}

Several existing works target at the adversarial attack against video processing systems, especially the object detectors, in the digital space. 
Xie et al.~\cite{xie2017adversarial} extended adversarial examples to semantic segmentation and object detection in the digital space. Lu et al.~\cite{lu2017no} demonstrated their adversarial examples against Faster-RCNN and YOLO generalize well across a sequences of digital images. When testing their samples in the physical world, most of the AEs cannot deceive the detectors even the AEs were poorly distorted against the background.
Moreover, the perturbations are quite large since they modified the whole stop sign when generating AE. Yang et al.~\cite{DBLP:journals/corr/abs-1810-05206} presented an interesting idea of creating a 3D mesh representation to attack object detectors digitally, but the effectiveness of the physical 3D adversarial objects are still unknown. 

There are few existing works attacking object detectors in the physical space. Shapeshifter~\cite{DBLP:journals/corr/abs-1804-05810} extended the EoT method~\cite{brown2017adversarial} to attack the Faster R-CNN object detector. Eykholt et al. proposed the physical attack to YOLO V2 object detector~\cite{song2018physical}. Both of the two works evaluated their AEs in some physical scenarios and worked well as expected. Since they were designed without considering the robustness against various physical conditions, those AEs are still limited in varying distance, angles, etc. In contrast, we designed more robust and practical AEs against the real world object detectors, which demonstrate better performance at both longer distance and wider angles.

There have been a lot of prior works~\cite{papernot2016limitations}~\cite{carlini2017towards}~\cite{kos2018adversarial}~\cite{yuan2018commandersong} on investigating the vulnerability of deep neural networks against adversarial examples. Szegedy et al~\cite{szegedy2013intriguing} showed surreptitious adversarial examples can mislead the DNN-based image classifiers. Goodfellow et al.~\cite{goodfellow6572explaining} found that a large fraction of adversarial examples are classified incorrectly by ImageNet  when perceived through the camera. Kurakin et al.~\cite{kurakin2016adversarial} demonstrated that the adversarial examples can still be effective to classifiers when printed out. Athalye et al.~\cite{athalye2017synthesizing} implemented a 3D printed adversarial object, which can deceive the neural networks at different orientations and scales. All these researches focus on adversarial samples against image classifiers, rather than object detectors.

\section{Conclusion}
\label{sec:conclusion}

In this paper, we presented  a robust and practical adversarial attack against the real world object detectors. In particular, we proposed feature-interference reinforcement, enhanced realistic constraints generation, nested AEs to improve the robustness of AEs in the physical world against various factors, like varying distances, angles, backgrounds, illumination, etc. The experimental results show that our adversarial examples are robust in the real world environments, capable of attacking the state-of-the-art real-time object detectors, e.g., YOLO V3 and faster-RCNN, at the distance ranging from 1$m$ to 25$m$ and angle ranging from $-60^\circ$ to $60^\circ$. The real-road tests, placing the object detectors in a car running at the speed of 30$km/h$, achieve the success rate over 72$\%$. Furthermore, the evaluation results also demonstrate high transferability of our AEs to other black box object detectors.



%
\begin{acks}
IIE authors are supported in part by National Key R\&D Program of China (No.2016QY04W0805), NSFC U1836211, 61728209, National Top-notch Youth Talents Program of China, Youth Innovation Promotion Association CAS, Beijing Nova Program, Beijing Natural Science Foundation (No.JQ18011) and National Frontier Science and Technology Innovation Project (No. YJKYYQ20170070).  
\end{acks}

%
\bibliographystyle{ACM-Reference-Format}
\bibliography{sample.bib}

%
\begin{onecolumn}
  
\section*{Appendix}
\label{appendix}


\begin{figure*}[ht]
\centering
\subfigure[]{

\begin{minipage}[t]{0.18\linewidth}
\centering
\epsfig{figure=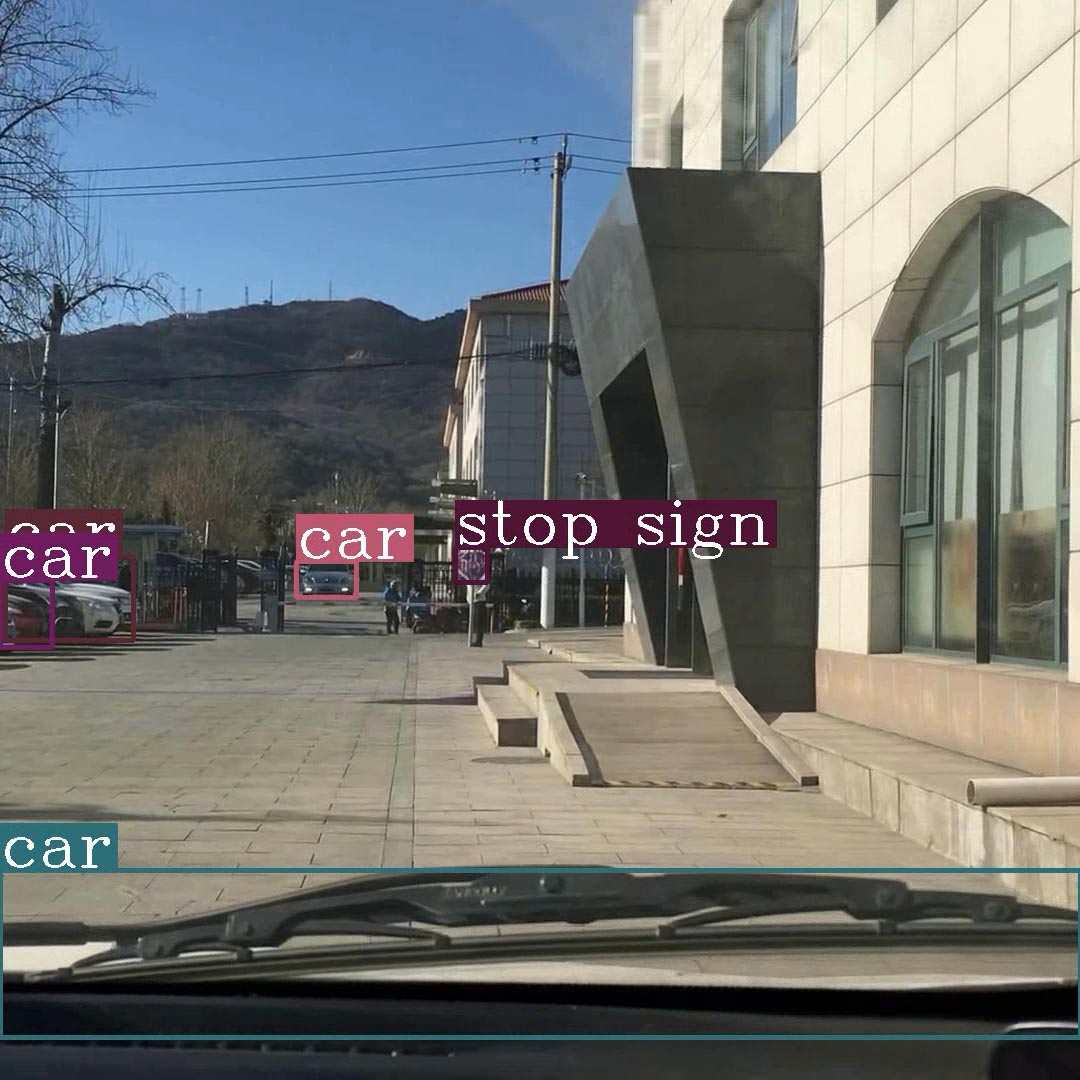, width=0.8\textwidth}

\epsfig{figure=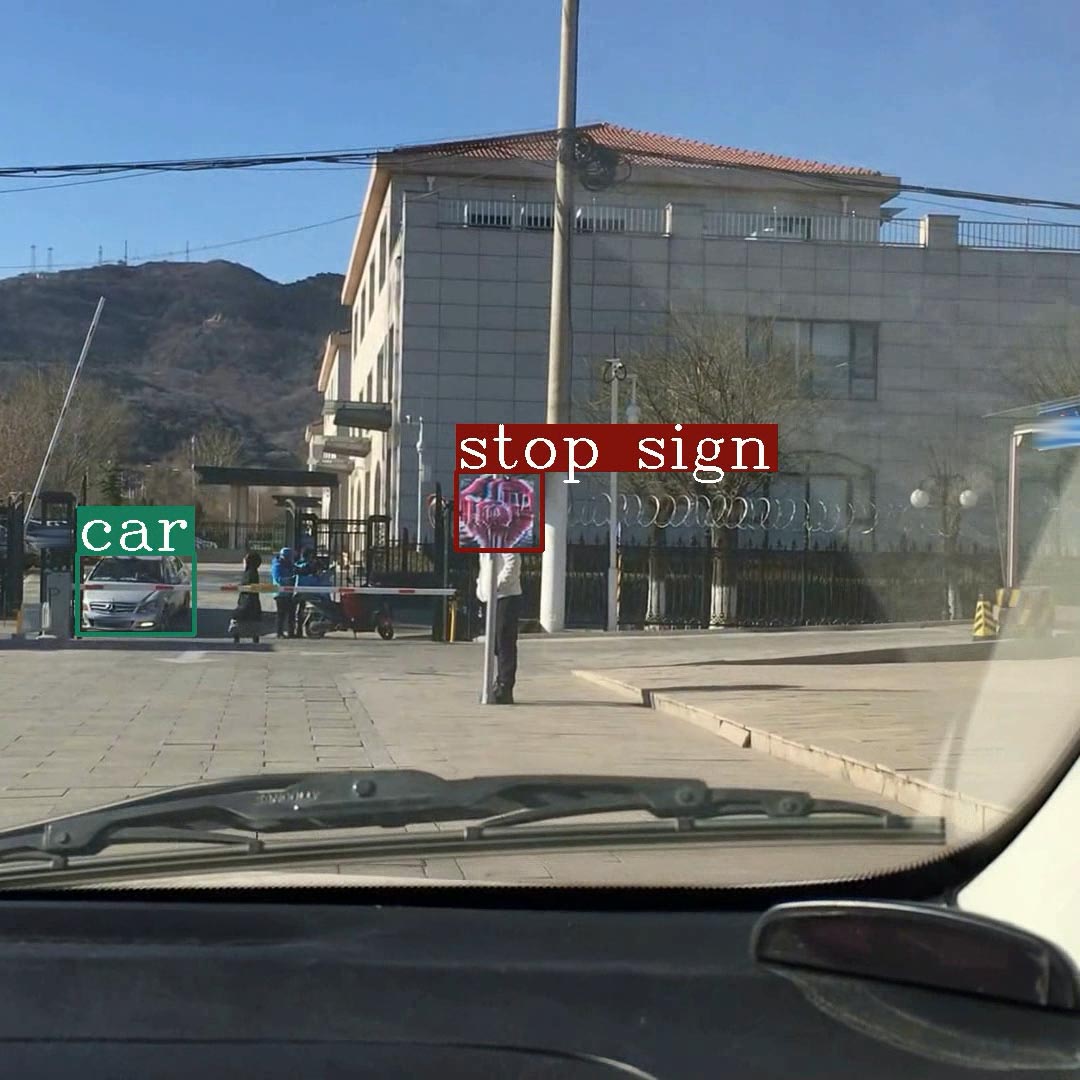, width=0.8\textwidth}

\epsfig{figure=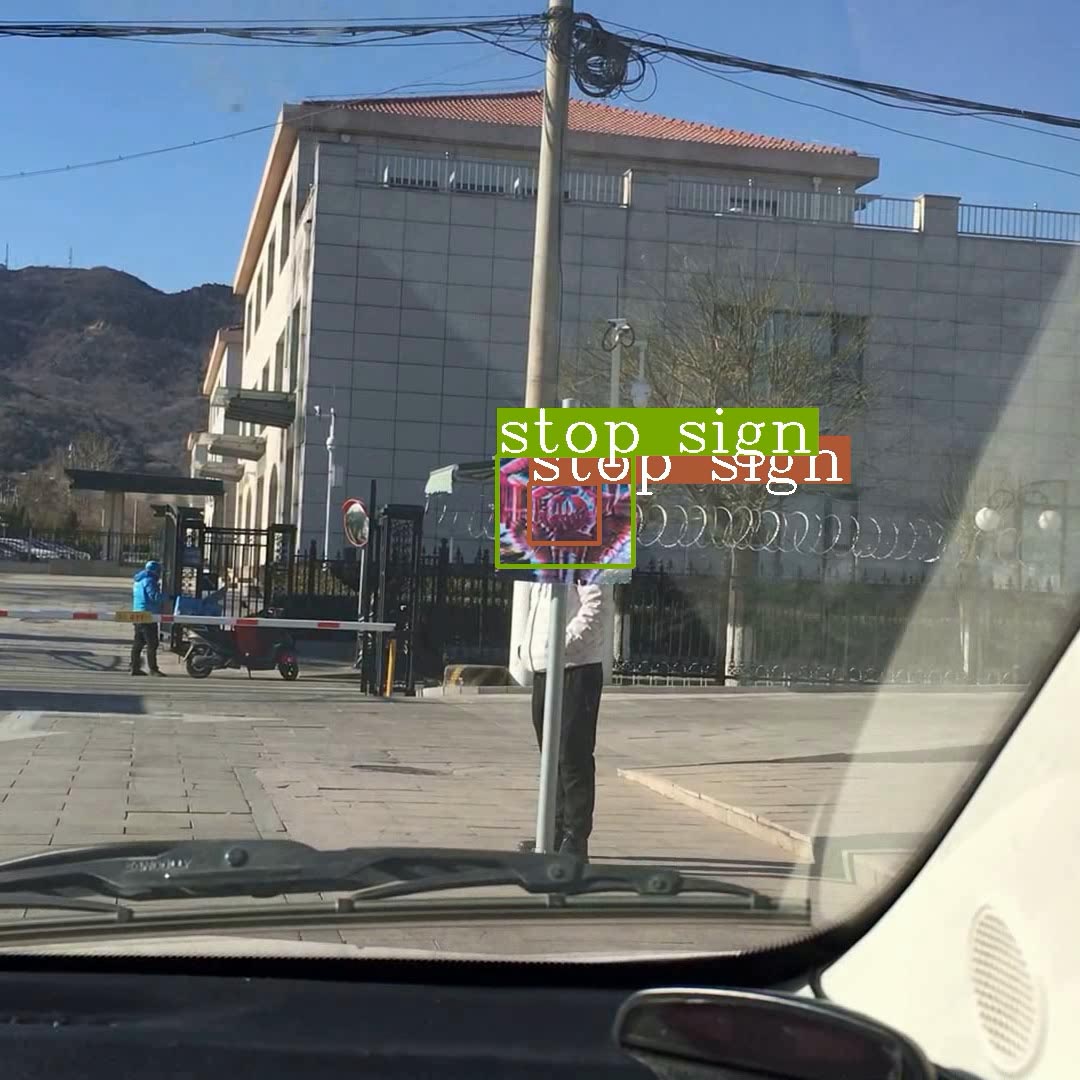, width=0.8\textwidth}
\end{minipage}%
}%
\subfigure[]{
\begin{minipage}[t]{0.18\linewidth}
\centering
\epsfig{figure=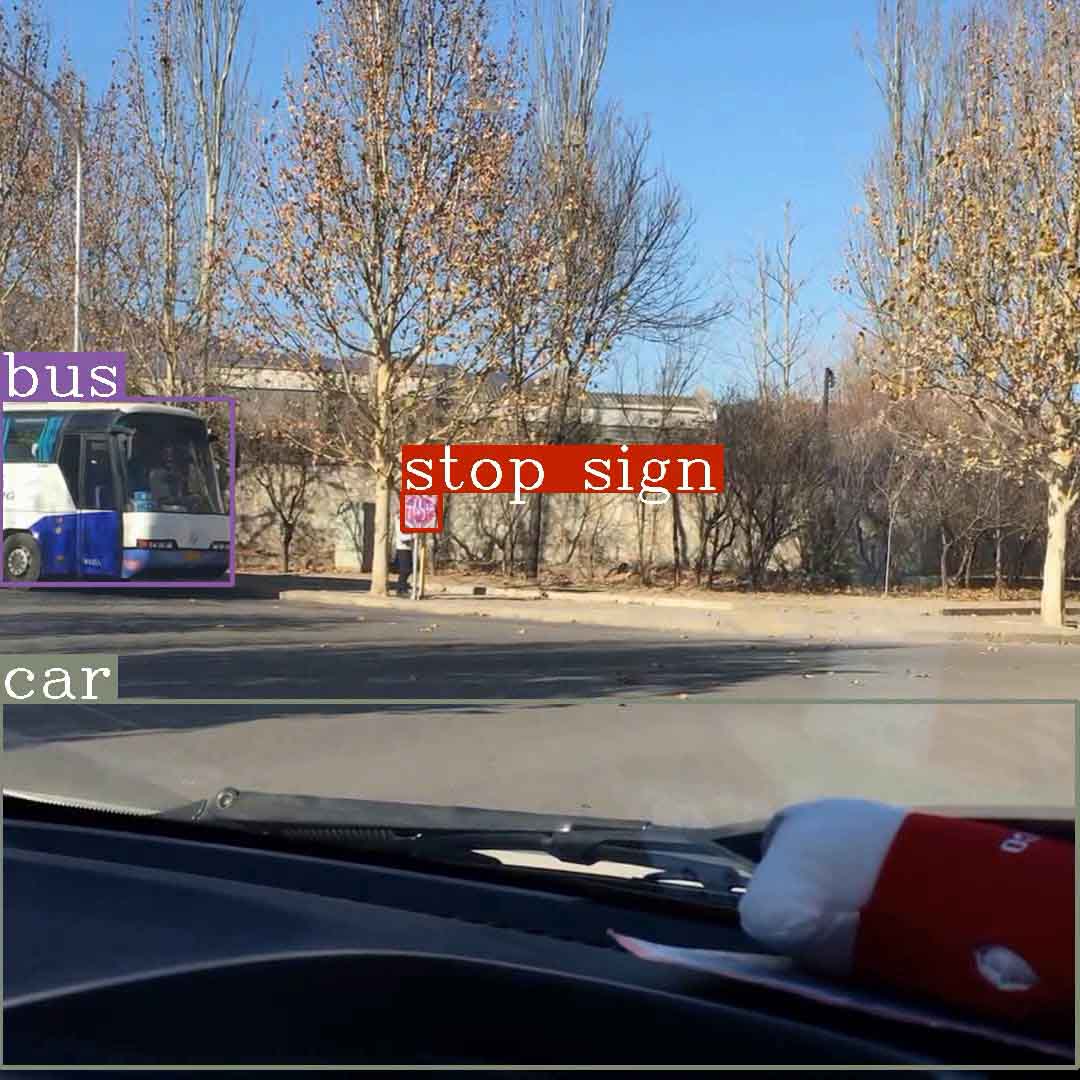, width=0.8\textwidth}

\epsfig{figure=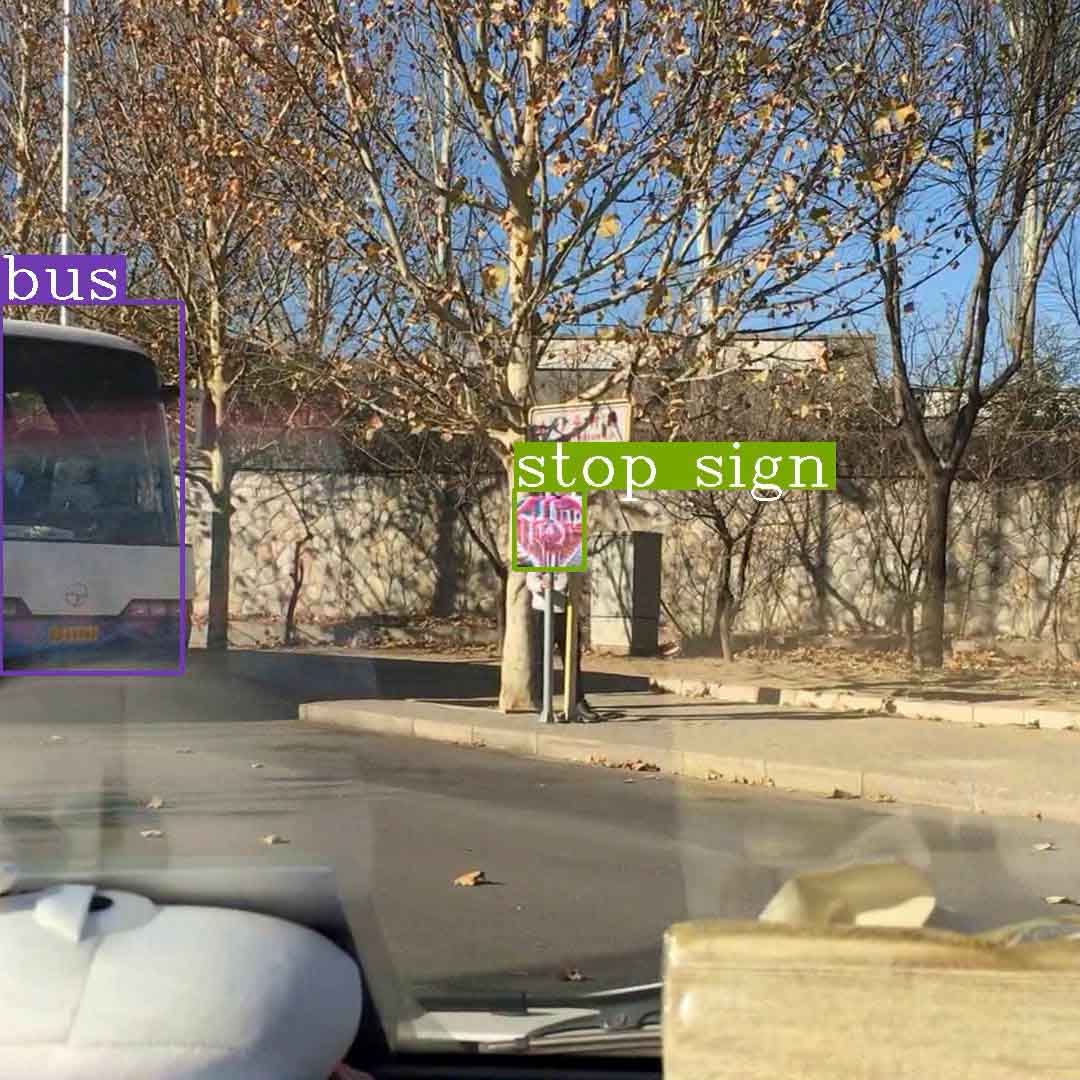, width=0.8\textwidth}

\epsfig{figure=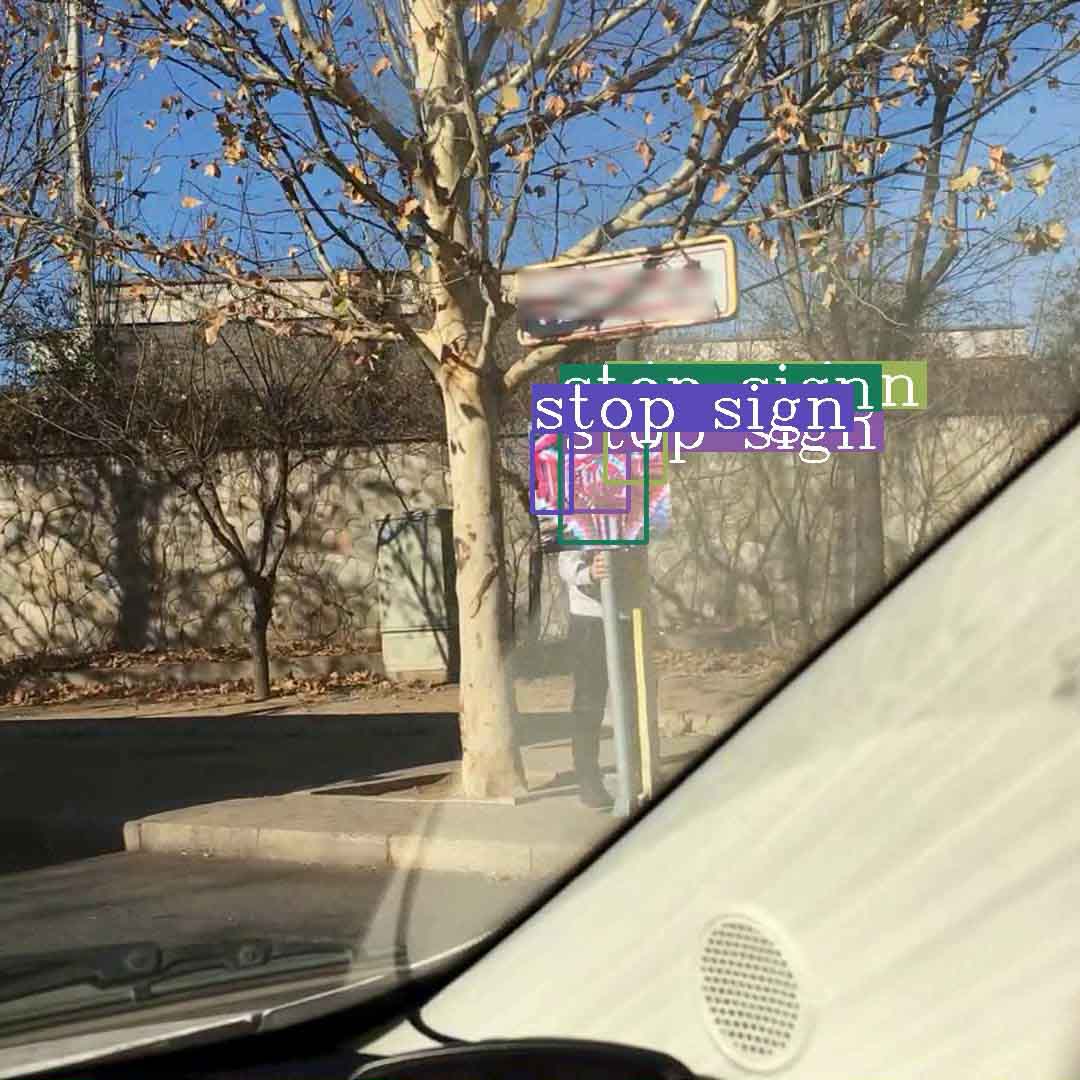, width=0.8\textwidth}
\end{minipage}%
}%
\subfigure[]{
\begin{minipage}[t]{0.18\linewidth}
\centering
\epsfig{figure=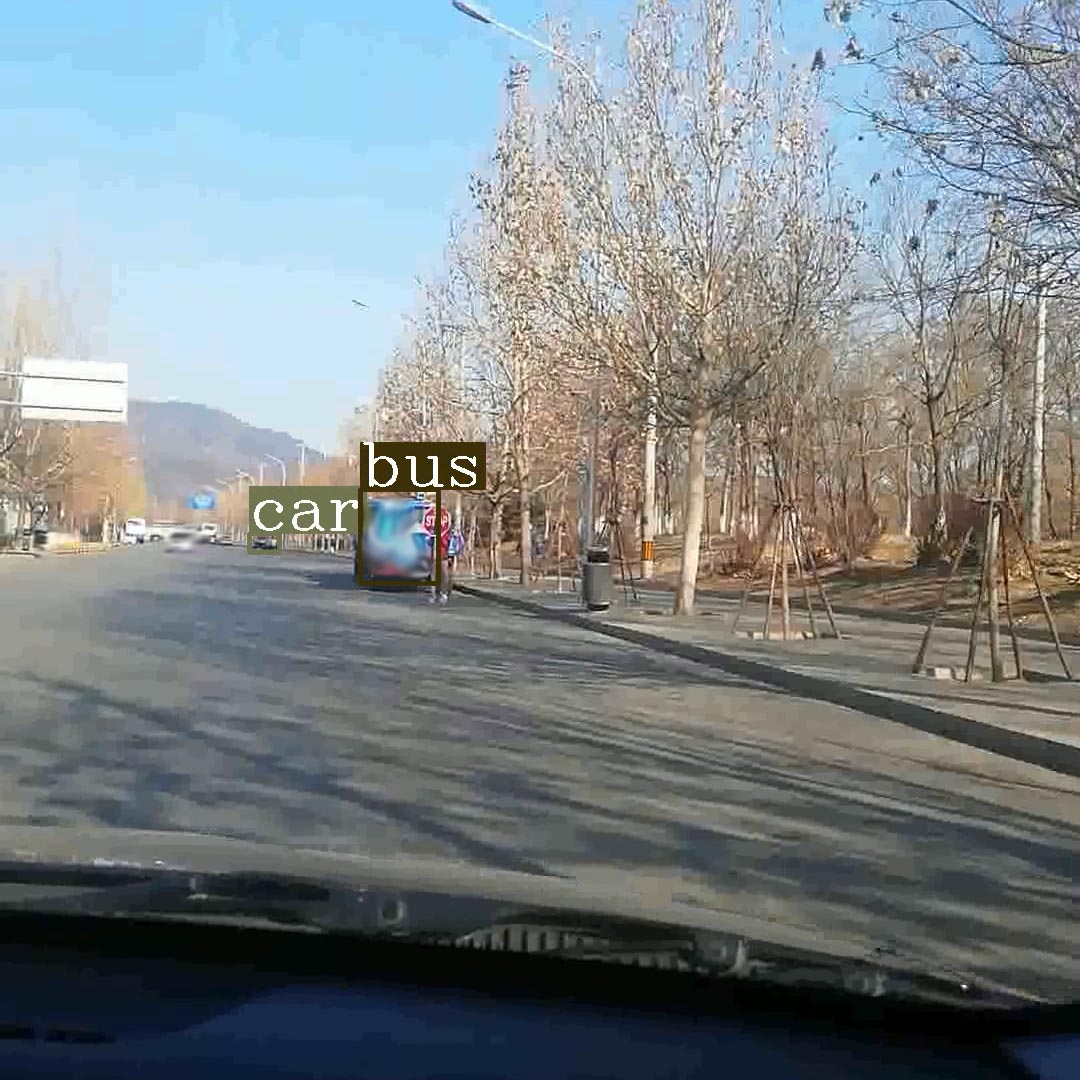, width=0.8\textwidth}

\epsfig{figure=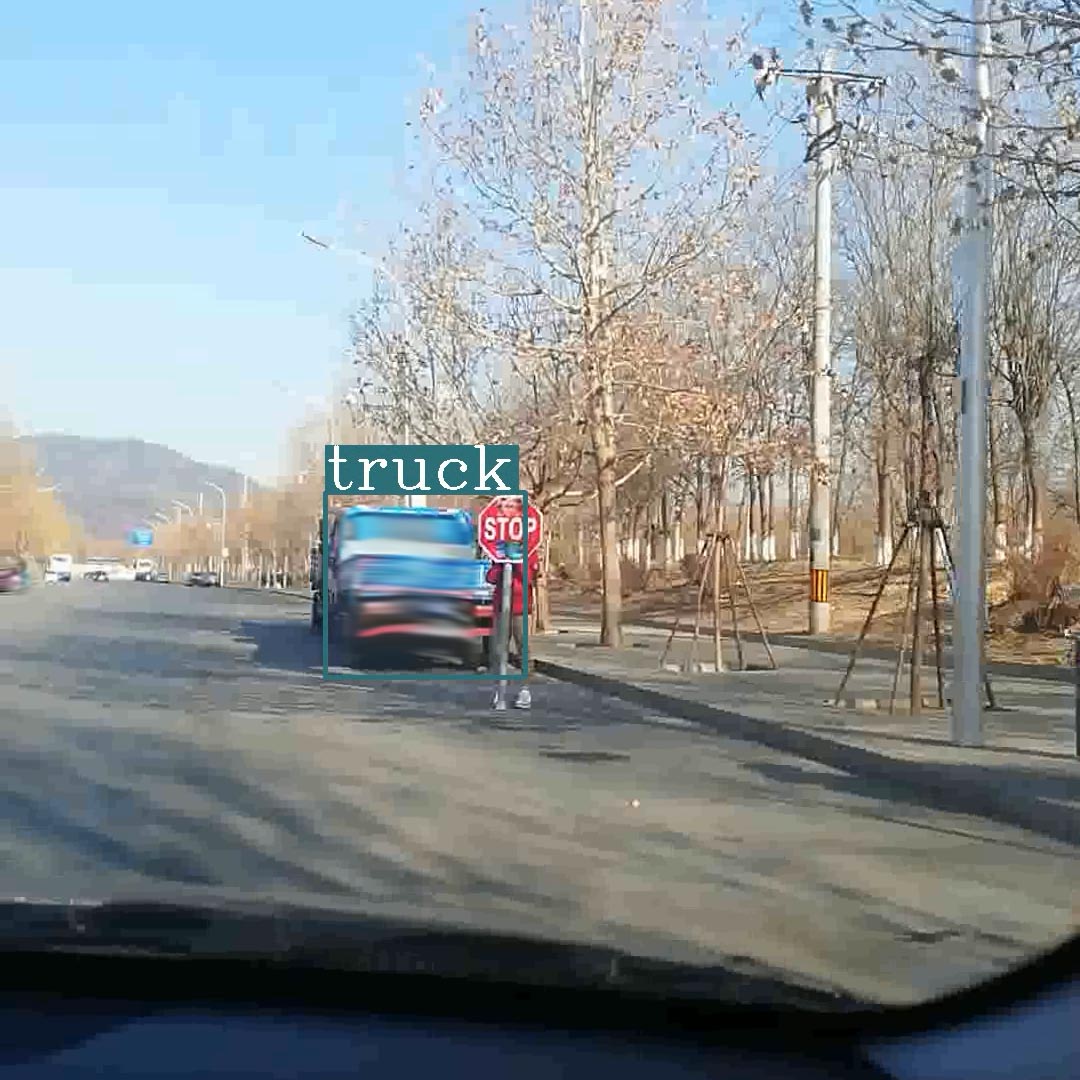, width=0.8\textwidth}

\epsfig{figure=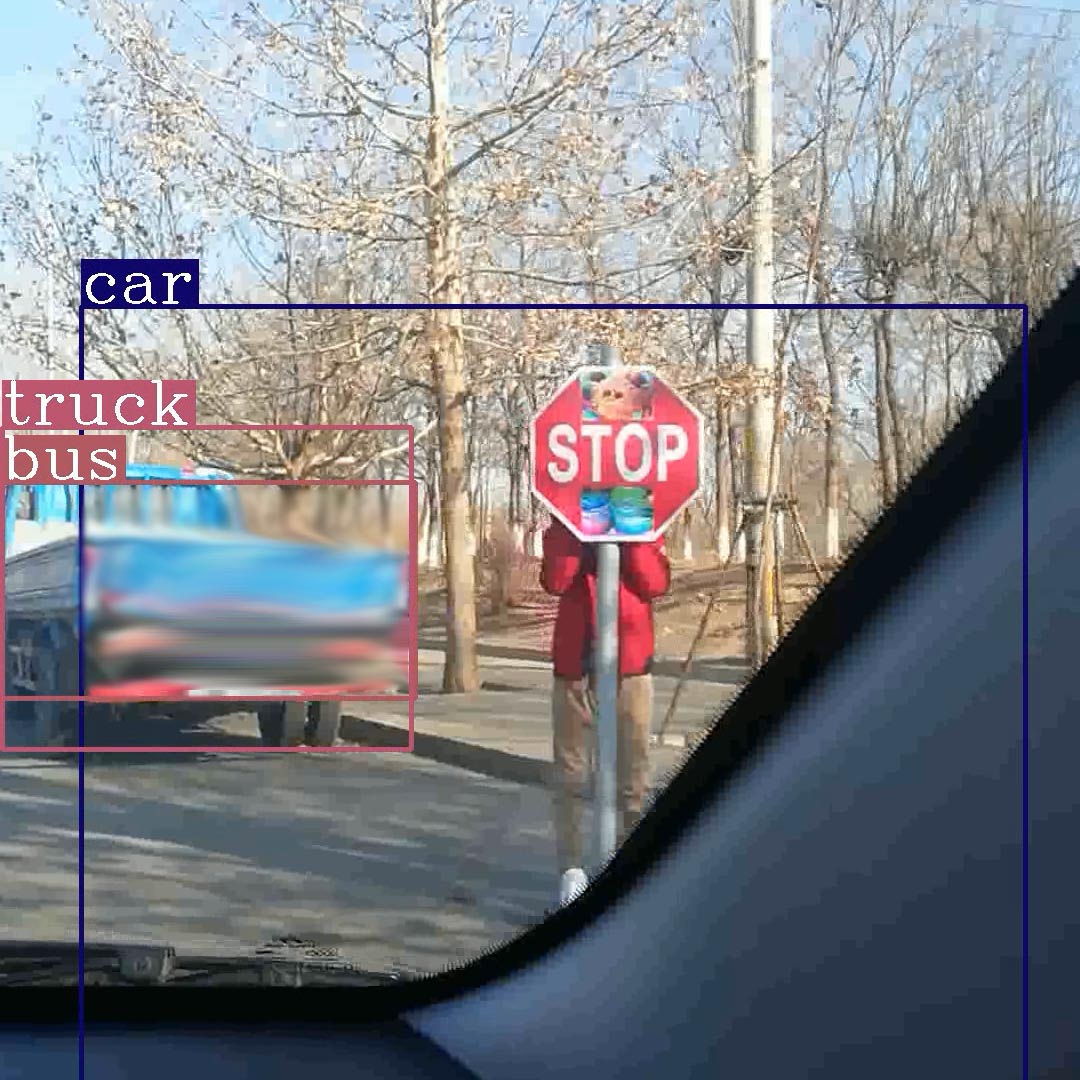, width=0.8\textwidth}
\end{minipage}%
}%
\subfigure[]{
\begin{minipage}[t]{0.18\linewidth}
\centering
\epsfig{figure=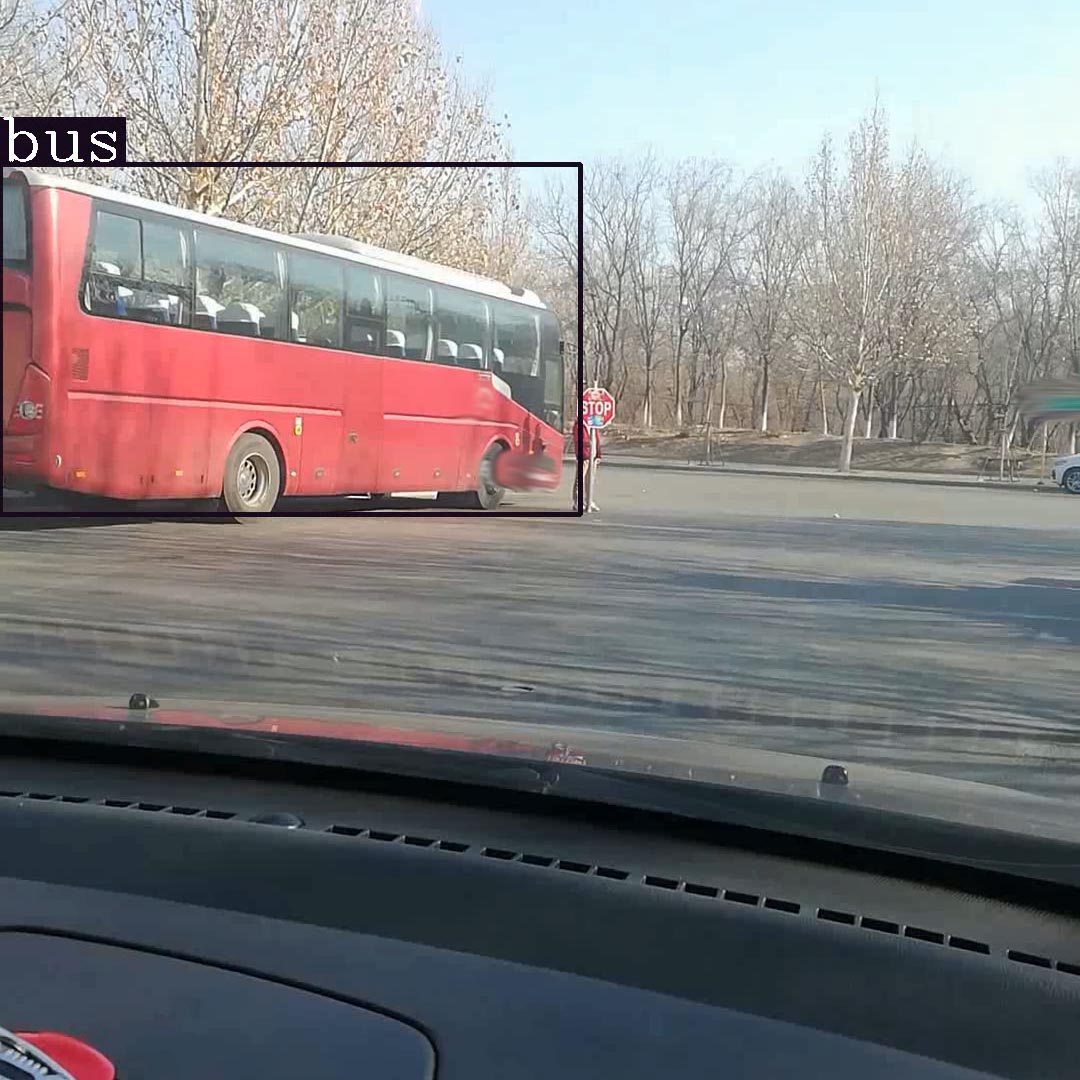, width=0.8\textwidth}

\epsfig{figure=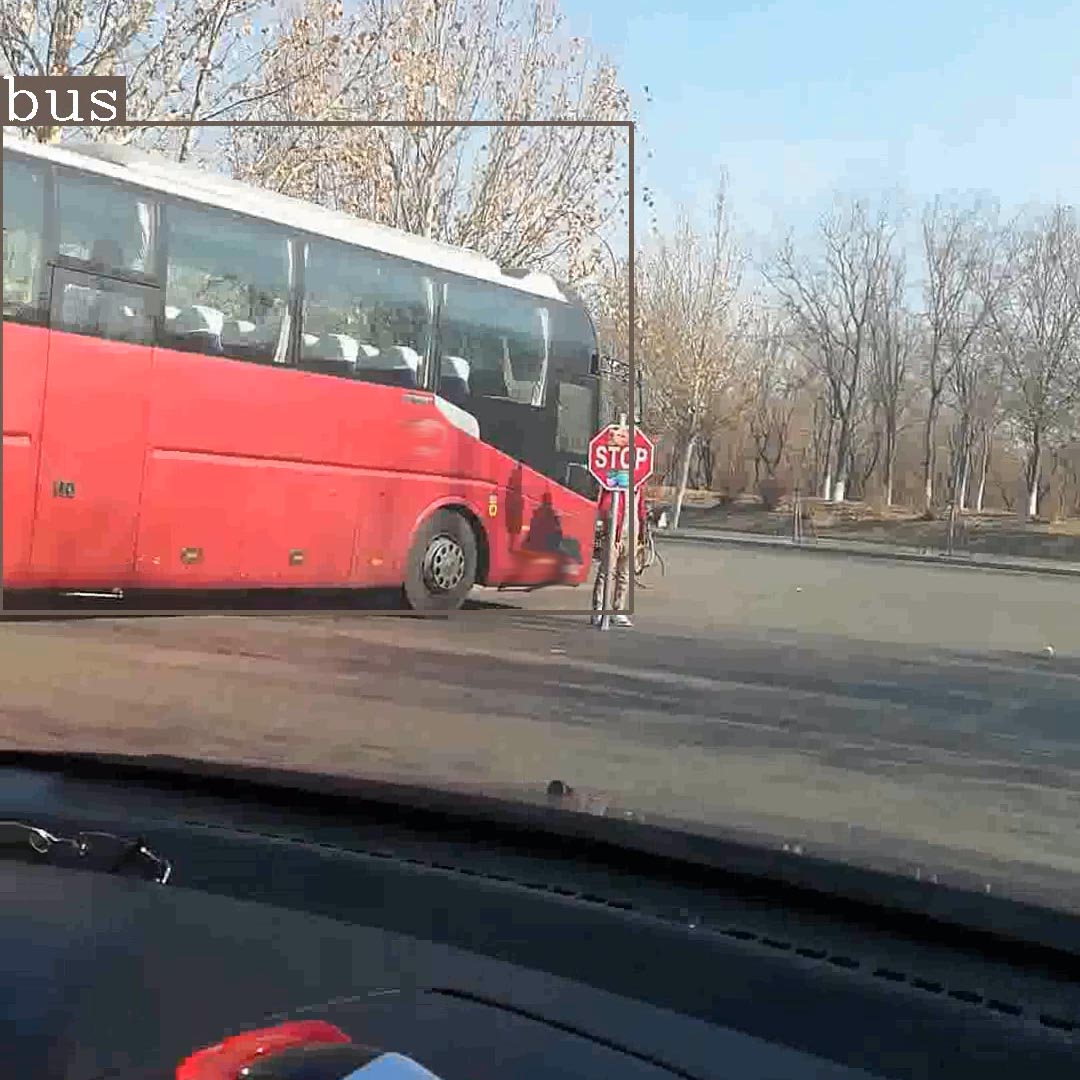, width=0.8\textwidth}

\epsfig{figure=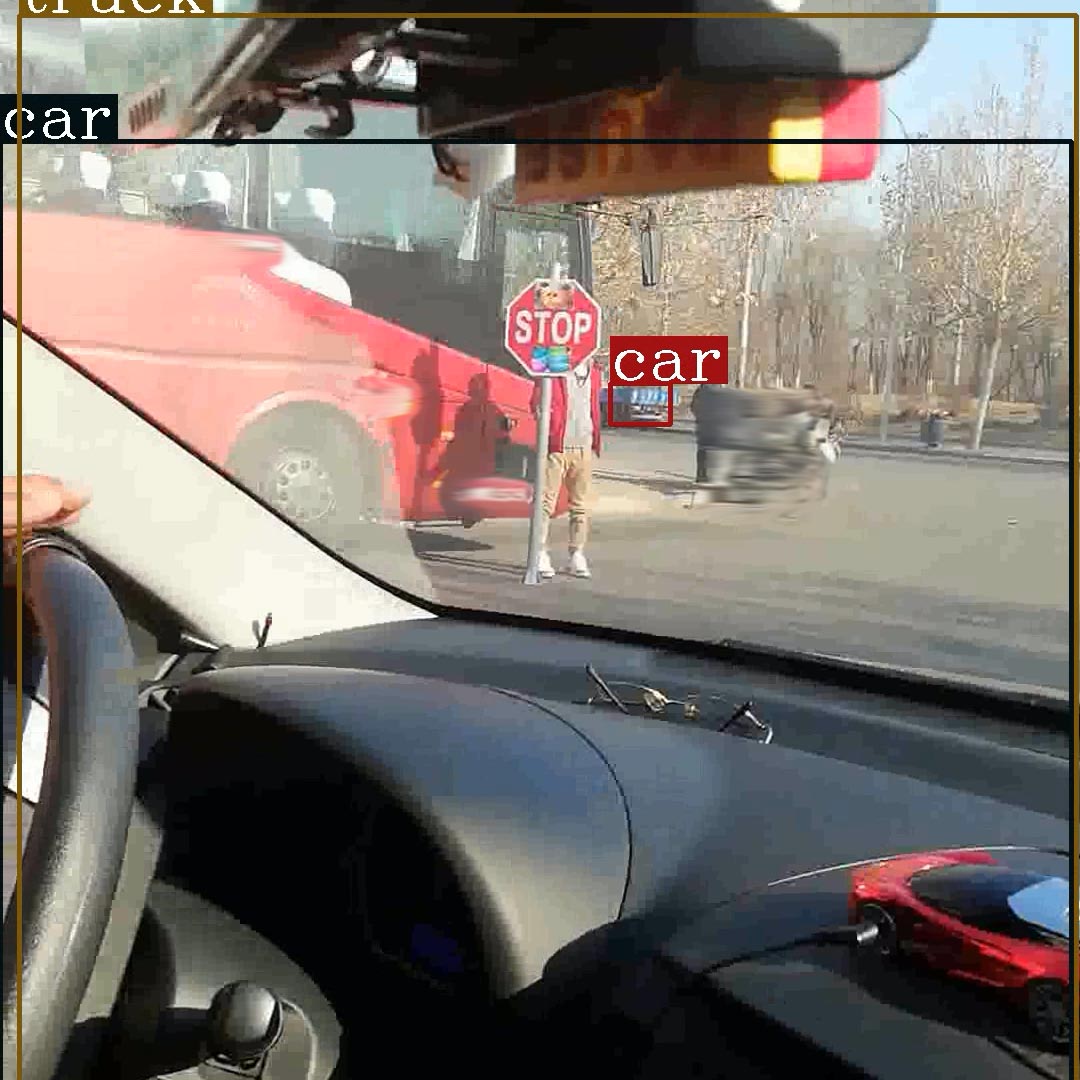, width=0.8\textwidth}
\end{minipage}%
}%
\caption{\textbf{Sample Frames of Real-road Driving Tests.} (a) The hiding attack on a straight road. (b) The hiding attack on a crossroad. (c) The appearing attack on a straight road. (d) The appearing attack on a crossroad. }
\label{real road driving tests}
\end{figure*}

\begin{figure*}[ht]
\centering
\subfigure[]{

\begin{minipage}[t]{0.18\linewidth}
\centering
\epsfig{figure=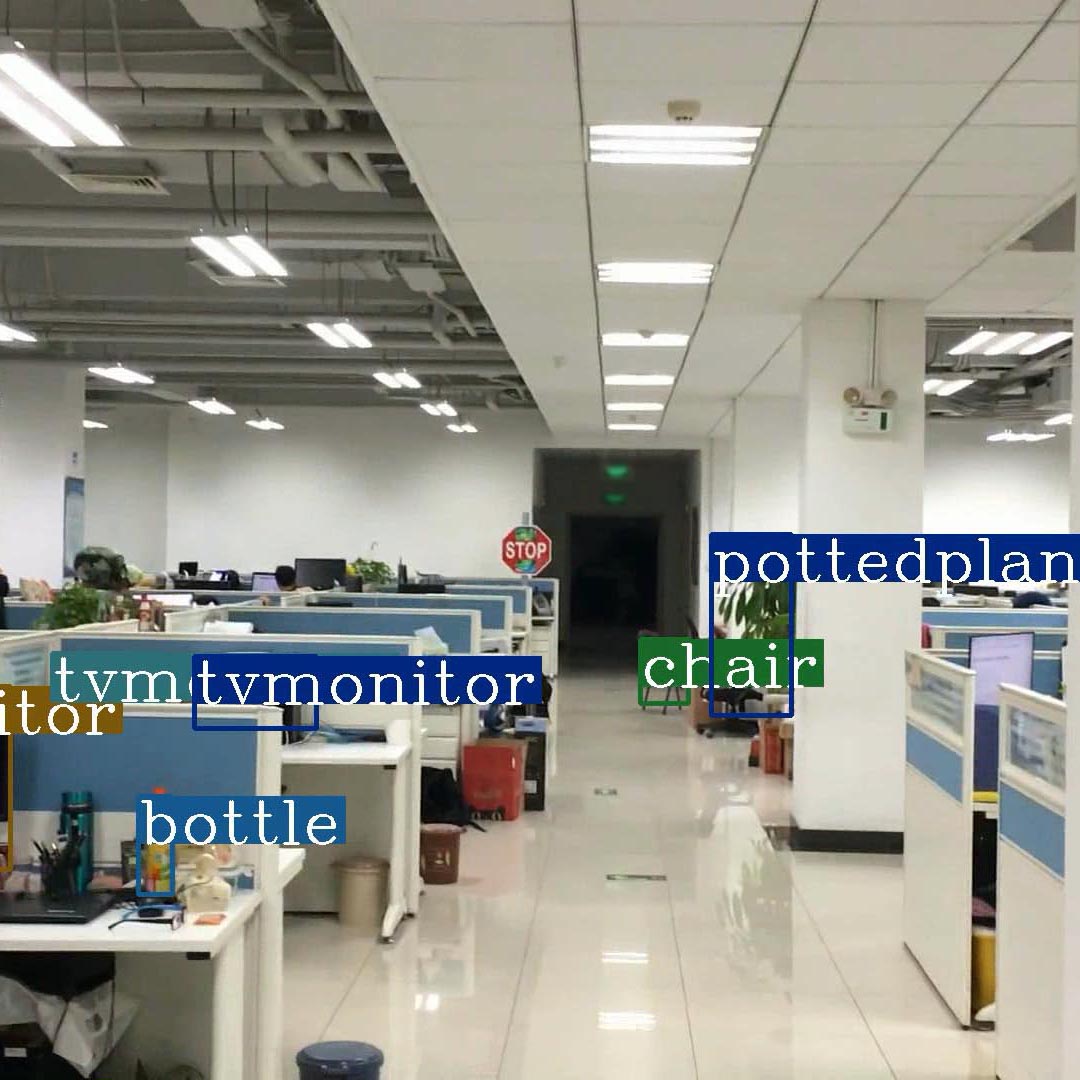, width=0.8\textwidth}

\epsfig{figure=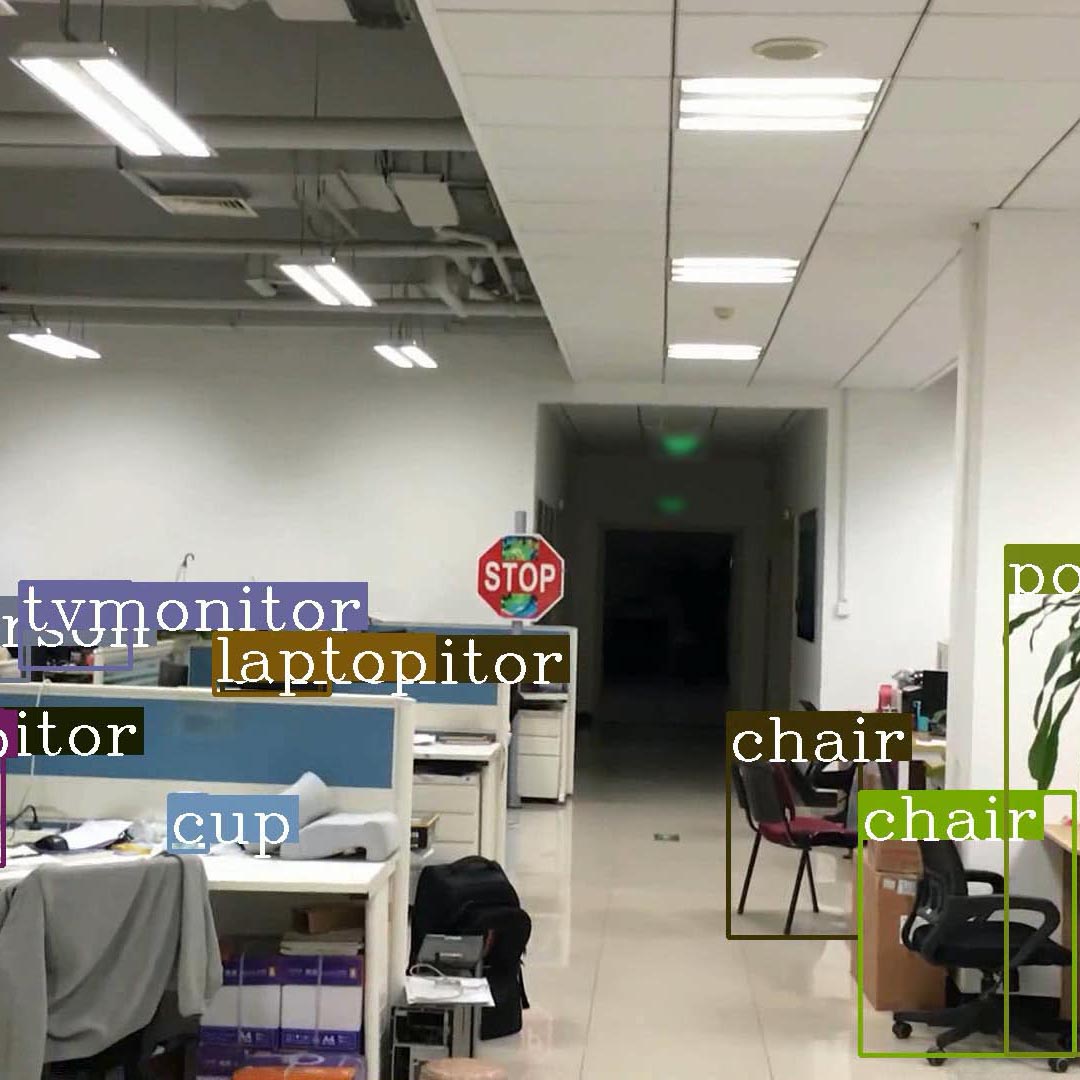, width=0.8\textwidth}

\epsfig{figure=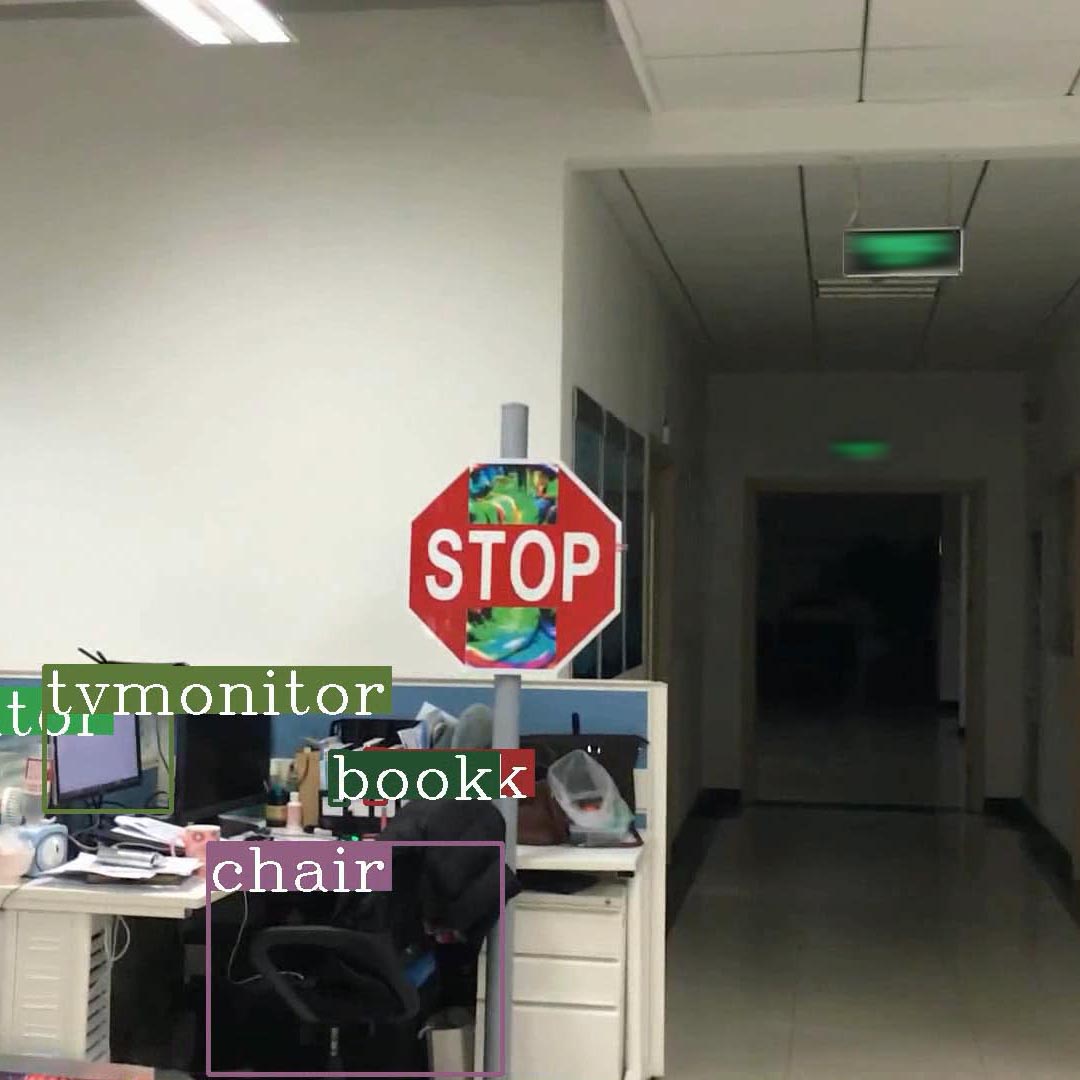, width=0.8\textwidth}
\end{minipage}%
}%
\subfigure[]{
\begin{minipage}[t]{0.18\linewidth}
\centering
\epsfig{figure=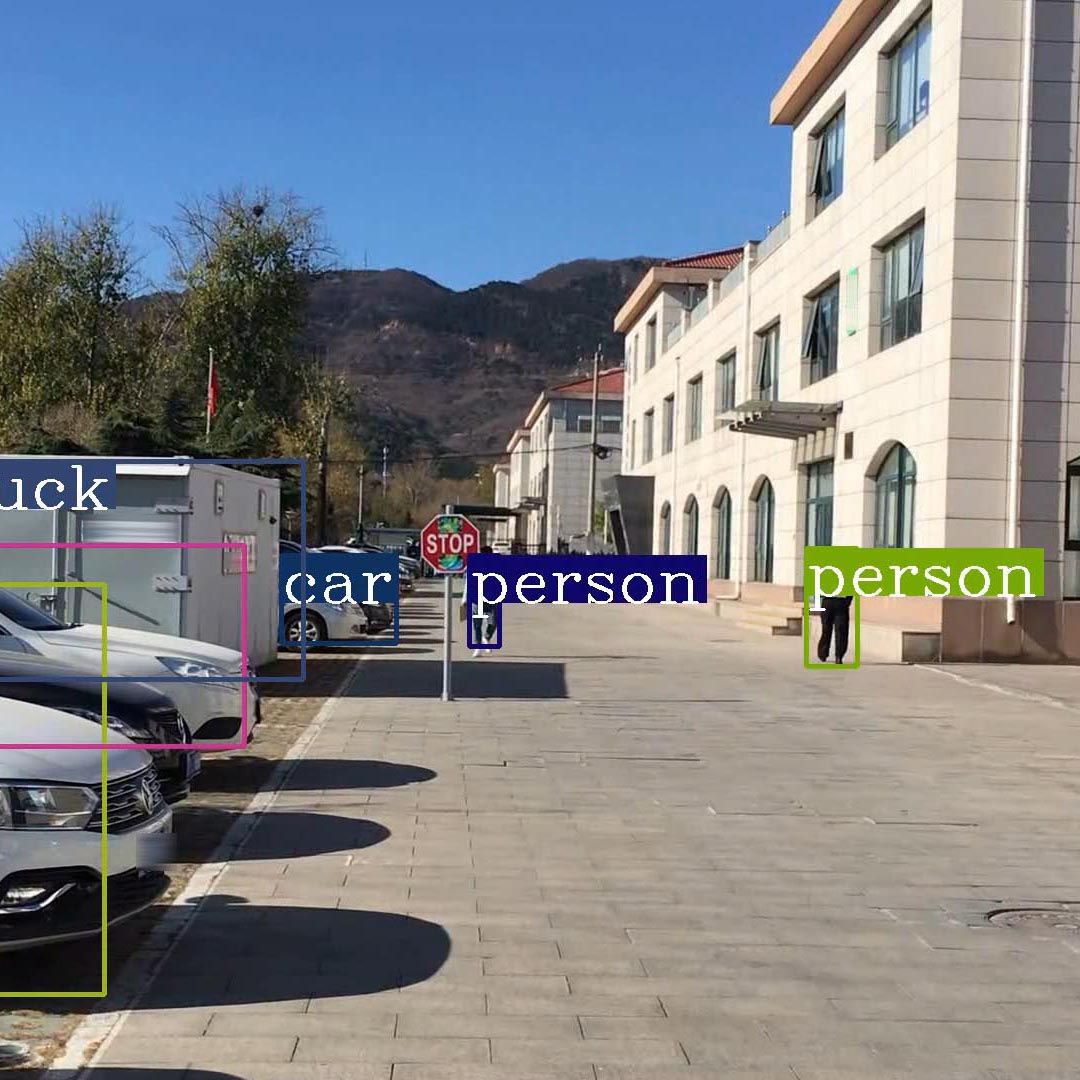, width=0.8\textwidth}

\epsfig{figure=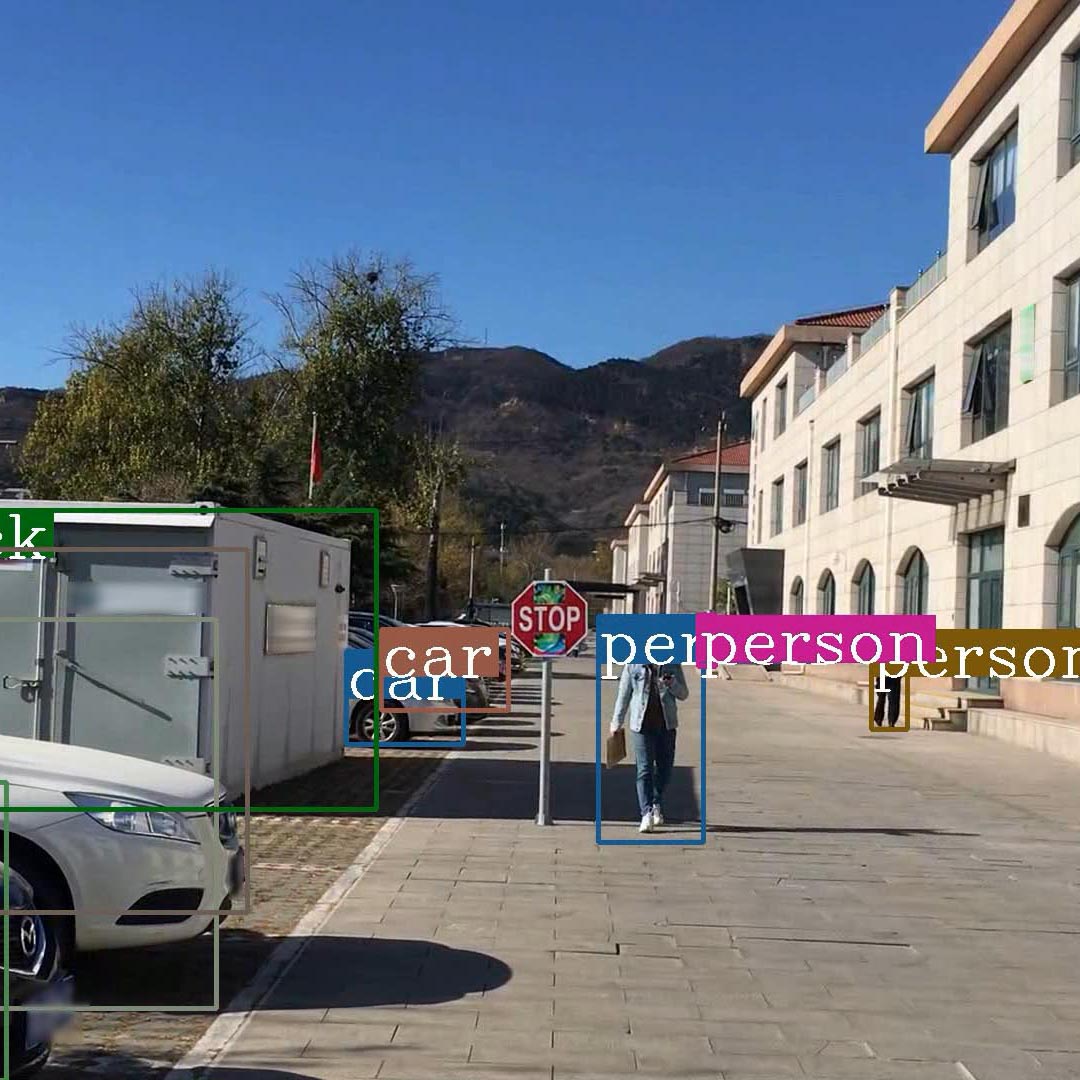, width=0.8\textwidth}

\epsfig{figure=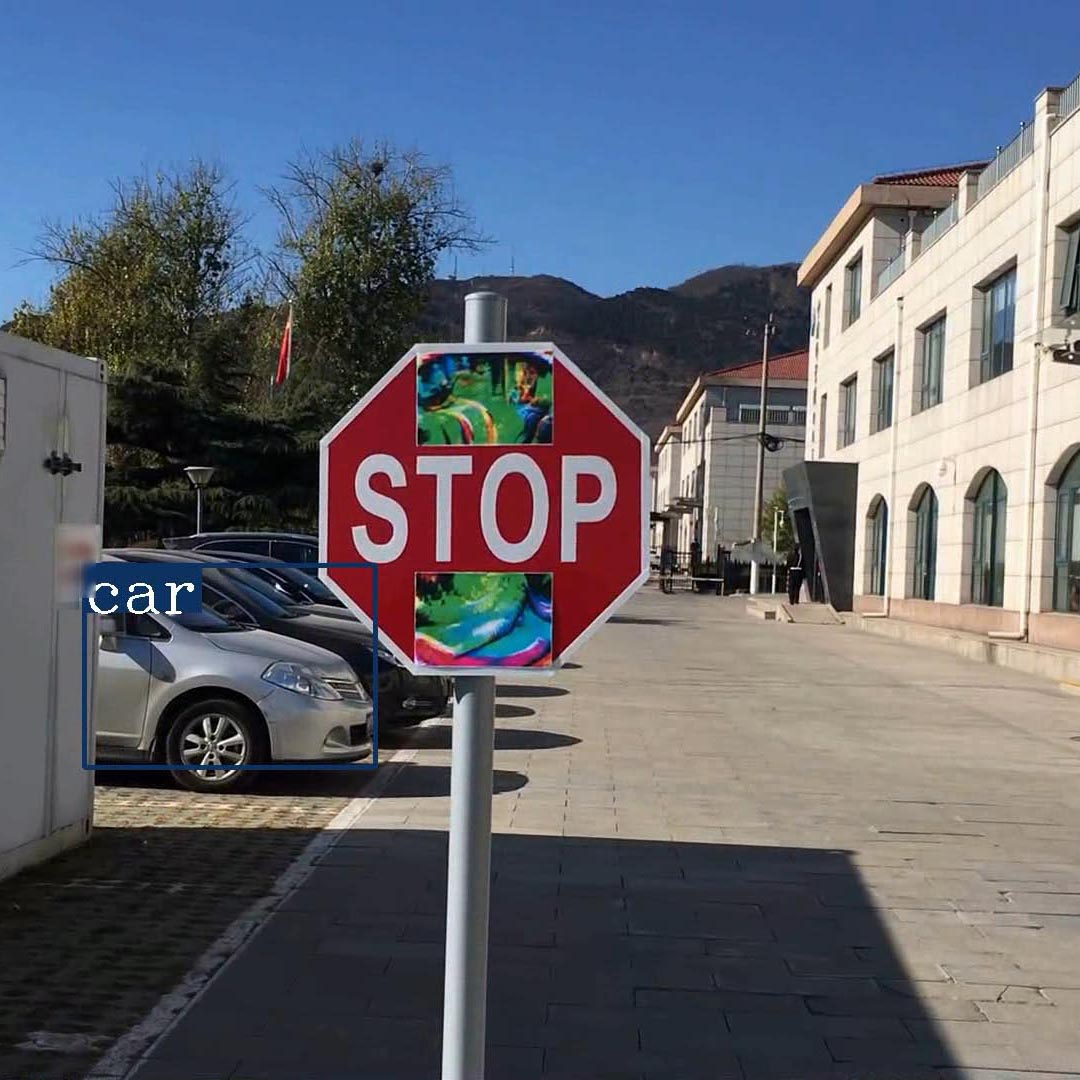, width=0.8\textwidth}
\end{minipage}%
}%
\subfigure[]{
\begin{minipage}[t]{0.18\linewidth}
\centering
\epsfig{figure=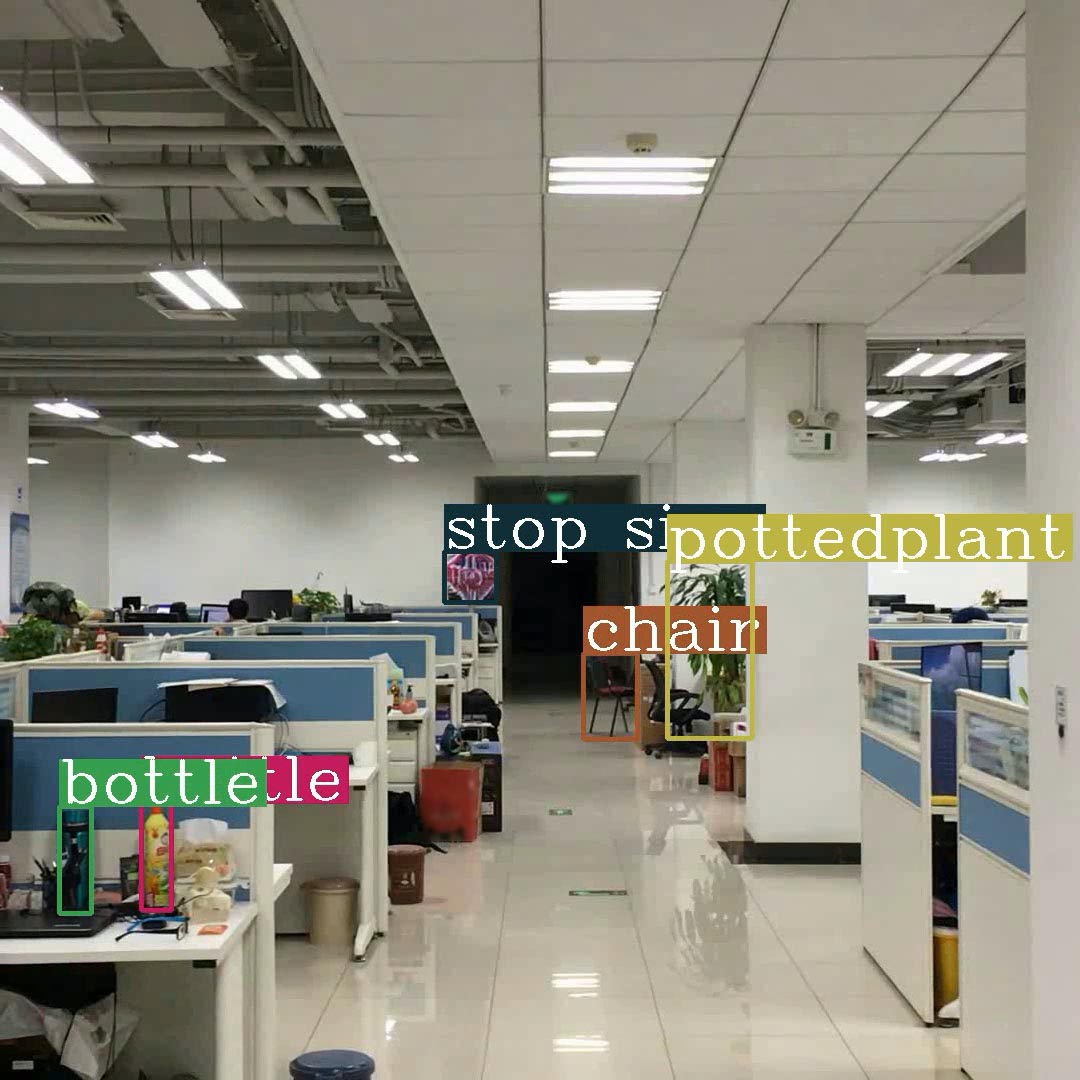, width=0.8\textwidth}

\epsfig{figure=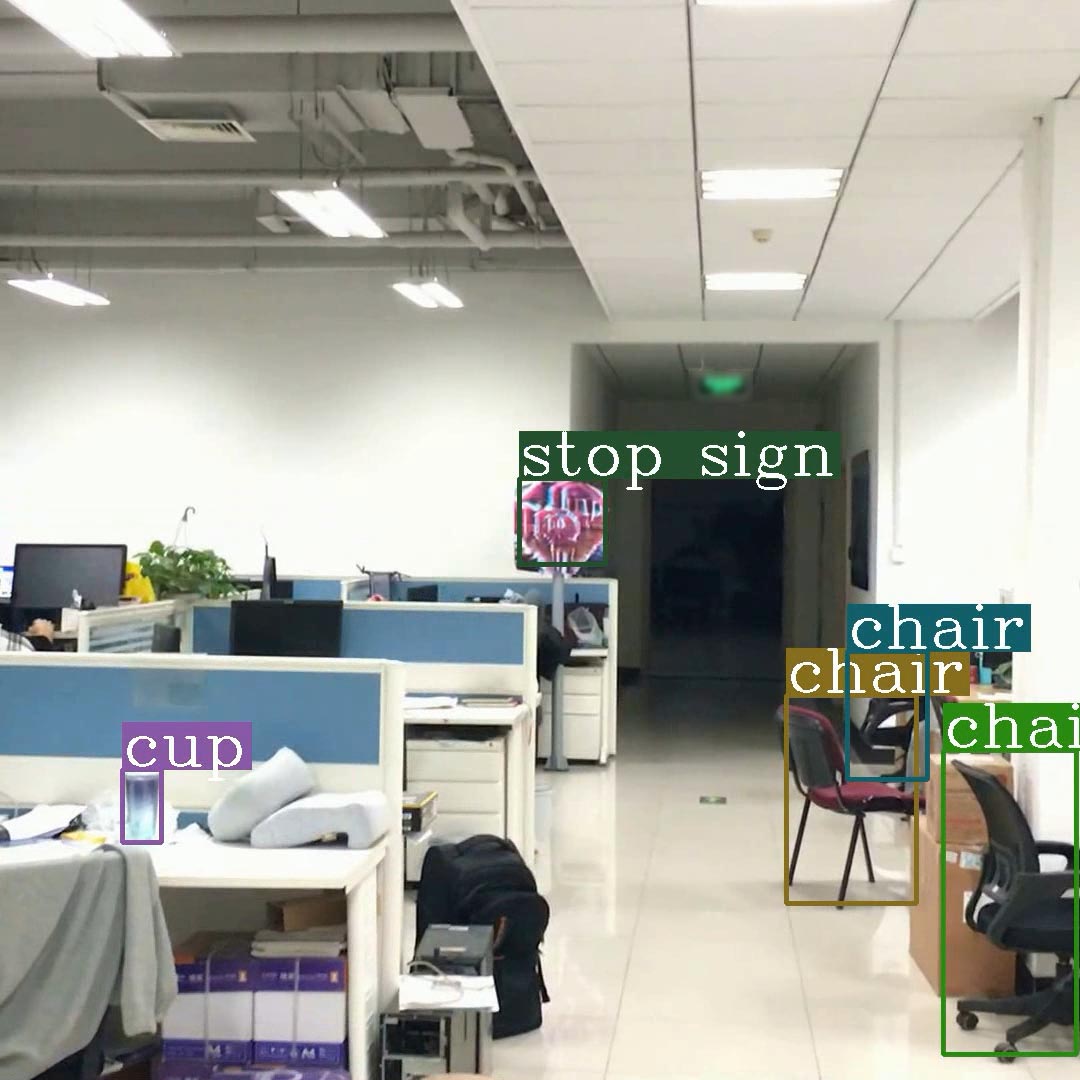, width=0.8\textwidth}

\epsfig{figure=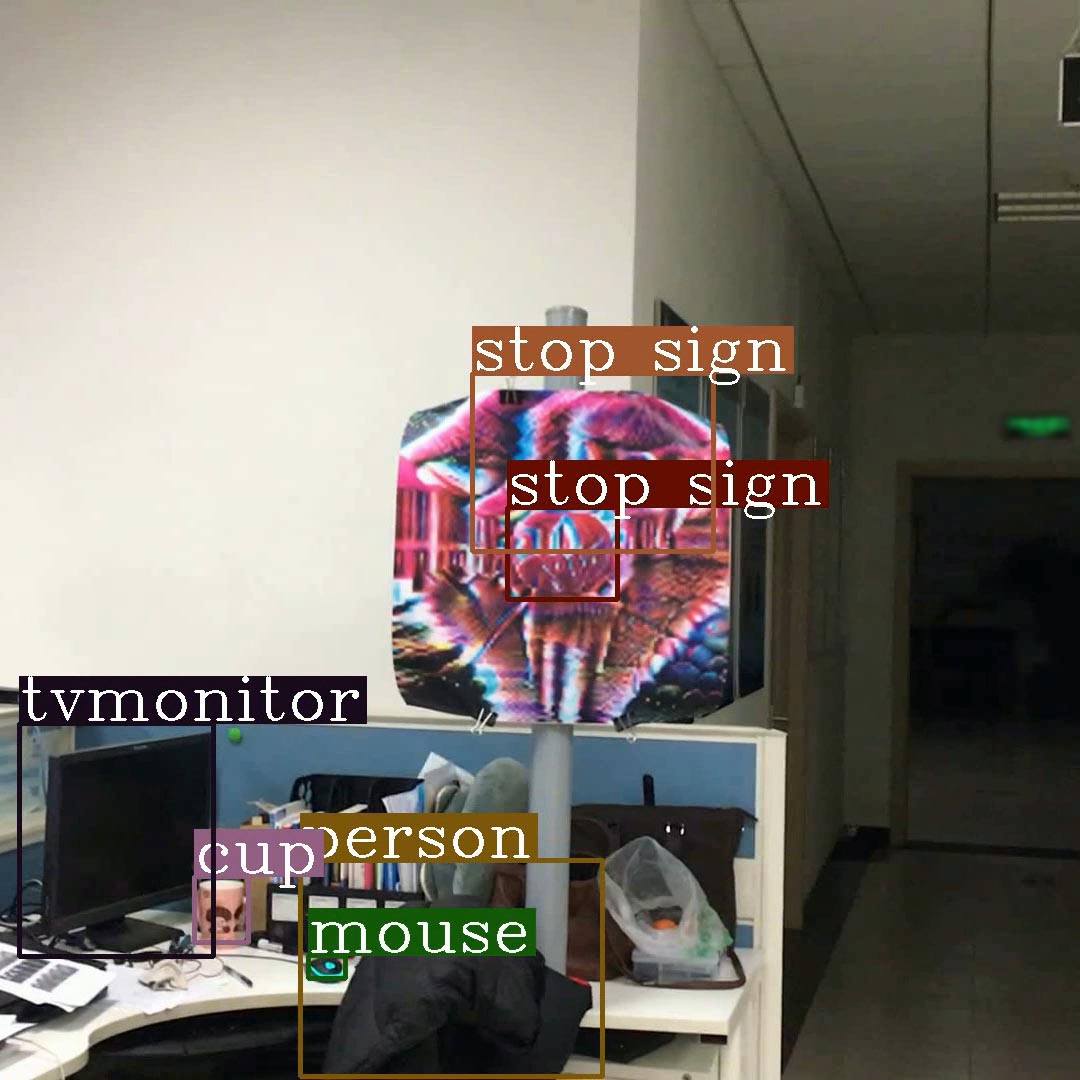, width=0.8\textwidth}
\end{minipage}%
}%
\subfigure[]{
\begin{minipage}[t]{0.18\linewidth}
\centering
\epsfig{figure=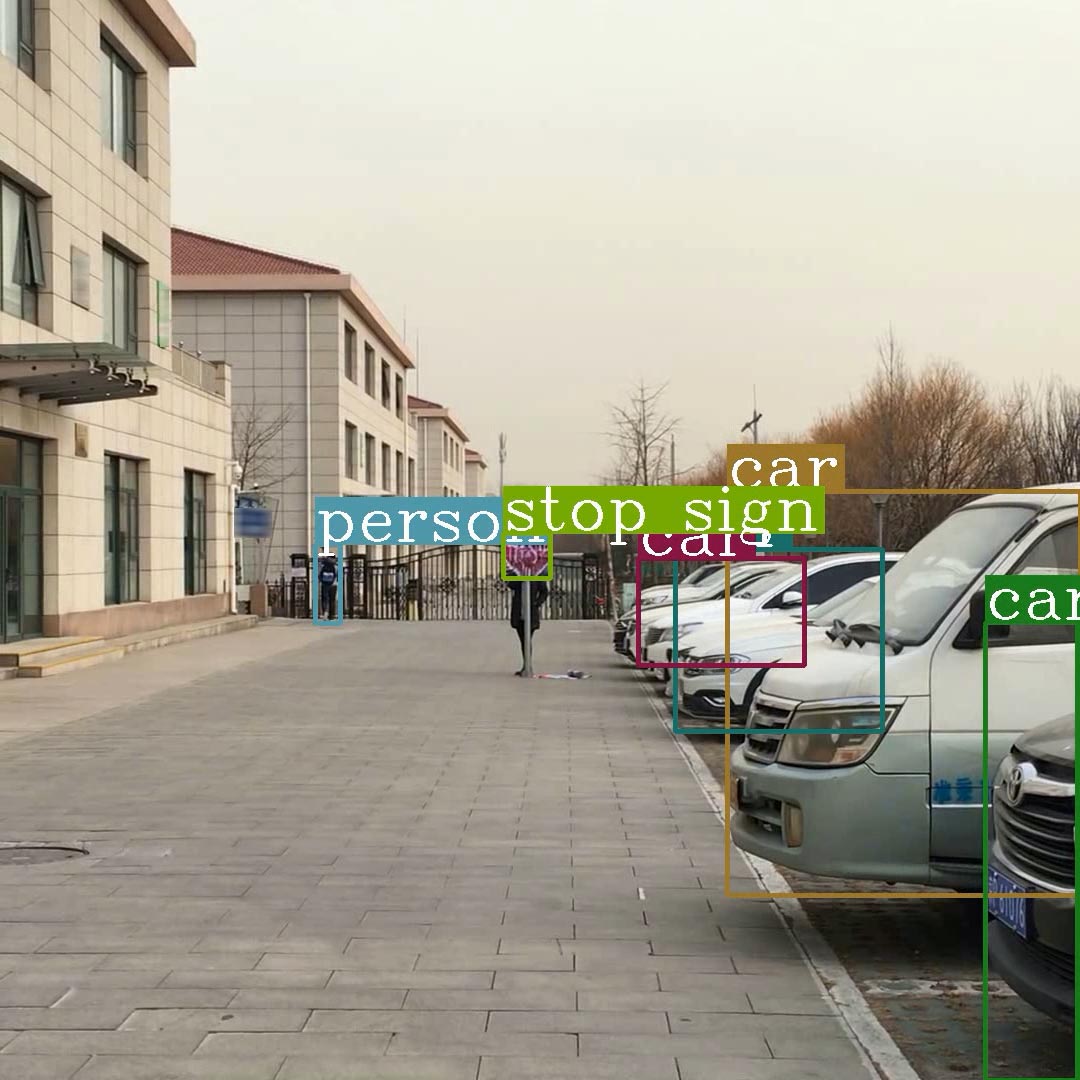, width=0.8\textwidth}

\epsfig{figure=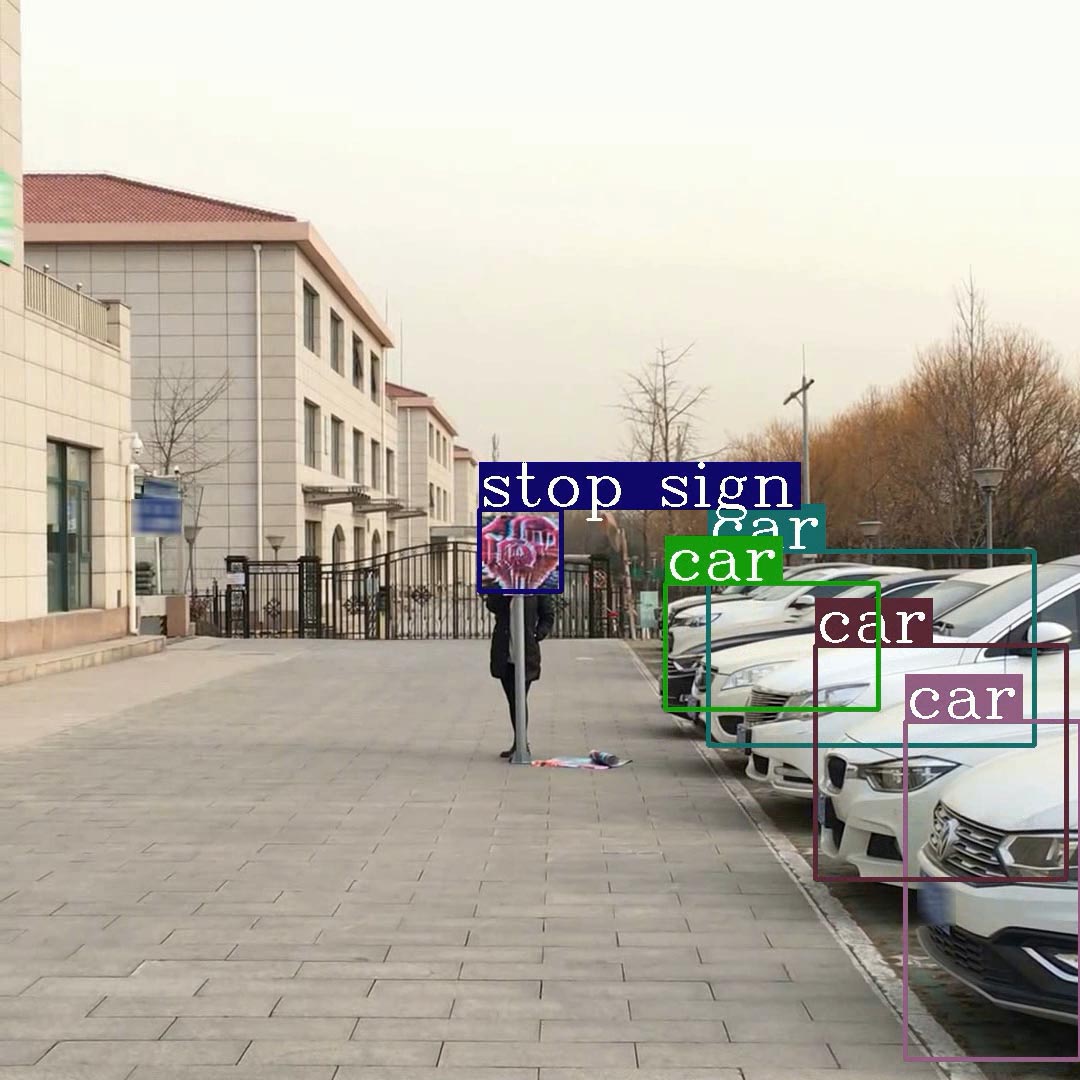, width=0.8\textwidth}

\epsfig{figure=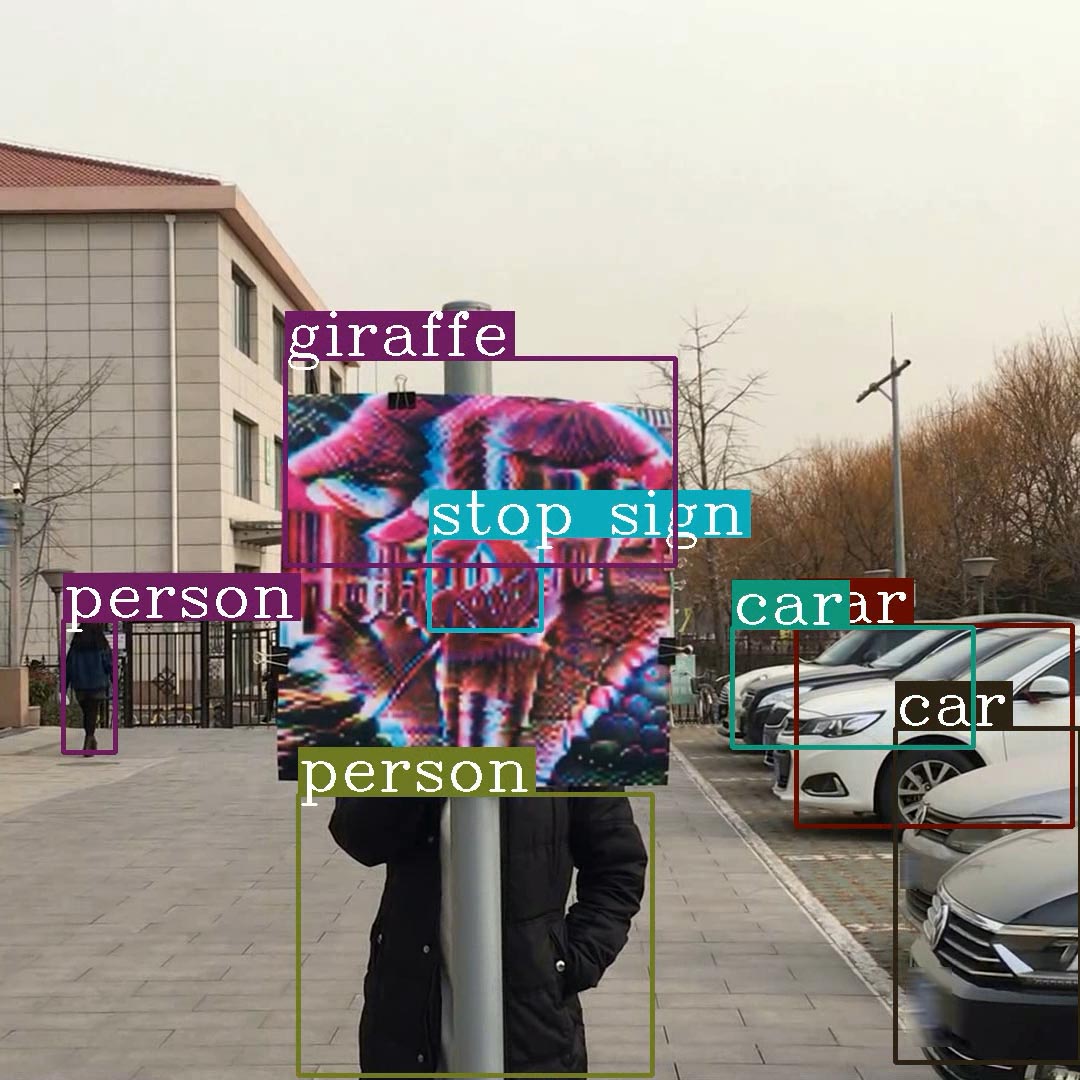, width=0.8\textwidth}
\end{minipage}%
}%
\caption{\textbf{Sample Frames of Hiding Attacks and Appearing Attacks at Different Distances.} (a) The hiding attack indoors. (b) The hiding attack outdoors. (c) The appearing attack indoors. (d) The appearing attack outdoors. }
\label{effectiveness against attack.}
\end{figure*}


\begin{figure*}[ht]
\centering
\subfigure[Hiding attacks at different angles ($0^\circ$, $30^\circ$, $45^\circ$, $60^\circ$).]{
\begin{minipage}[t]{0.20\linewidth}
\centering
\epsfig{figure=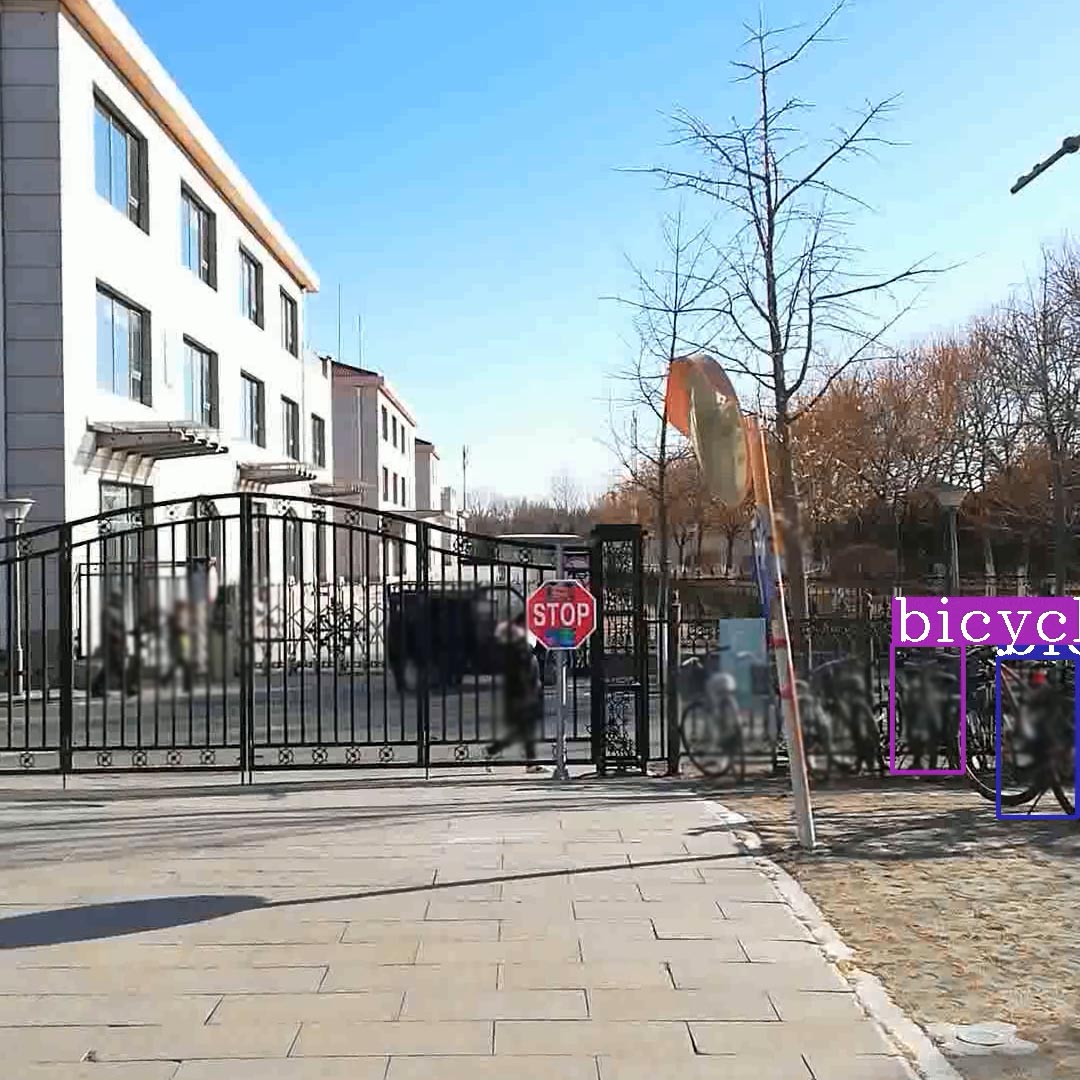, width=0.9\textwidth} 
\end{minipage}%

\begin{minipage}[t]{0.20\linewidth}
\centering
\epsfig{figure=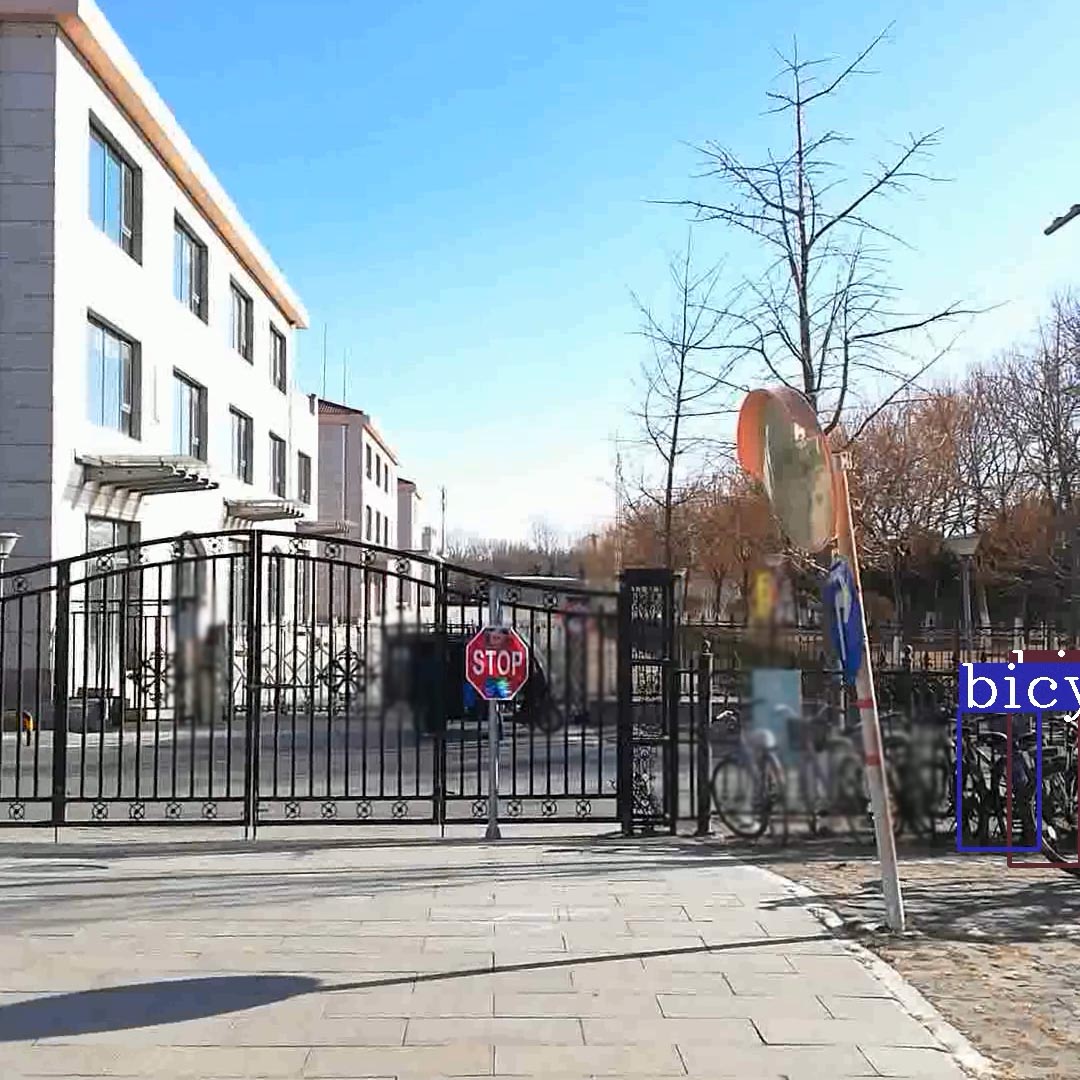, width=0.9\textwidth} 
\end{minipage}%

\begin{minipage}[t]{0.20\linewidth}
\centering
\epsfig{figure=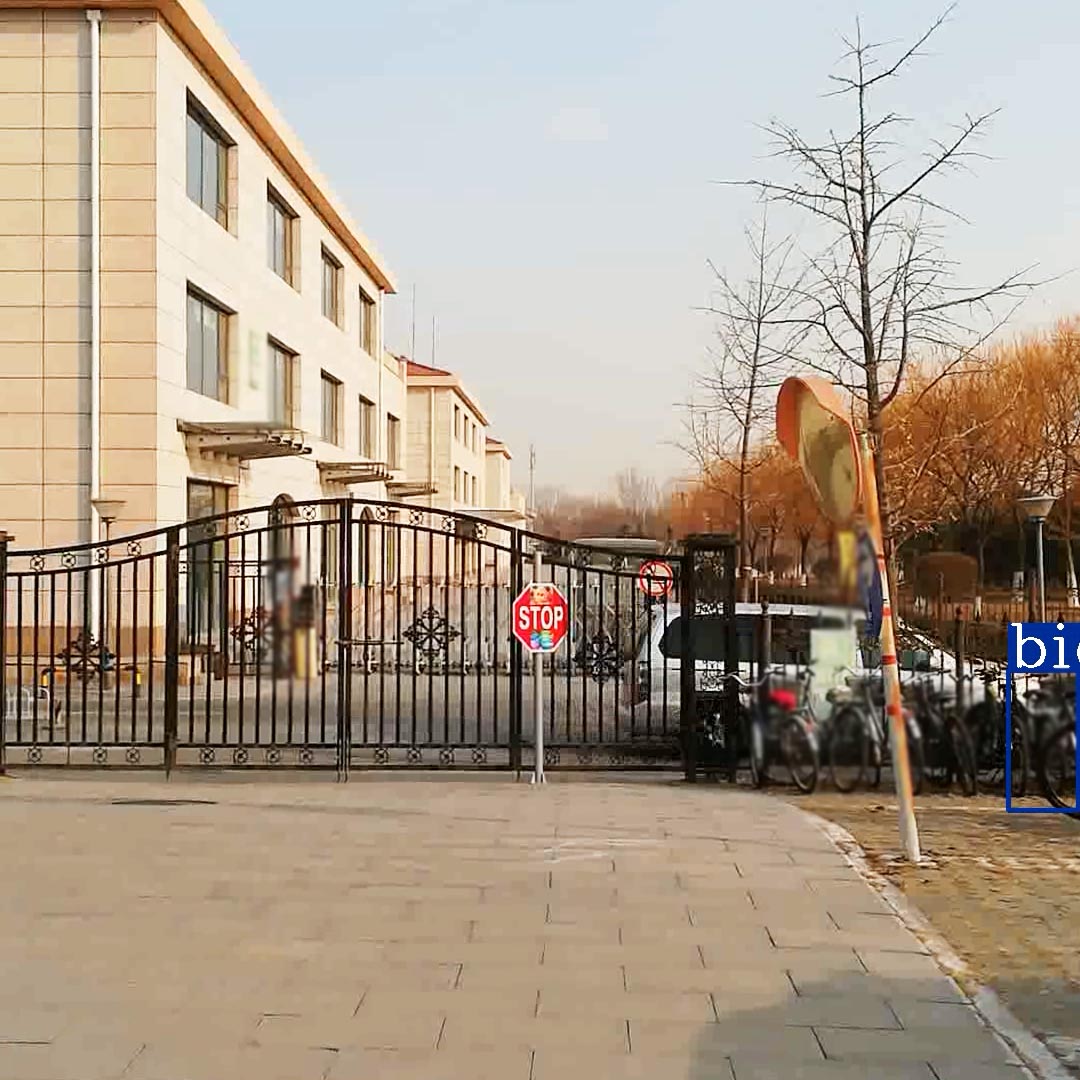, width=0.9\textwidth} 
\end{minipage}%

\begin{minipage}[t]{0.20\linewidth}
\centering
\epsfig{figure=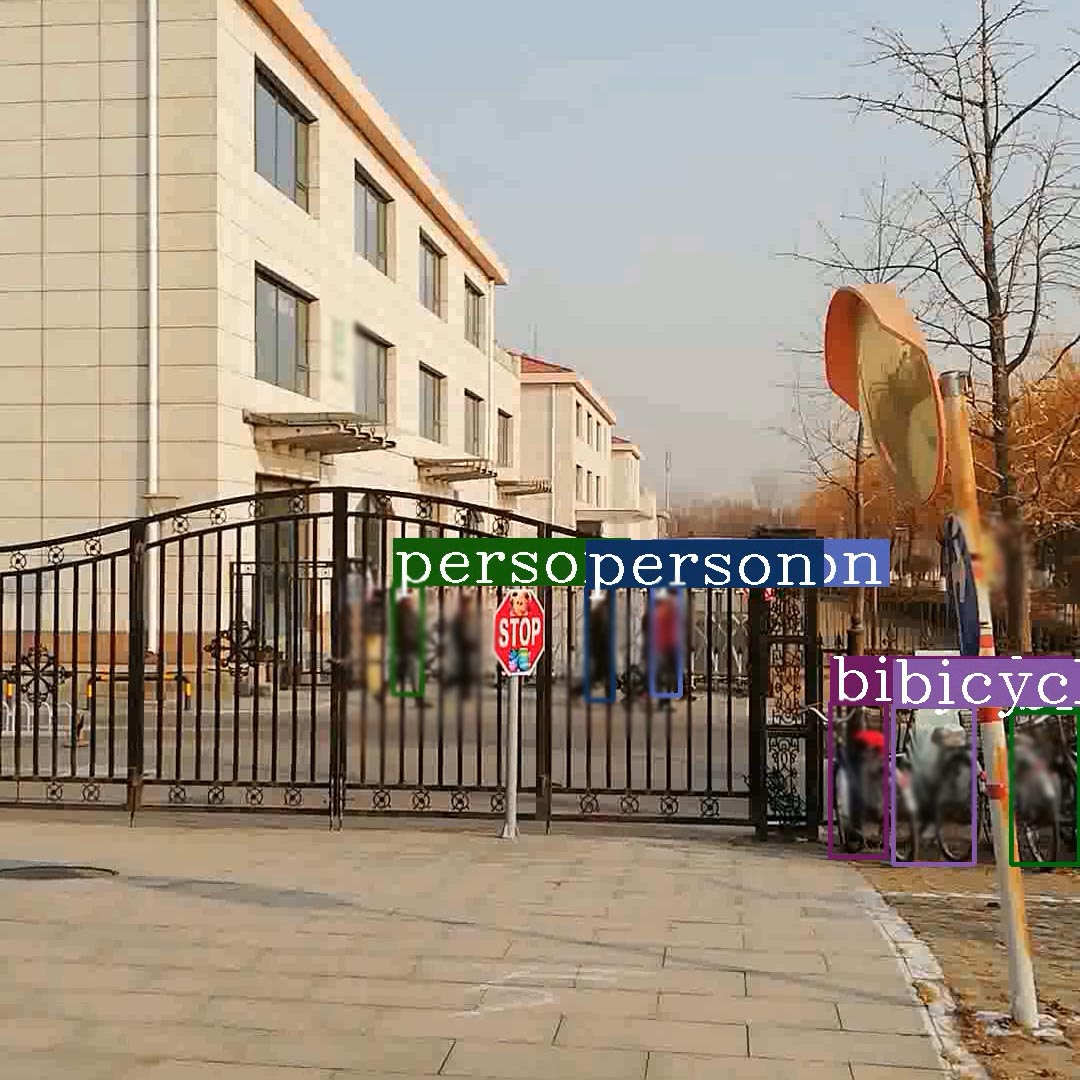, width=0.9\textwidth} 
\end{minipage}%
}%

\centering
\subfigure[Appearing attacks at different angles ($0^\circ$, $30^\circ$, $45^\circ$, $60^\circ$).]{
\begin{minipage}[t]{0.20\linewidth}
\centering
\epsfig{figure=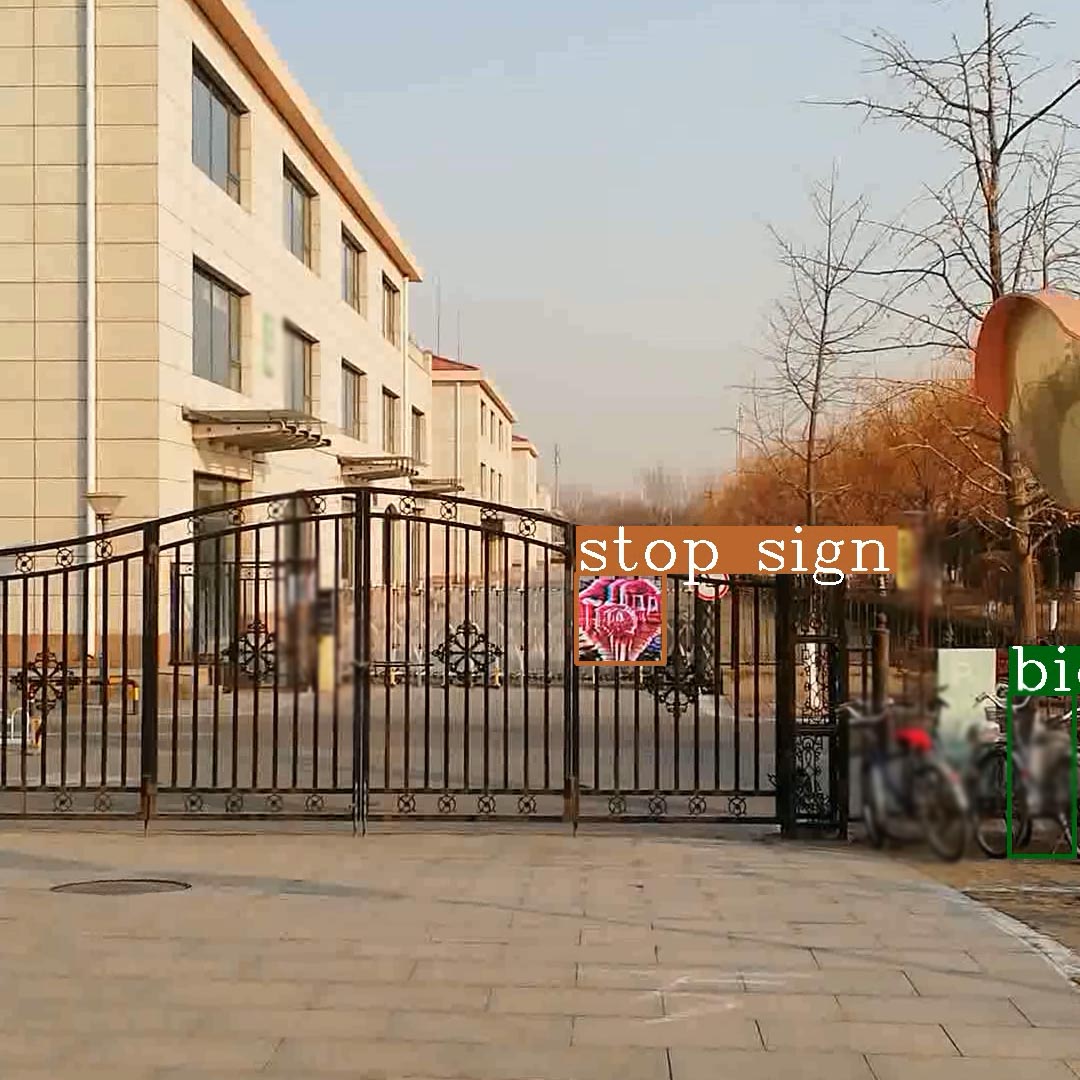, width=0.9\textwidth} 
\end{minipage}%

\begin{minipage}[t]{0.20\linewidth}
\centering
\epsfig{figure=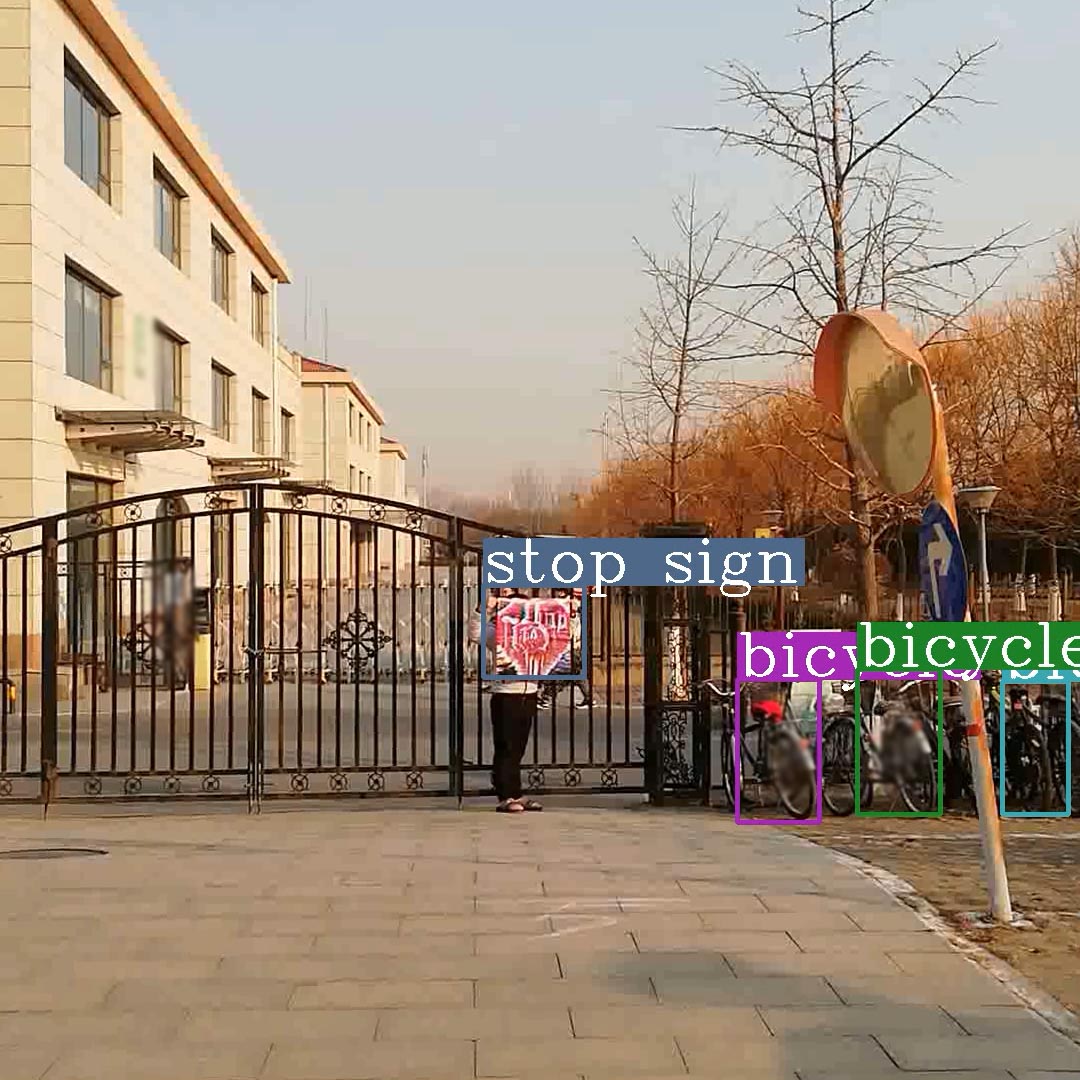, width=0.9\textwidth} 
\end{minipage}%

\begin{minipage}[t]{0.20\linewidth}
\centering
\epsfig{figure=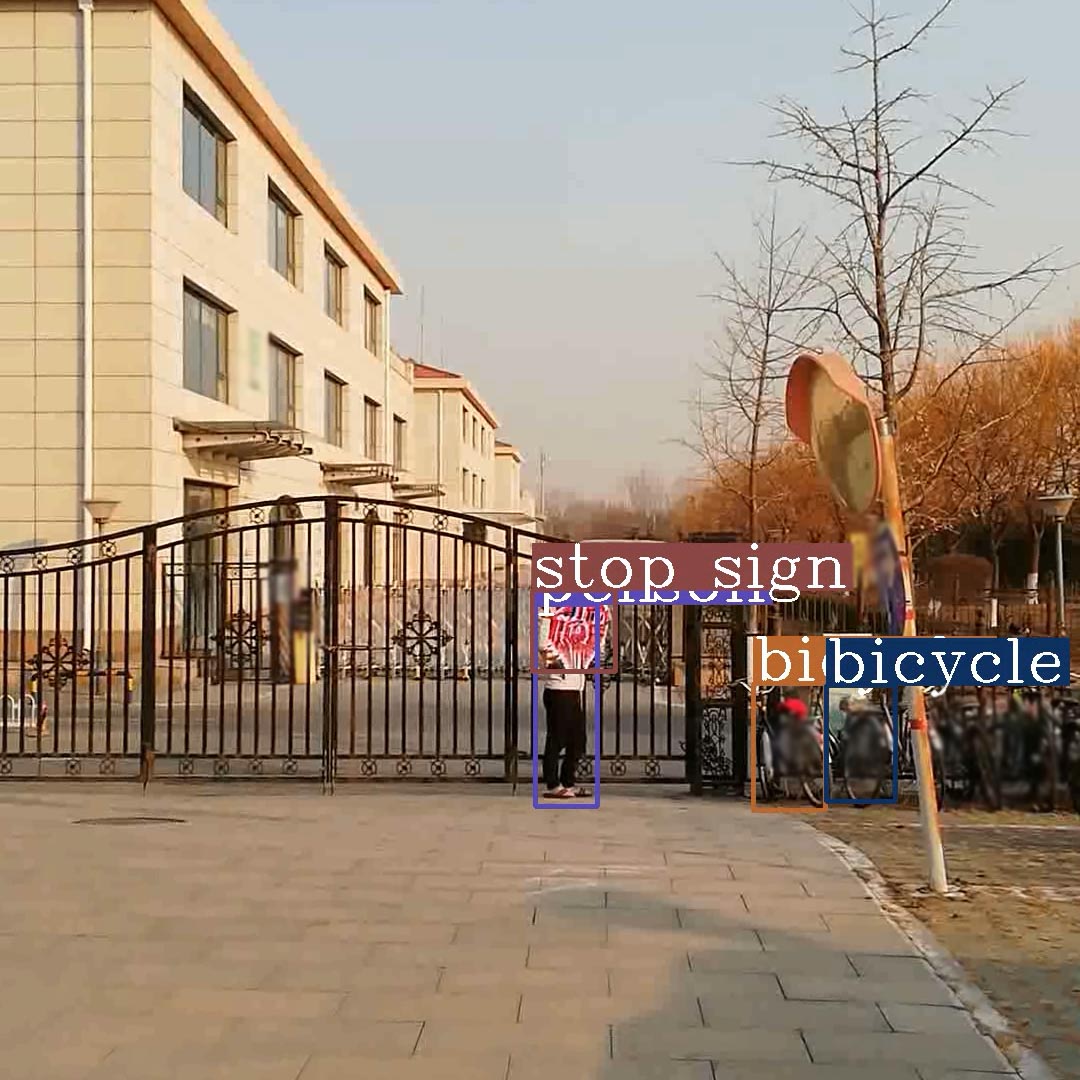, width=0.9\textwidth} 
\end{minipage}%

\begin{minipage}[t]{0.20\linewidth}
\centering
\epsfig{figure=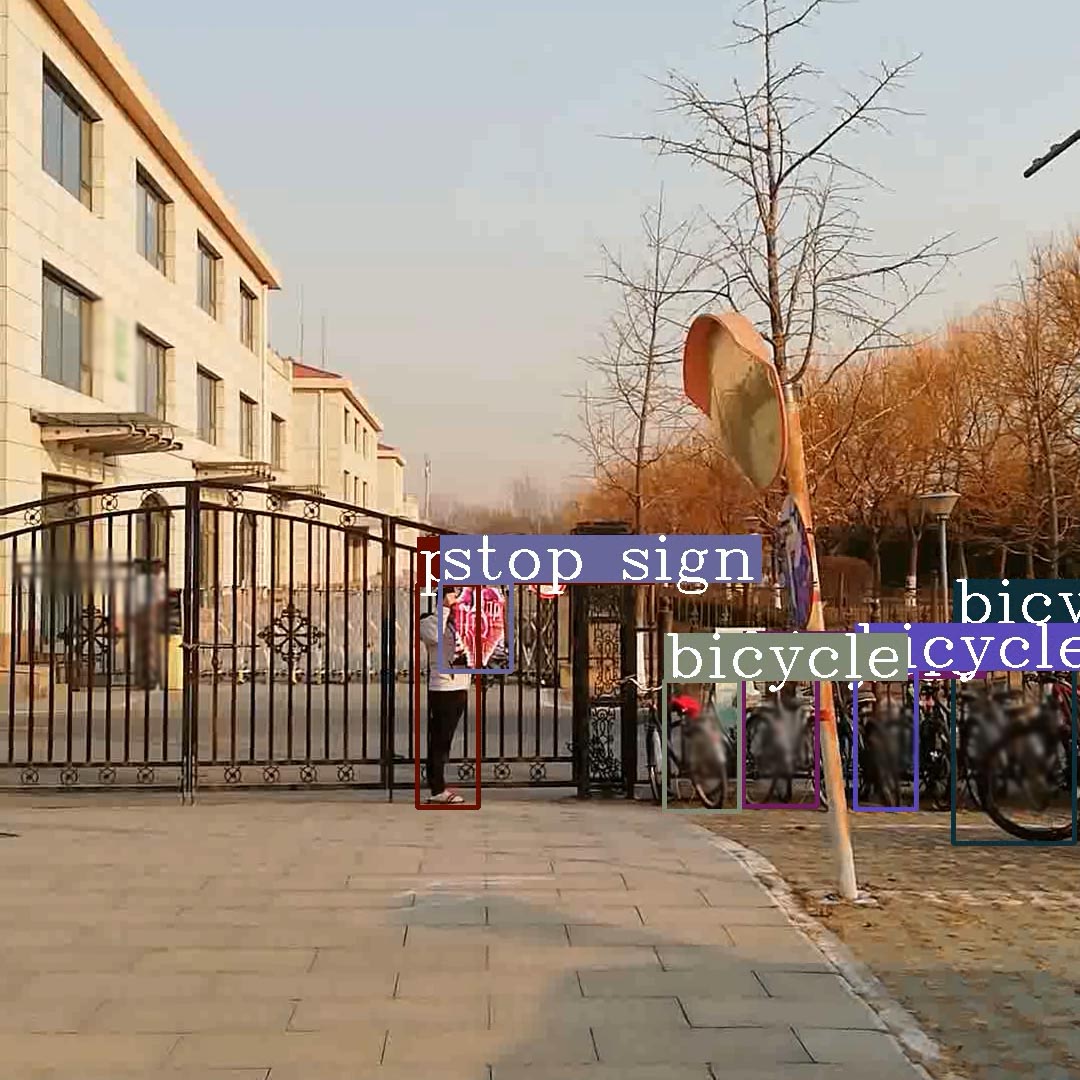, width=0.9\textwidth} 
\end{minipage}%
}%

\caption{\textbf{Sample Frames of Hiding Attacks and Appearing Attacks at Multiple Angles.}  }
\label{Appearance attack in real road test.}
\end{figure*}


\begin{table*}[ht]

\centering
\footnotesize
\caption{Success Rate of the Original Stop Sign on YOLO V3 }
\label{Success rate of the real stop sign on YOLO V3}
\begin{tabular}{m{3cm}
<{\centering}|m{2cm}
<{\centering}|m{2cm}
<{\centering}|m{2cm}
<{\centering}|m{2cm}
<{\centering}}
\hline
\textbf{Success rate($\%$)}& \textbf{5$m\sim$10$m$}& \textbf{10$m\sim$15$m$}& \textbf{15$m\sim$20$m$}& \textbf{20$m\sim$25$m$}\\
\hline \hline
\textbf{$0^\circ$} & \text{100}& \text{100}& \text{100} &\text{93} \\ \hline
\textbf{$30^\circ$}& \text{100} &\text{100} &\text{100} &\text{84} \\ \hline
\textbf{$45^\circ$}& \text{100} &\text{100} &\text{100} &\text{90} \\ \hline
\textbf{$60^\circ$}& \text{100} &\text{98} &\text{93}&\text{72} \\ \hline
\end{tabular}

\end{table*}

\end{onecolumn}


\end{document}